\documentclass{article}

\usepackage[preprint]{neurips_2024}

\usepackage[utf8]{inputenc} 
\usepackage[T1]{fontenc}    
\usepackage{hyperref}       
\usepackage{url}            
\usepackage{booktabs}       
\usepackage{amsfonts}       
\usepackage{nicefrac}       
\usepackage{microtype}      
\usepackage{xcolor}         

\usepackage{graphicx}%
\usepackage{multirow}%
\usepackage{amsmath,amssymb,amsfonts}%
\usepackage{amsthm}%
\usepackage{mathrsfs}%
\usepackage[title]{appendix}%
\usepackage{xcolor}%
\usepackage{textcomp}%
\usepackage{manyfoot}%
\usepackage{booktabs}%
\usepackage{algorithm}%
\usepackage{algorithmicx}%
\usepackage{algpseudocode}%
\usepackage{listings}%
\usepackage{graphicx}
\usepackage{subcaption}
\usepackage{xcolor}
\usepackage{enumitem}
\definecolor{customgreen}{HTML}{548235}
\definecolor{customblue}{HTML}{0070C0}
\definecolor{customblue2}{HTML}{4472C4}

\title{Reverse Physician-AI Relationship: Full-process Clinical Diagnosis Driven by a Large Language Model}

\author{Shicheng Xu$^{1,2,\dagger}$, \; Xin Huang$^{3,\dagger}$, \; Zihao Wei$^{1,2,\dagger}$, \; 
\textbf{Liang Pang$^{1}$\thanks{Corresponding Author \; $\dagger$ Equal Contributions.}, \;   Huawei Shen$^{1}$, \; Xueqi Cheng$^{1}$} \\
$^1$State Key Laboratory of AI Safety, Institute of Computing Technology, CAS\\
$^2$University of Chinese Academy of Sciences \; $^3$Peking University Third Hospital \\
\texttt{\{xushicheng21s,weizihao, pangliang,shenhuawei,cxq\}@ict.ac.cn \; hay221@163.com}
}

\begin{document}

\maketitle

\begin{abstract}
Full-process clinical diagnosis in the real world encompasses the entire diagnostic workflow that begins with only an ambiguous chief complaint. While artificial intelligence (AI), particularly large language models (LLMs), is transforming clinical diagnosis, its role remains largely as an assistant to physicians. This AI-assisted working pattern makes AI can only answer specific medical questions at certain parts within the diagnostic process, but lack the ability to drive the entire diagnostic process starting from an ambiguous complaint, which still relies heavily on human physicians. This gap limits AI’s ability to fully reduce physicians' workload and enhance diagnostic efficiency. To address this, we propose a paradigm shift that reverses the relationship between physicians and AI: repositioning AI as the primary director, with physicians serving as its assistants. So we present DxDirector-7B, an LLM endowed with advanced deep thinking capabilities akin to human ``slow thinking,'' enabling it to autonomously drive the full-process diagnosis with minimal physician involvement. Furthermore, DxDirector-7B establishes a robust accountability framework for misdiagnoses, delineating responsibility between AI and human physicians. In evaluations across rare, complex, and real-world cases under full-process diagnosis setting, DxDirector-7B not only achieves significant superior diagnostic accuracy but also substantially reduces physician workload than state-of-the-art medical LLMs, such as MedFound-176B, as well as general-purpose LLMs such as GPT-4o and DeepSeek-V3-671B. Fine-grained analyses across multiple clinical departments and tasks validate its efficacy, with expert evaluations indicating its potential to serve as a viable substitute for medical specialists. These findings mark a new era where AI, traditionally a physicians' assistant, now effectively drives the entire diagnostic process to drastically reduce physicians' workload, indicating an efficient and accurate diagnostic solution.

\end{abstract}

\section{Introduction}\label{sec1}
Full-process clinical diagnosis encompasses the entire diagnostic workflow connected with clinical decision making~\cite{banning2008review}, beginning with a patient's vague chief complaint. Physicians must iteratively make differential diagnoses, design and interpret a series of appropriate diagnostic tests, and progressively refine their understanding of the patient's clinical information before reaching a definitive diagnosis~\cite{bergus1998clinical,pinnock2014learning,singh2008reducing}. This complex process demands not only extensive medical knowledge and advanced reasoning skills~\cite{woods2006value} but also imposes a substantial workload on physicians. Despite rigorous professional training, the misdiagnosis rate in clinical practice remains close to 20\%~\cite{graber2013incidence,liu2025generalist}. The growing patient demand continues to outpace the diagnostic capacity of physicians, underscoring the urgent need for more efficient and scalable diagnostic solutions~\cite{weinhold2014understanding,world2016global}.

Recent advances in large language models (LLMs), a rapidly evolving artificial intelligence technology, have demonstrated remarkable capabilities in language comprehension and generation. This progress has spurred growing interest in their potential applications in clinical diagnosis. Emerging studies suggest that LLMs exhibit promising diagnostic performance~\cite{clusmann2023future,omiye2024large,pfohl2024toolbox,johri2025evaluation}, prompting the development of medical-specialized LLMs~\cite{singhal2023large,singhal2025toward,zakka2024almanac,liu2025generalist,chen2023meditron,moor2023med,qiu2024towards}. However, the role of LLMs in real-world diagnosis are limited as only assistants for physicians. This limitation arises primarily because current LLMs excel in diagnosing cases with comprehensive clinical data—such as symptoms, medical history, and diagnostic test results, or clear instructions~\cite{liu2025generalist, singhal2025toward,chen2023meditron}—whereas real-world full-process clinical diagnosis often begins with only a patient’s vague chief complaint~\cite{bergus1998clinical,elstein1978medical,singh2008reducing} (Fig.~\ref{framework} (a) and (b)). This makes LLMs can only provide assistance in making final diagnosis or answering specific medical questions at certain parts within the diagnostic process, while much more work of the diagnostic process still highly relies on human physicians, such as clinical reasoning, condition assessment, and designing diagnostic tests to progressively enrich the clinical information.

\begin{figure}[t]
    \centering
    \includegraphics[width=0.9\linewidth]{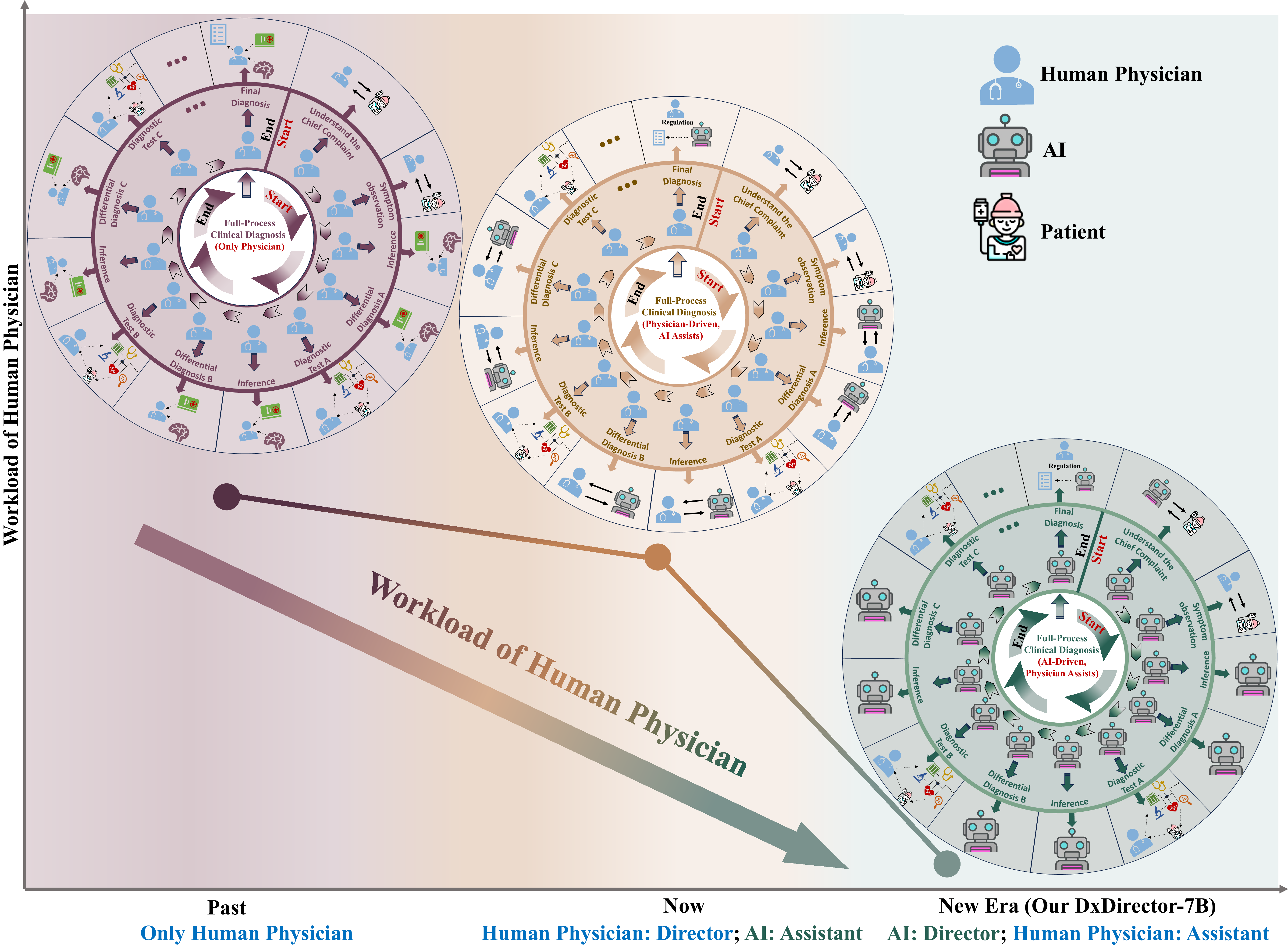}
        \caption{Workflow of full-process diagnosis in past, now and the new era. The inner circle represents the directorship (deciding the specific clinical problems) of multi-step dynamic clinical diagnosis, and the outer circle represents the specific execution of the corresponding steps (solving the clinical problems). In the past, all work is done by human physicians. Now, AI, especially LLM, has reduced physicians' workload to a certain extent, but AI can only serve as an assistant to answer questions designed by physicians at specific steps, and lacks the ability to drive the full diagnosis process starting from the chief complaint, which still relies heavily on physicians. Our DxDirector-7B, marks a new era that AI can drive the full-process diagnosis, only needs physicians as assistants to conduct some clinical operations at necessary steps, reducing the workload of physicians as much as possible.}
        \label{timeline}
\end{figure}

\begin{figure}[t]
    \centering
        \includegraphics[width=0.9\linewidth]{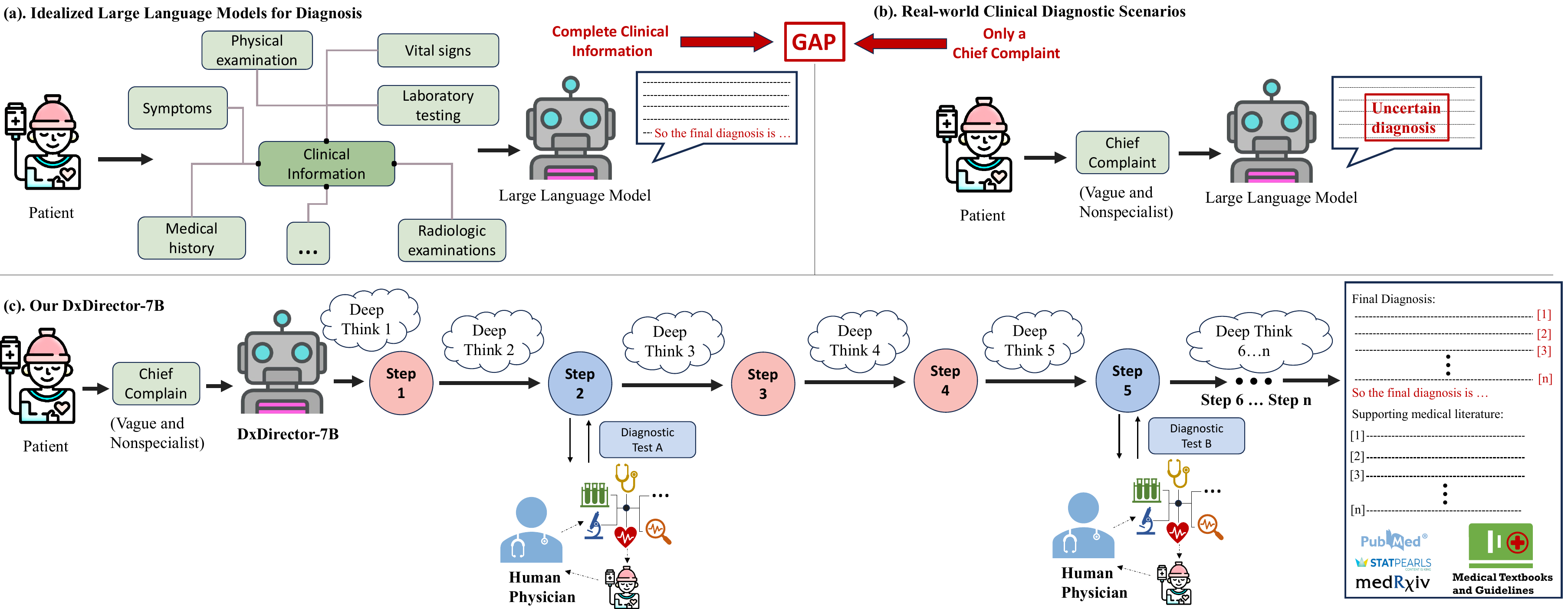}
        \caption{Comparison between our DxDirector-7B and existing LLMs in full-process clinic diagnosis. \textbf{(a). }Existing LLMs are still limited to answering questions with complete clinical information. \textbf{(b). }However, the real-world clinical diagnosis only begins with the patient’s vague and nonspecialist chief complaint. This gap makes that LLMs can only provide assistance in making final diagnosis while much more work before this still highly relies on human physicians. \textbf{(c). }Our DxDirector-7B addresses this by driving the full-process clinical diagnosis step-by-step, only requesting assistance from physicians at necessary steps with the principle of minimizing physicians’ involvement.}
        \label{framework}
\end{figure}

To address this challenge, we propose a paradigm shift in the role of LLMs in clinical diagnosis. Unlike existing LLMs, which function solely as assistants to physicians, our approach redefines this relationship by positioning the LLM as the primary director of the diagnostic process, with physicians serving as its assistants (Fig.~\ref{timeline}). As described in the new era of Fig.~\ref{timeline}, LLM drives the full-process clinical diagnosis. At the beginning, LLM can only access the patient's vague chief complaint, and it gradually clarifies the patient's condition, designs appropriate diagnostic tests, infers complex medical knowledge and clinical phenomena, and finally makes the diagnosis. During this process, LLM can dynamically request assistance from physicians only when it faces the clinical operations that the computer program-based LLM cannot complete, such as symptom observation, laboratory testing, physical examination, and so on. Physicians are LLM's assistants to finish its requests and input the results back to it. LLM continues the subsequent diagnosis. The pattern of LLM requesting assistance adheres to the principle of minimizing physicians' involvement, thereby reducing the workload and requirement for medical expertise of physicians as much as possible.

Building on this design, we introduce DxDirector-7B, an LLM with advanced deep thinking capabilities (like human ``slow thinking''), capable of autonomously driving the full-process clinical diagnosis starting from a vague chief complaint. DxDirector-7B progressively executes the entire clinic diagnosis step-by-step, it performs deep thinking to determine the optimal strategy at each step and seeks assistance from human physicians only at necessary steps with the principle of minimizing physicians' involvement. It dynamically assesses whether sufficient clinical information has been gathered to establish a final diagnosis or whether further diagnostic steps are required. The final diagnostic output includes a clear and comprehensive summary of the entire diagnosis process, with each fine-grained medical knowledge attached by authoritative medical literature, thereby enhancing the verifiability of AI-generated diagnoses. Furthermore, this structured output establishes a robust accountability framework between physicians and the LLM, ensuring traceability in misdiagnosis.

We evaluate DxDirector-7B in the full-process clinical diagnosis setting using both real-world scenarios and four authoritative publicly available datasets. The evaluation datasets comprise 26,018 cases, including rare, complex, and diagnostically challenging cases reported in NEJM Clinicopathologic Cases~\cite{brodeur2024superhuman}, cases from the U.S. Medical Licensing Examination~\cite{jin2021disease}, and cases of real-world inpatients from officially certified Grade 3A hospitals in China. To ensure a comprehensive assessment, we conduct fine-grained evaluations across 19 clinical departments (e.g., neurosurgery, oncology, endocrinology) and 12 clinical tasks (e.g., diagnosis, differential diagnosis, treatment). Experimental results indicate that in terms of full-process clinical diagnostic accuracy, our DxDirector-7B significantly surpasses the medical adaptad LLMs with dozens of times more parameters, such as MedFound-176B~\cite{liu2025generalist} and OpenbioLLM-70B. It also significantly surpasses the current strongest commercial general-purpose LLMs with nearly 100 times more parameters, such as GPT-4o, o1-preview, o3-mini,and Deepseek-V3-671B. Notably, DxDirector-7B achieves this superior performance while requiring significantly lower physician involvement than all comparison LLMs. These findings show that DxDirector-7B achieves the best accuracy with significantly lower computational and training costs, requiring the lowest human physicians' efforts in the entire diagnosis process. Evaluations with the participation of medical specialists show that in real-world full-process diagnostic scenarios, the diagnoses generated by our DxDirector-7B achieve substitution for medical specialists in 60\% to 75\% of cases in many departments such as pulmonology and gastroenterology. Further analysis highlights its ability to provide a fine-grained verification framework for AI-generated diagnoses, establishing a robust accountability mechanism between physicians and AI in misdiagnosis.

This paper marks a new era where AI, traditionally a physicians' assistant, now leads the entire diagnostic process with minimal physician involvement. It advances effective AI deployment in full-process clinical workflows of the real world, reducing the workload of physicians to the greatest extent possible and indicating the efficient, accurate and scalable diagnostic solution.

\section{Results}\label{sec2}
\vspace{-0.75em}

In this section, we first give an overview of our DxDirector-7B and its training method. Then, we present the comprehensive experimental results about comparing DxDirector-7B with the most advanced medically adapted LLMs and commercial general-purpose LLMs.
\begin{figure}[t]
    \centering
        \includegraphics[width=1.0\linewidth]{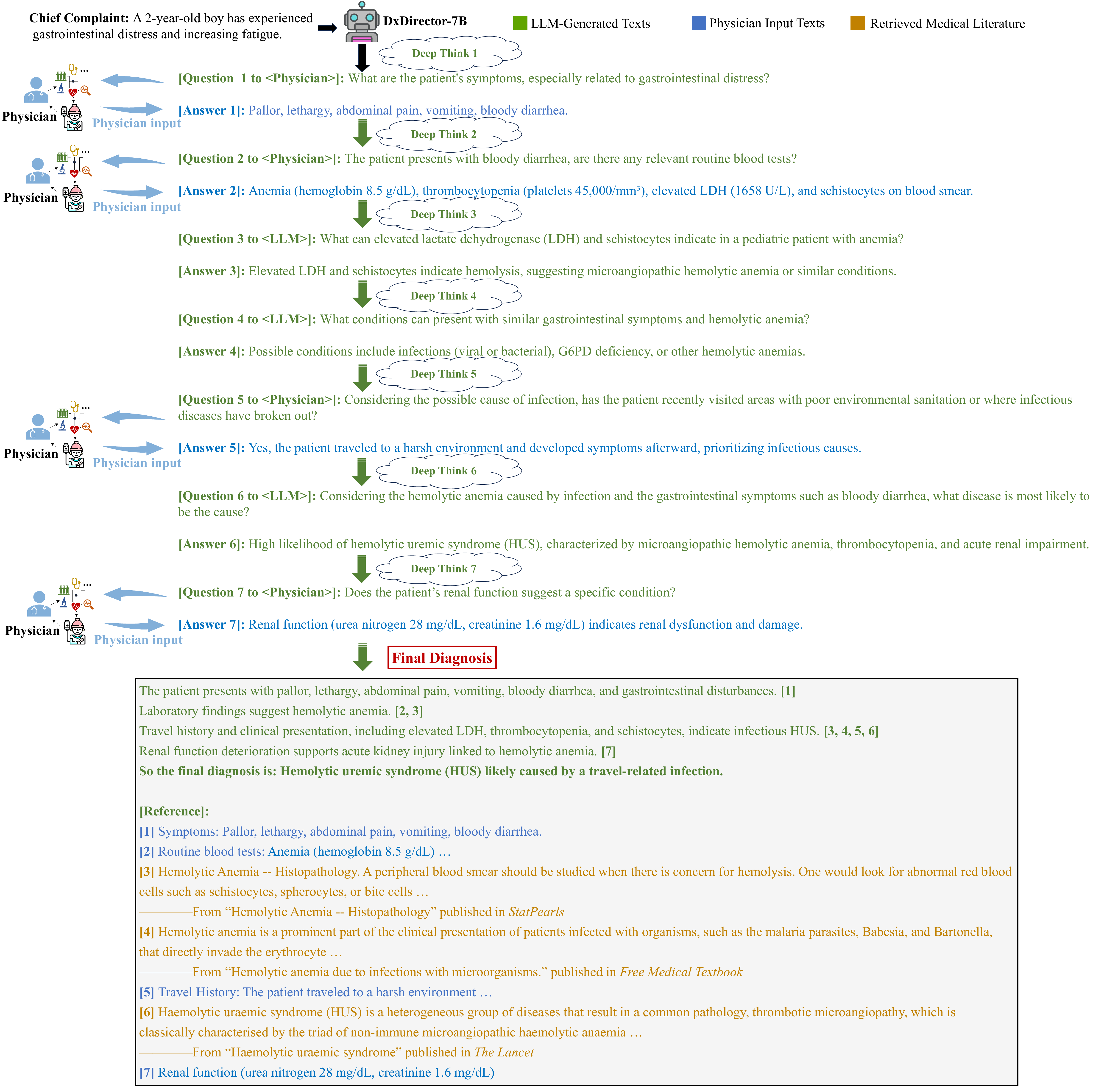}
        \caption{A case of DxDirector-7B performing the full-process diagnosis starting with only a chief complaint. Most of this process is driven by DxDirector-7B step-by-step reasoning (\textcolor{customgreen}{green texts}) and physicians only need to follow its instructions to complete some basic clinical operations (\textcolor{customblue}{blue texts}).}
        \label{DxDirector}
\end{figure}

\vspace{-0.5em}
\subsection{Overview and Training Method of DxDirector-7B}
\subsubsection{Overview}
The overall workflow of our DxDirector-7B is shown in Fig.~\ref{framework}~(c) and the practical case of DxDirector-7B is illustrated in Fig.~\ref{DxDirector}. As shown in Fig.~\ref{framework} (a) and (b), current LLMs typically rely on complete clinical information for diagnosis, a scenario that rarely aligns with real-world clinical practice, where initial information often consists solely of a vague chief complaint. This means human physicians still need to pay much workload between getting chief complaint and making final diagnosis, and LLMs are just assistants in certain parts of this entire complex process. To address this disparity, we introduce DxDirector-7B, an advanced LLM with powerful deep thinking ability to drive the full-process clinical diagnosis starting with only a patient's vague chief complaint, which is much closer to real-world clinical diagnosis than existing LLMs.

In Fig.~\ref{framework}~(c), DxDirector-7B systematically executes complex clinical diagnosis in a stepwise manner, interconnected by a detailed reasoning process—termed ``deep thinking''—that mimics human ``slow thinking'' cognitive strategy. This deep thinking incorporates current clinical data and integrates overarching diagnostic objectives, guiding the identification of critical questions at each diagnostic stage (denoted as "\texttt{[Question]}" in Fig.~\ref{DxDirector}). If the question requires objective medical knowledge or inference (automatically marked with ``\texttt{<LLM>}''), DxDirector-7B will generate the answer to the question by itself. If the question requires clinical operations for diagnostic test, such as medical imaging, physical examinations, laboratory tests that computer program-based LLMs cannot complete, it will actively request assistance from human physicians (automatically marked with ``\texttt{<Physician>}''), who will input the results of the operation as an answer. The deep thinking at each step ensures that the step is correct and efficient, so that the diagnosis can be completed accurately while relying on the minimal human physicians' efforts.

When DxDirector-7B determines that the diagnosis is complete, it synthesizes the preceding steps into a succinct summary ("\texttt{[Final Diagnosis]}" in Fig.~\ref{DxDirector}). It can attach the authoritative medical literature retrieved by a medical search model or physicians' operations involved in each step. This improves the verifiability of the generated diagnosis at fine-grained level. Additionally, it clearly delineates the actions of LLMs and physicians to establish a precise accountability framework, which is critical in potential medical incidents. More practical cases are in Extended Data Fig.~\ref{case1} to Fig.~\ref{case3-bs}.

Compared to existing state-of-the-art LLMs, DxDirector-7B achieves superior diagnostic accuracy while markedly reducing both the clinical workload and expertise required of human physicians. As discussed in Fig.~\ref{timeline}, the development of DxDirector-7B signals a paradigm shift in clinical practice, fundamentally redefining the collaborative dynamics between AI and healthcare professionals and providing an efficient, accurate and scalable diagnostic solution.

\subsubsection{Training Method}
Our training method for DxDirector-7B includes three stages: (1) Continued pre-training on medical data; (2) Instruction-tuning for full-process diagnosis; (3) Step-level strategy preference optimization.

The first stage is consistent with existing medical-adaptation methods for LLMs~\cite{chen2023meditron,liu2025generalist}. We continued pre-train Llama-2-7B~\cite{touvron2023Llama} on large-sclae medical data such as clinical guidelines, PubMed papers, and so on. This stage enables the general LLMs to acquire medical knowledge, which forms the foundation for its clinical diagnosis capabilities. Details about this can be found in Section~\ref{pretrain}.

The second stage is instruction-tuning for full-process clinical diagnosis. This stage enables our DxDirector-7B to drive the full-process clinical diagnosis solely starting with ambiguous chief complaints, through the step-by-step reasoning and continuous deep thinking. The training dataset is constructed based on publicly available medical question-answering data~\cite{jin2021disease}. We use general-purpose LLM GPT-4o to convert patients' case reports in the datasets into step-by-step reasoning, and use powerful reasoning LLM o1-preview to enrich the thinking process of each step (details of this can be found in Section~\ref{sft_section}). The automated construction process of this synthetic data is supervised by medical experts. After data construction, we get $10,178$ high-quality instruction-response pairs covering multiple clinical tasks such as diagnosis, differential diagnosis, designing treatment plan, screening, analyzing etiology, and so on. Instruction tuning based on this dataset endows our DxDirector-7B with the preliminary capability to drive a full-process clinical diagnosis and perform deep thinking. The technical details of training can be found in Section~\ref{sft_section}.

The third stage is step-level strategy preference optimization. We call the question to be solved in each step derived by deep thinking of DxDirector-7B as “strategy”. After the second stage, our DxDirector-7B can generate the strategy step-by-step just like Fig.~\ref{framework} (c). The third stage enables DxDirector-7B to implicitly compare multiple potential strategies in deep thinking at each step and select the optimal strategy. This ensures that each step in complex clinical reasoning is correct and efficient, so that the diagnosis can be completed accurately while relying on the minimal human physicians’ efforts. The optimization of this stage is performed at step-level. In training data construction, we use multiple sampling to make DxDirector-7B generate multiple different strategies for each step (given the same prefix) and assign different rewards to these strategies. The reward value is determined by both the correctness of the final answer and the quantified physician workload derived from the strategies. Strategies with more correct answers are assigned higher rewards. For strategies with the same correct answers, the strategies that seek more assistance from human physicians will have a lower reward value. In training, DxDirector-7B learns to refine deep thinking to generate the strategy with the highest reward by reward-based reinforcement learning~\cite{wirth2017survey} with the principle of ensuring correctness while minimizing the workload of human physicians. Details can be found in Section~\ref{sspo}. 

\vspace{-0.8em}
\subsection{Overview of Experiments}
\paragraph{Evaluation Datasets}
The evaluation datasets consist of two parts: one is four publicly available medical datasets evaluated automatically based on their provided correct answers, and the other is set of cases in real-world clinical diagnosis with the evaluation participated by medical specialists.

\vspace{-0.2em}
For the publicly available datasets, we first collect raw data and then reconstruct them to simulate full-process clinical diagnosis scenarios, in which LLM is initially only provided with the patient's chief complaint while additional clinical information that helps make the definitive diagnosis needs to be gradually inferred or obtained through its active reasoning process. Four datasets are utilized: (1) \textbf{NEJM Clinicopathologic Cases}~\cite{brodeur2024superhuman}, it covers 344 clinical cases published by the New England Journal of Medicine between 2014 and 2024. These cases are highly complex, rare, and educationally significant. (2) \textbf{RareArena}~\footnote{\url{https://github.com/zhao-zy15/RareArena}} is a dataset of nearly 50,000 rare disease diagnoses extracted from case summaries in PubMed Central, covering 4,597 rare disease types. We use the rare disease confirmation of it, which covers 22,901 data samples. (3) \textbf{ClinicalBench}~\cite{yan2024clinicallab} is a multi-departmental clinical diagnostic evaluation benchmark includes 1,500 real-world cases that cover 150 diseases. (4) \textbf{US Medical License Exam}~\cite{jin2021disease}, it is a set of 1,273 challenging medical questions in the US Medical License Exam. There are many tasks in this dataset such as diagnosis, differential diagnosis, treatment planning, and so on.

\vspace{-0.2em}
To simulate the full-process clinical diagnosis beginning with only a patient's initial chief complaint, we reconstruct the four datasets as follows. For each data instance, we first employ GPT-4o API~\footnote{\url{https://api.openai.com/v1/chat/completions}.} to extract all clinical information (patient's profile, disease symptoms and histories, drug dosage requirements, diagnostic test results, and so on.). Next, we utilize GPT-4o API to transform medically precise clinical descriptions into vague chief complaints characteristic of real patients. Both steps leverage the in-context learning approach\cite{dong2024survey}, guided by explicit instructions and exemplar cases curated by medical experts. In this way, each data instance is reformulated as a triplet comprising a clinical diagnostic question, an initial patient chief complaint, and detailed clinical information. At the beginning of the diagnosis, LLMs can only access the chief complaint, while additional clinical information needs to be gradually inferred or obtained through its active reasoning.

\vspace{-0.2em}
For the real-world clinical diagnosis, we construct real clinical diagnostic scenario within an officially certified Grade 3A hospital in China~\footnote{Grade 3A hospitals are the highest level hospitals in China’s “three-grade, six-class” classification system.}. This evaluation covers 160 real cases across 9 different clinical departments. We introduce the medical specialists in each department to participate in the evaluation of the diagnostic results generated by LLMs. The specific details about this can be found in Section~\ref{beiyi}. This experiment has been approved by the hospital's Ethics Review Committee (IRB00006761-M20250173). To safeguard patient privacy, any personally identifiable information (PII) or other sensitive details have been manually identified and removed by the medical team.

\vspace{-0.5em}
\paragraph{Full-process Clinical Diagnosis Setting for Evaluation}
Based on above datasets, we construct a full-process clinical diagnosis setting to evaluate the performance of various LLMs. In this setting, each data instance consists of a question, a patient's chief complaint and detailed clinical information. Initially, the LLM has access solely to the chief complaint and is tasked with addressing questions related to diagnosis, treatment strategies, etiology, and so on. Any additional clinical details must subsequently be inferred or actively obtained through stepwise reasoning. When encountering tasks need clinical operations that the computer program cannot complete, such as symptom observation, laboratory testing, physical examination, and so on, the LLM proactively requests assistance from human physicians. To automatically simulate this physician interaction on large-scale dataset, we implement an AI agent powered by GPT-4o, which receives real-time queries from the LLM, interprets the requested clinical information, and provides relevant data extracted from detailed clinical information to LLM, allowing LLM to continue reasoning. This framework effectively replicates realistic interactions in full-process clinical diagnosis, where LLM asks the assistance from physicians. Our DxDirector-7B has ability to actively perform full-process clinical diagnosis for patient's chief complaint while all baselines do not. Because existing LLMs tend to directly make a diagnosis, even when the current clinical information is vague and insufficient (as shown in Fig~\ref{framework} (a) and (b)). So we design specific prompts (Supplementary Fig. 21) to guide the baselines in completing this with multi-round conversation between themselves and the simulated physicians.

\vspace{-1.0em}
\paragraph{Baselines}
The baselines in the experiments can be divided into two categories:
\vspace{-0.8em}
\begin{enumerate}[leftmargin=*, noitemsep] 
    \item One is the current most powerful commercial general-purpose large language models including Deepseek-v3-671B~\cite{liu2024deepseek}, GPT-4o~\cite{achiam2023gpt}, OpenAI o1-preview~\cite{jaech2024openai}, OpenAI o3-mini~\footnote{\url{https://openai.com/index/o3-mini-system-card/}}, Gemini-2.0-flash~\cite{team2023gemini}. These LLMs boast hundreds of billions of parameters and are developed by tech giants (OpenAI, Google, Microsoft and Deepseek) at immense training costs. Recent study has shown that they have promising performance in making the final clinical diagnosis~\cite{brodeur2024superhuman,kanjee2023accuracy}.
    \item The other is the open source LLMs specifically optimized for medical domain including Meditron-70B~\cite{chen2023meditron}, OpenbioLLM-70B~\footnote{\url{https://huggingface.co/aaditya/Llama3-OpenBioLLM-70B\#}}, Clinical Camel-70B~\cite{toma2023clinical} and Meditron-176B~\cite{liu2025generalist} and a open source general LLM Llama-3-70B~\cite{grattafiori2024Llama}. They have significantly larger parameters than our DxDirector-7B (70B, 176B vs. our 7B), which means they are more expensive to train and infer.
\end{enumerate}

\vspace{-1.2em}
\subsection{Accuracy of Clinical Diagnosis} \label{sec:acc}

\vspace{-0.25em}
This section reports the experimental results about clinical diagnostic accuracy of various LLMs on NEJM Clinicopathologic Cases, RareArena and ClinicalBench in full-process diagnosis setting. Our DxDirecotr-7B achieves the highest accuracy on all these three datasets and outperforms human physicians on complex cases. The detailed results and analysis are as follows. 

\begin{figure}[H]
  \centering
    \begin{subfigure}[b]{0.95\textwidth}
    \includegraphics[width=\linewidth]{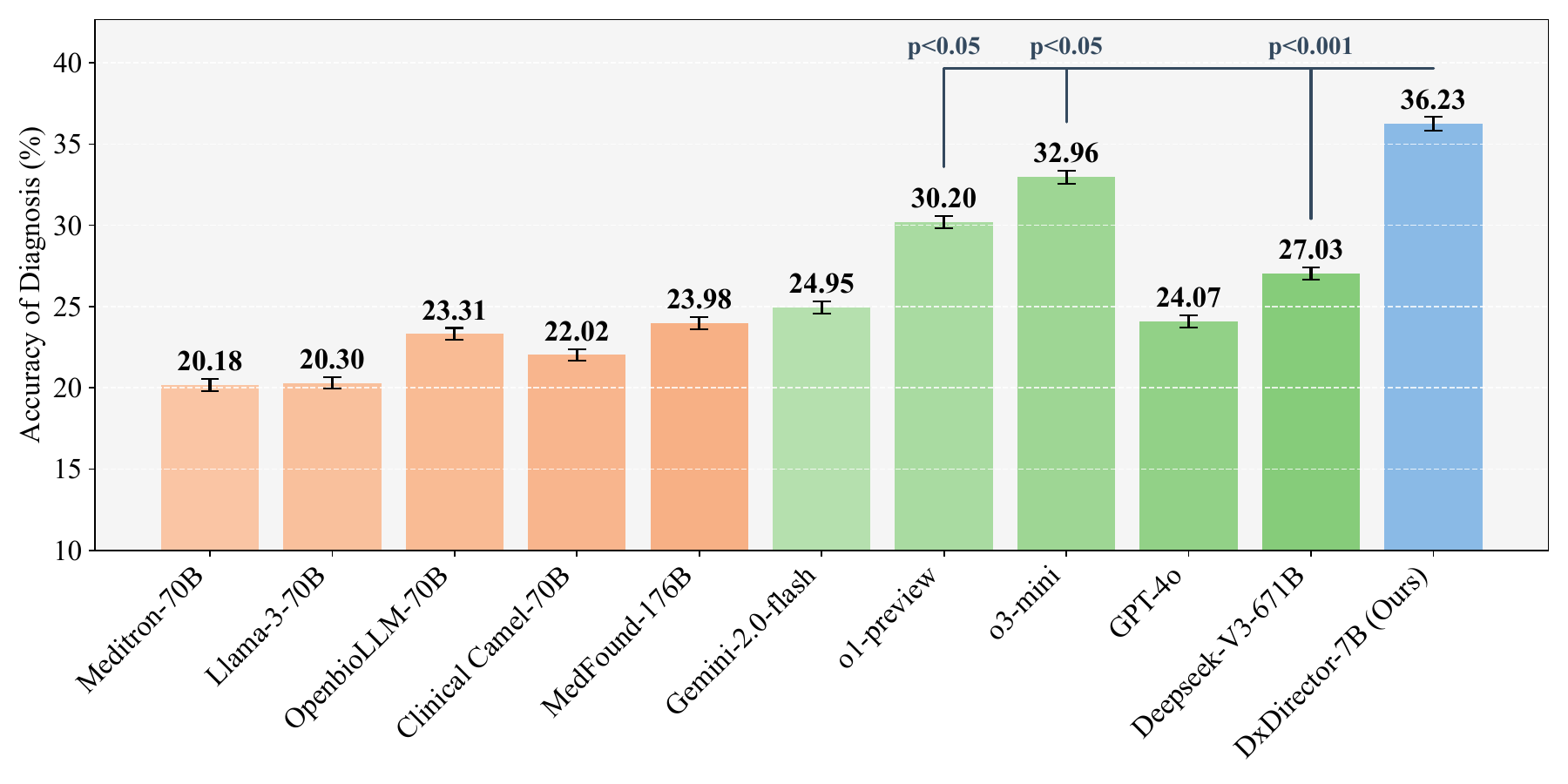}

    \vspace{-1em}
    \caption{Rare Disease Cases (RareArena).}
    \label{rare_acc}
  \end{subfigure}
  
  \vspace{1em}
  \begin{subfigure}[b]{0.95\textwidth} 
    \includegraphics[width=\linewidth]{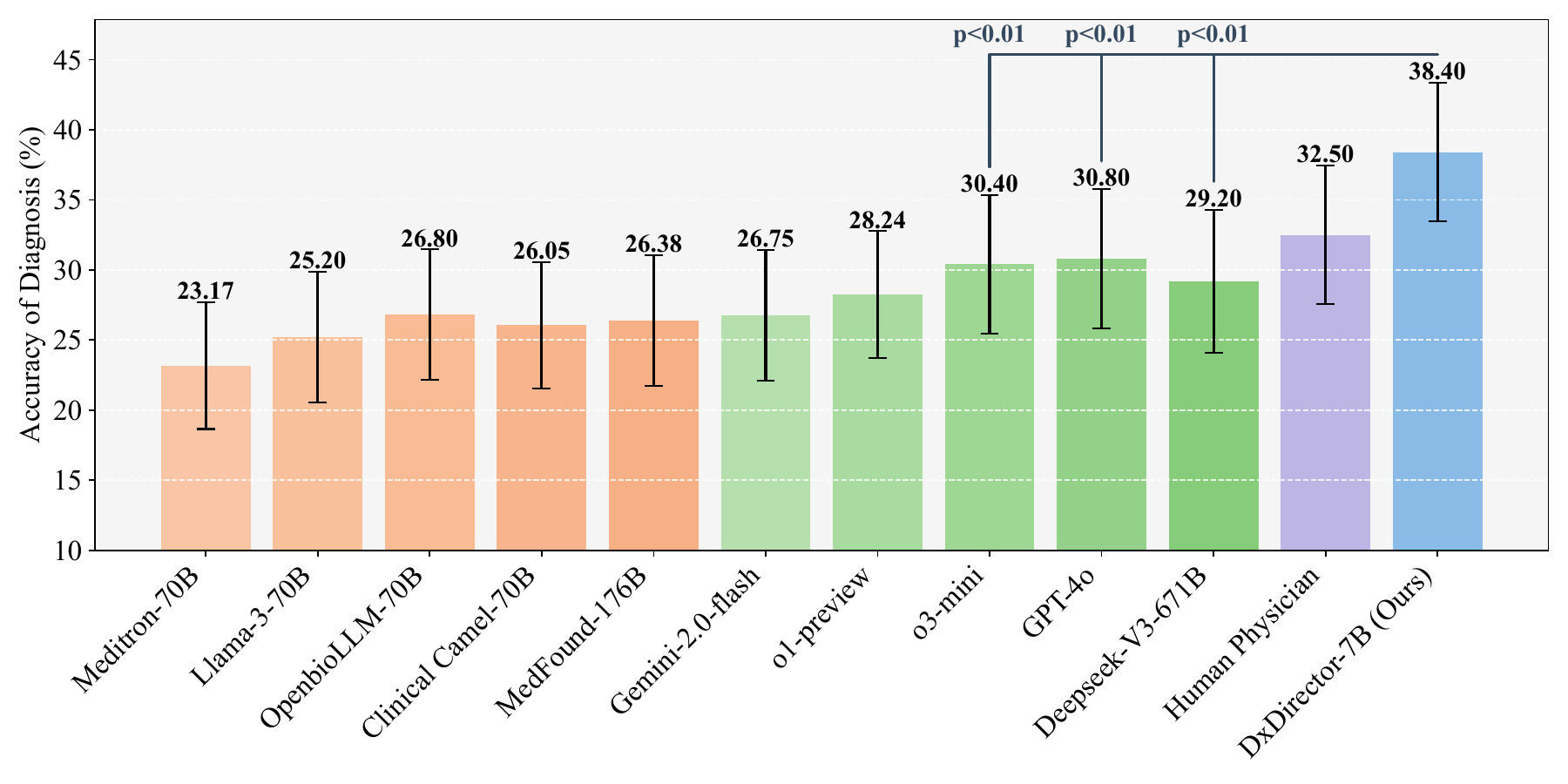}
    
    \vspace{-0.5em}
    \caption{Complex Cases (NEJM Clinicopathologic Cases). Accuracy of human physician is from~\cite{brodeur2024superhuman}.}
    \label{NEJM_acc}
  \end{subfigure}
  
  \vspace{1em}
  \begin{subfigure}[b]{0.95\textwidth}
    \includegraphics[width=\linewidth]{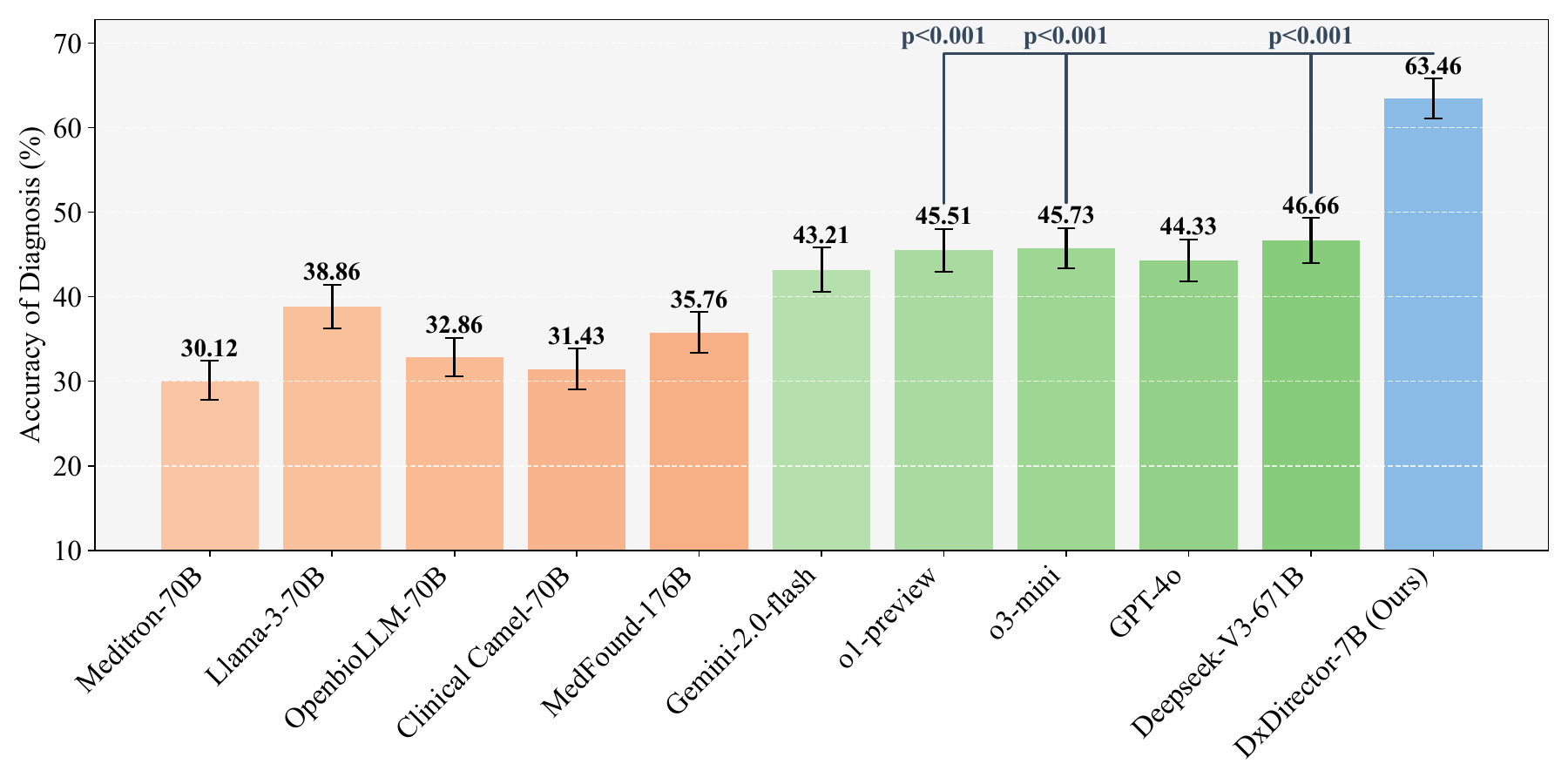}
    \vspace{-2em}
    \caption{Real-world Cases (ClinicalBench).}
    \label{clinical_acc}
  \end{subfigure}
  \caption{Accuracy of diagnoses generated by different LLMs across different datasets in full-process diagnosis setting. Bars are annotated with the accuracy of each LLM. Error bars reflect $95\%$ confidence intervals determined by non-parametric bootstrap procedure with 1,000 samples on RareArena and ClinicalBench, and 200 samples on NEJM Cases. We perform statistical significance tests utilizing the two-side McNemar test between DxDirector-7B and the top-3 baselines on each dataset, with p-value levels annotated on the bars.}
  \label{diag_acc}
\end{figure}

The evaluation on RareArena reveals the capacity of LLMs to diagnose rare diseases—a challenging domain that requires expertise in conditions characterized by low prevalence, encompassing 4,597 distinct pathologies across 22,901 clinical cases. As illustrated in Fig.\ref{rare_acc}, under the full-process diagnostic setting, our DxDirector-7B achieves the highest accuracy at $36.23\%$. This represents a $3.27\%$ absolute advantage over the strongest commercial LLM (o3-mini: $32.96\%$) and a $12.25\%$ lead against medically adapted LLMs (MedFound-176B: $23.98\%$), despite using $25$ times fewer parameters than medically adapted LLMs and nearly $100$ times fewer than commercial LLMs like Deepseek-V3-671B ($27.03\%$). Besides, the stark contrast between commercial LLMs highlights reasoning's critical role—GPT-4o ($24.07\%$) underperforms o3-mini by $8.89\%$ and o1-preview ($30.20\%$) by $6.13\%$, despite comparable medical knowledge memorization. Both o1-preview and o3-min are LLMs with powerful reasoning ability. This suggests that stronger logical reasoning enables better synthesis of sparse symptom patterns in rare disease diagnosis, a capability GPT-4o lacks despite superior general intelligence~\cite{qiu2025quantifyingreasoningabilitiesllms}. Our DxDirector-7B amplifies this advantage through deep thinking like human at each step, achieving higher parameter efficiency while delivering superior accuracy.

The evaluation on NEJM Clinicopathologic Cases benchmark reveals critical insights into the capabilities and limitations of LLMs in complex clinical reasoning. Fig.~\ref{NEJM_acc} presents the diagnostic accuracy of baselines and our DxDirector-7B in the setting of full-process clinical diagnosis. Our DxDirector-7B achieves the best accuracy ($38.4\%$) and outperforms human physicians. Further analysis reveals three pivotal findings. First, existing medical-domain adaptation methods of LLMs provides limited benefits on full-process diagnosis: LLMs pretrained on large-scale medical data (Meditron-70B: $23.17\%$; OpenbioLLM-70B: $26.80\%$; MedFound-176B: $26.38\%$) show marginal gains over the generalist Llama-3-70B ($25.20\%$) at equivalent (70B) or even more (176B) parameters ($\Delta \leq 1.60\%$). Second, while trillion-parameter commercial general-purpose LLMs (GPT-4o: $30.8\%$; Deepseek-V3-671B: $29.2\%$) surpass medically adapted LLMs, all remain statistically inferior to human physicians ($32.5\%$), exposing fundamental limitations in existing LLMs for full-process clinical diagnosis. Third, DxDirector-7B achieves the best accuracy---a $5.9\%$ absolute improvement over physicians and $7.6\%$ over GPT-4o---despite using merely $4\%$--$10\%$ parameters of medically adapted LLMs (7B vs.\ 70B, 176B) and nearly $1\%$ of commercial general-purpose LLMs. This demonstrates the effectiveness and efficiency of our training method in allowing LLMs to think deeply like ``slow thinking'' at each reasoning step, which enables human-surpassing diagnostic accuracy. The results redefine optimization strategies for medical LLMs, proving that lightweight models with powerful deep thinking ability can master complex full-process clinical reasoning, rather than brute-force scaling or narrow pretraining on large scale medical data.

The ClinicalBench—spanning $1,500$ real-world cases across $150$ diseases—reveals the performance of LLMs in real-world full-process clinical diagnosis. Results in Fig.\ref{clinical_acc} show DxDirector-7B achieves the highest accuracy at $63.46\%$, outperforming the strongest commercial model (Deepseek-V3-671B: $46.66\%$) by $16.8\%$ and medically adapted LLMs (Clinical Camel-70B: $31.43\%$; OpenbioLLM-70B: $32.86\%$; MedFound-176B: $35.76\%$) by $27.70\%$ to $32.03\%$, despite using much fewer parameters. In addition to the conclusions consistent with NEJM and RareArena, the results on ClinicalBench suggest two important findings: first, compared to NEJM and RareArena, our DxDirector-7B achieves the largest absolute improvement, which shows the significant advantages of our DxDirector-7B in real-world diagnosis. Second, sole medical-domain training of LLMs cannot be efficiently translated to full-process diagnosis in clinical practice, as OpenbioLLM-70B ($32.86\%$) and MedFound-176B ($35.76\%$) shows worse performance than general LLM Llama-3-70B ($38.86\%$). This indicates ``slow thinking'' plays a more important role than sole medical adaptation in driving full-process diagnosis.

\vspace{-0.5em}
\subsection{Quantitative Analysis of Human Physicians' Workload} \label{sec:work}

\vspace{-0.5em}
This section analyzes the workload needed to by paid by human physicians when LLMs drive the full-process clinical diagnostic. In our constructed AI-driven full-process clinical diagnosis setting, to maximize the potential of LLMs and reduce the workload of human physicians as much as possible, human physicians only need to work as assistants to follow the instructions of LLMs to complete clinical operations that LLMs cannot achieve, such as observing symptoms, physical examinations, laboratory tests, and so on. An ideal LLM should be capable of precisely identifying the essential clinical tasks that truly require human intervention, adhering to the principle of minimizing physician involvement while ensuring diagnostic accuracy. To quantitatively assess this, we introduce two key metrics: (1) the total number of clinical operations that the LLM requests physicians to perform throughout the diagnostic process (where fewer requests indicate greater efficiency), and (2) the proportion of operations that are truly useful for making an accurate diagnosis out of all requested operations. (the higher the better). The specific clinical operations required by LLMs can be found in word cloud analysis of Supplementary Fig. 1 to Fig. 10.

Combing the findings in Section~\ref{sec:acc} and~\ref{sec:work}, in the full-process diagnostic setting, our DxDirector-7B not only achieves significant superior diagnostic accuracy but also substantially reduces physician workload than state-of-the-art LLMs, indicating the efficient, accurate and scalable diagnostic solution. The detailed results and analysis are as follows.

\begin{figure}[H]
  \centering
    
  \begin{subfigure}[b]{0.9\textwidth}
    \includegraphics[width=\linewidth]{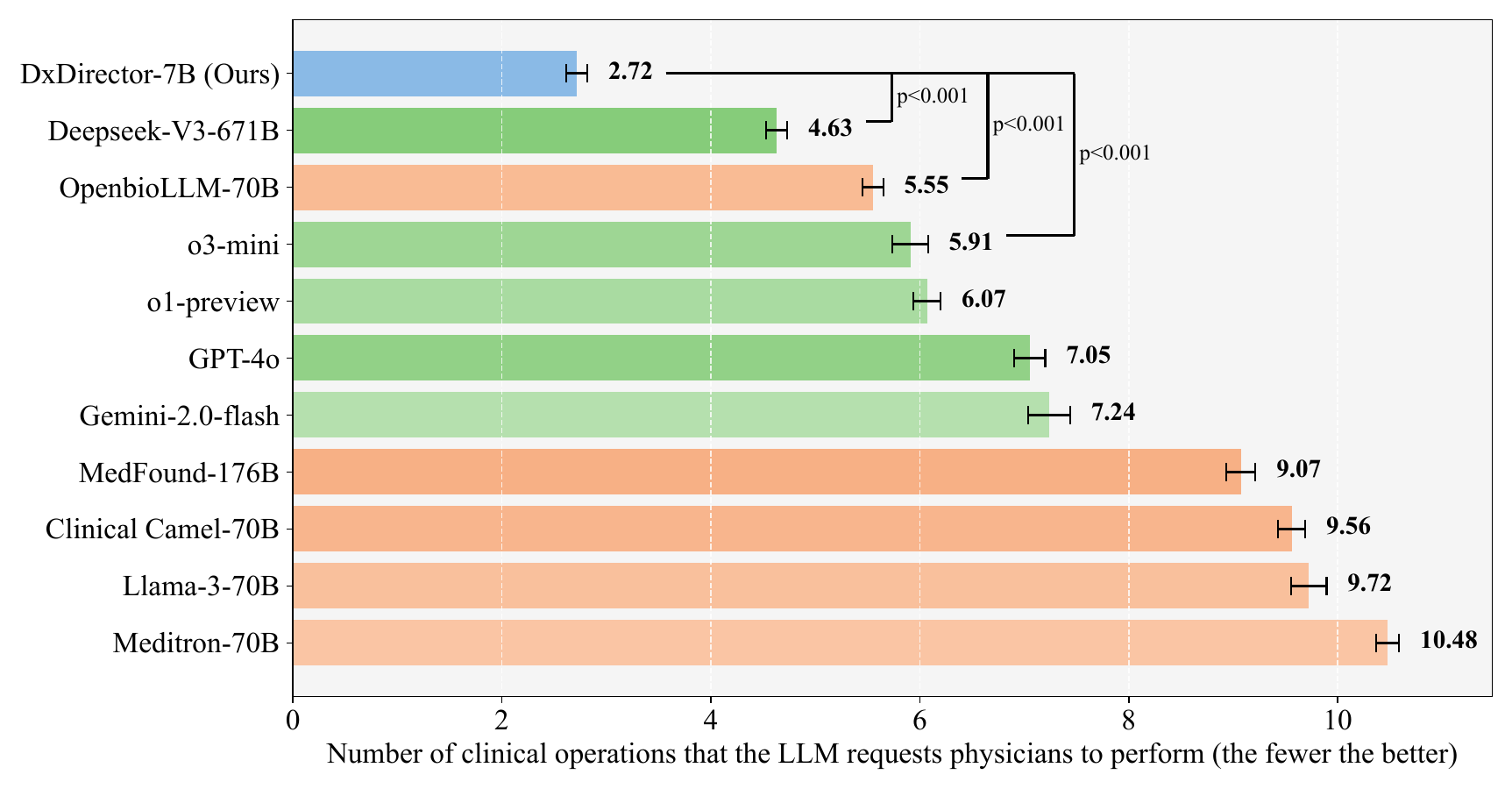}
    
    \vspace{-0.5em}
    \caption{Rare Disease Cases (RareArena).}
    \label{rare_count}
  \end{subfigure}
  
\vspace{0.5em}
  \begin{subfigure}[b]{0.9\textwidth} 
    \includegraphics[width=\linewidth]{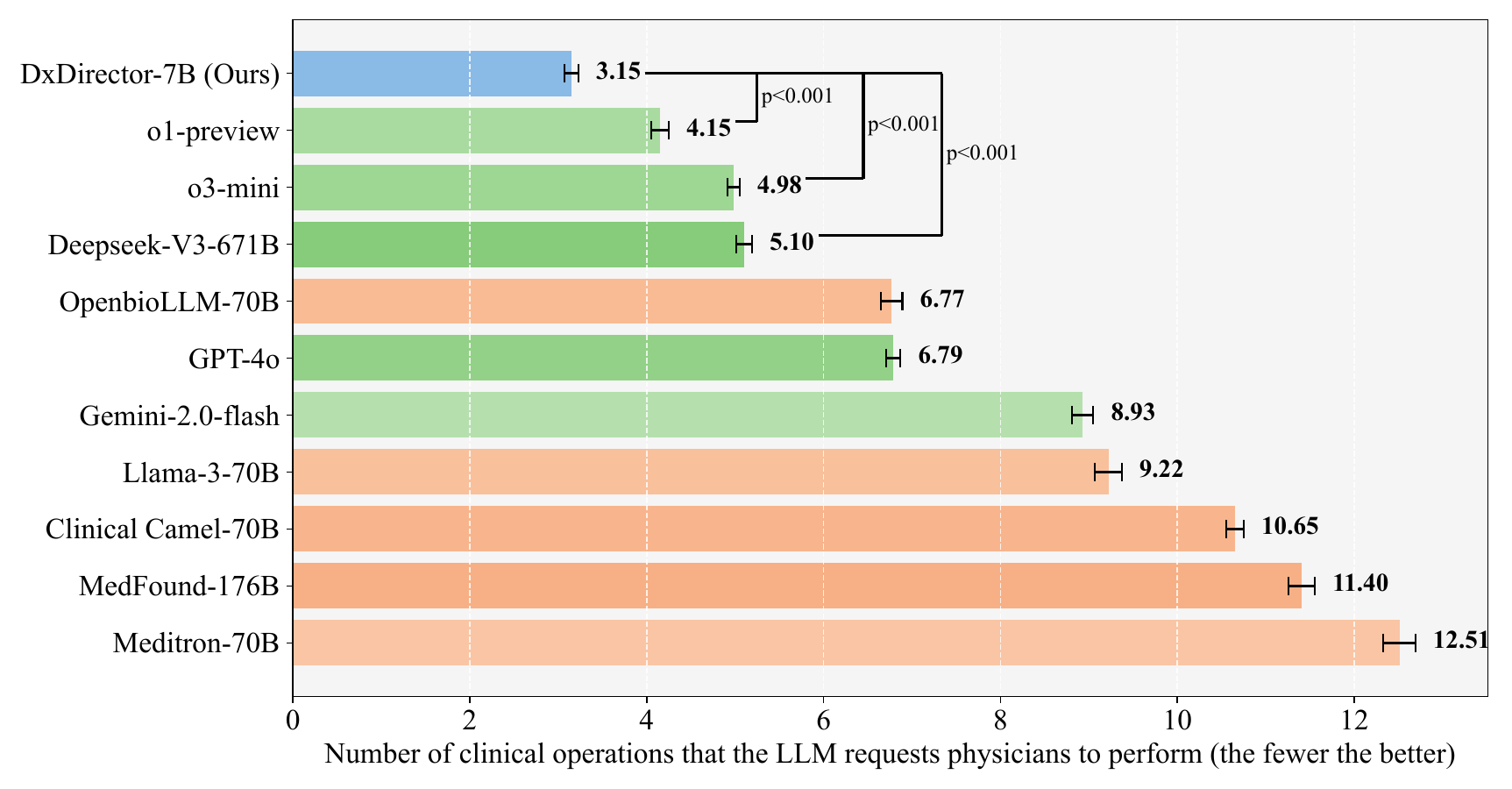}
        
    \vspace{-0.5em}
    \caption{Complex Cases (NEJM Clinicopathologic Cases).}
    \label{NEJM_count}
  \end{subfigure}

\vspace{0.5em}
  \begin{subfigure}[b]{0.9\textwidth}
    \includegraphics[width=\linewidth]{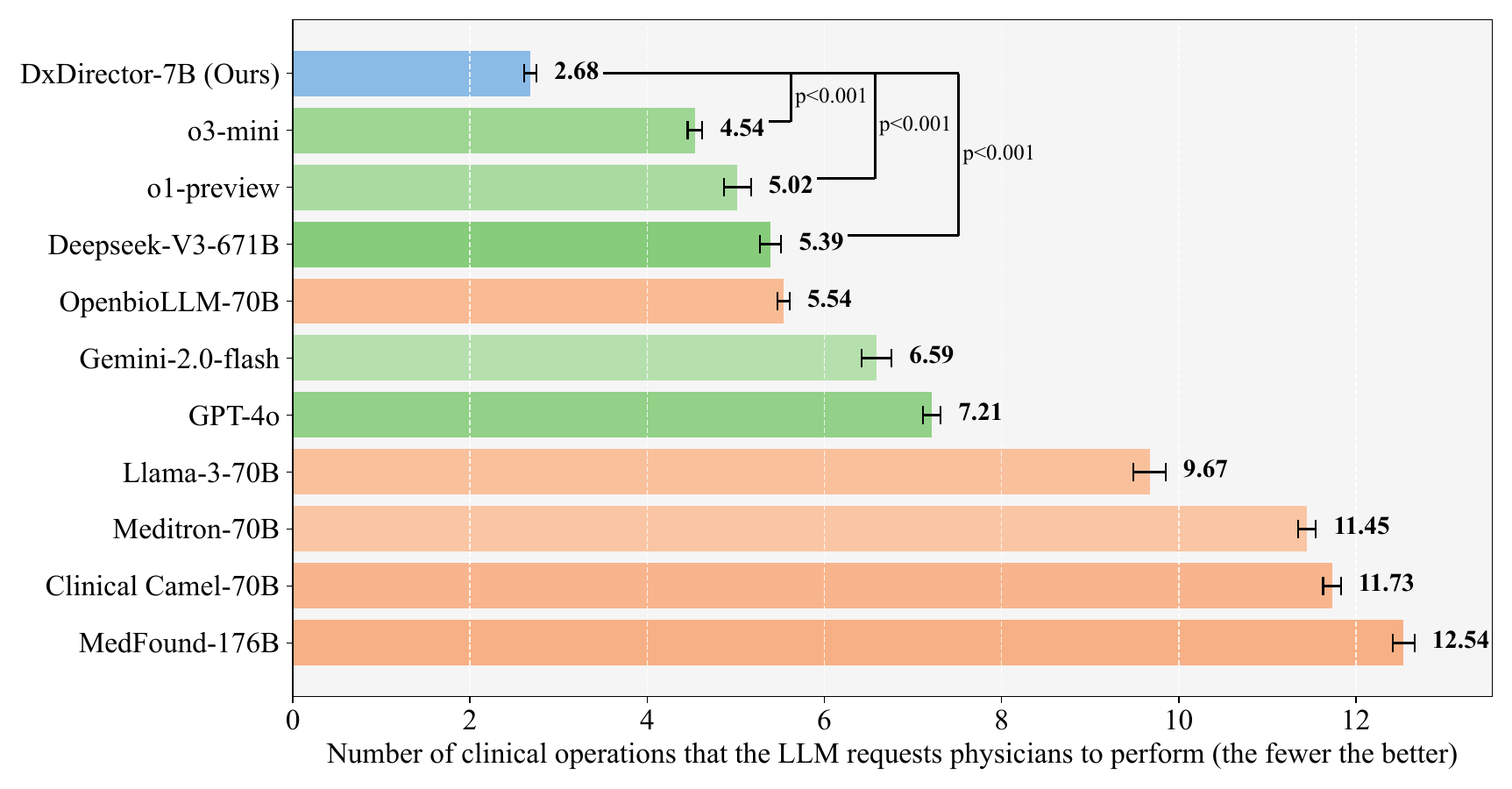}
            
    \vspace{-0.5em}
    \caption{Real-world Cases (ClinicalBench).}
    \label{clinical_count}
  \end{subfigure}
  
  \caption{Number of clinical operations that LLMs request physicians to perform in the entire diagnosis process (the fewer the better). Error bars reflect $95\%$ confidence intervals determined by non-parametric bootstrap procedure with 1,000 samples on RareArena and ClinicalBench, and 200 samples on NEJM Cases. We perform statistical significance tests utilizing two-side Mann-Whitney U test between DxDirector-7B and the top-3 baselines, with p-value
levels annotated on the bars.}
  \label{fig:count}
\end{figure}

For the first metric, the average statistical results over three datasets are shown in Fig.~\ref{rare_count},~\ref{NEJM_count} and~\ref{clinical_count}. Specifically, DxDirector-7B effectively complete the entire diagnostic process with an average of approximately $3$ clinical operations across diverse diagnostic scenarios, including rare diseases (RareArena), complex cases (NEJM), and real-world clinical contexts (ClinicalBench). This efficiency notably surpass that of all baseline LLMs. By comparison, general-purpose commercial LLMs typically necessitate between 4 and 8 operations, while open-source medically adapted LLMs exhibit the poorest performance, often requiring nearly 10 clinical operations. Within baseline comparisons, commercial general-purpose LLMs such as o3-mini and o1-preview, benefiting from robust reasoning capabilities, consistently require fewer operations than other LLMs, including GPT-4o. Enhanced reasoning capacity allows these LLMs to effectively leverage available clinical data to make accurate clinical decision, thus minimizing additional operational demands—an attribute particularly exemplified by our DxDirector-7B that can perform ``slow thinking'' like human before making the specific strategy at each diagnostic step.

\begin{figure}[H]
  \centering
  \begin{subfigure}[b]{0.9\textwidth}
    \includegraphics[width=\linewidth]{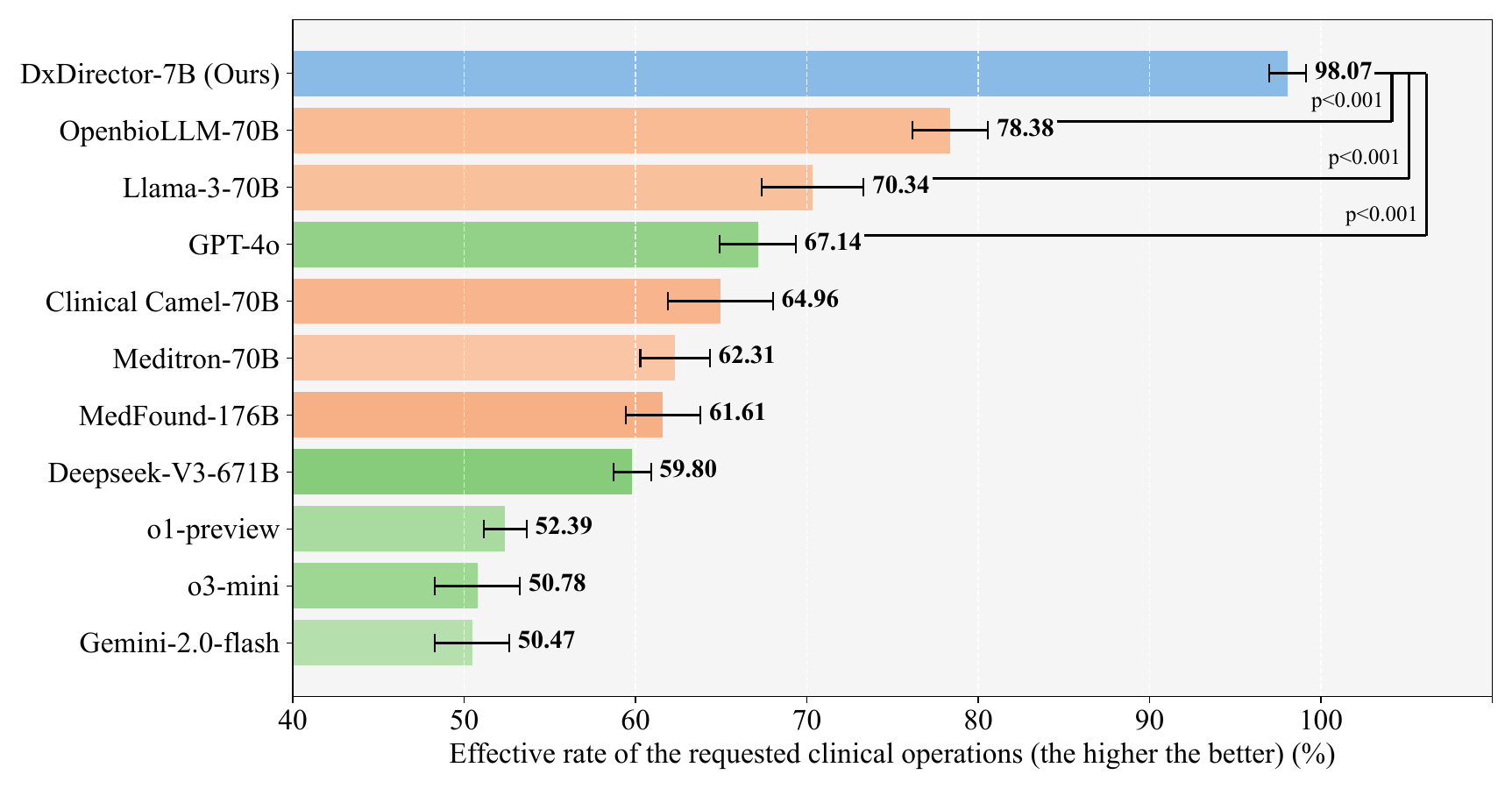}
            
    \vspace{-0.5em}
    \caption{Rare Disease Cases (RareArena).}
    \label{rare_rate}
  \end{subfigure}
  
    \vspace{0.5em}
  \begin{subfigure}[b]{0.9\textwidth} 
    \includegraphics[width=\linewidth]{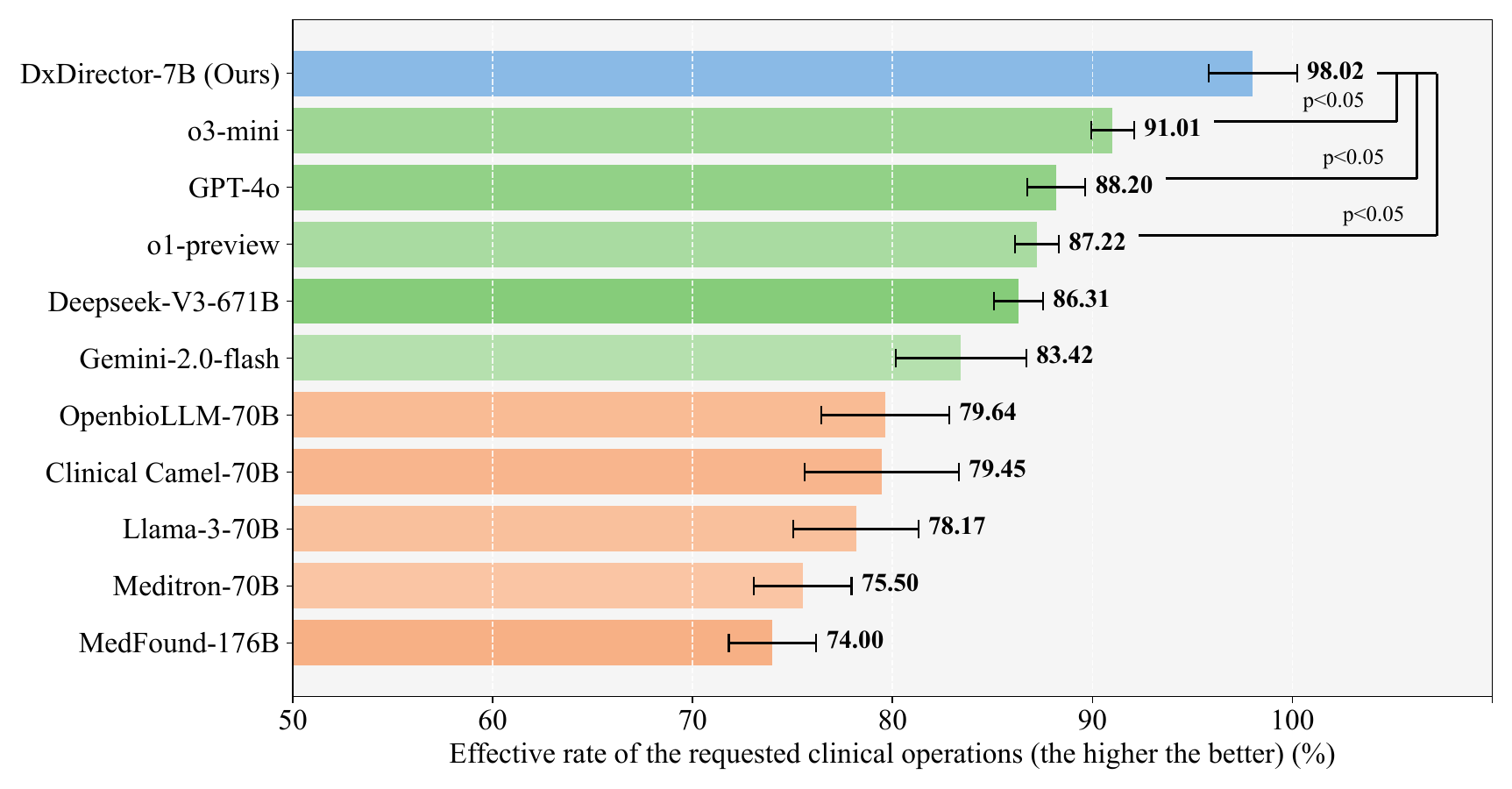}
            
    \vspace{-0.5em}
    \caption{Complex Cases (NEJM Clinicopathologic Cases).}
    \label{NEJM_rate}
  \end{subfigure}

    \vspace{0.5em}
  \begin{subfigure}[b]{0.9\textwidth}
    \includegraphics[width=\linewidth]{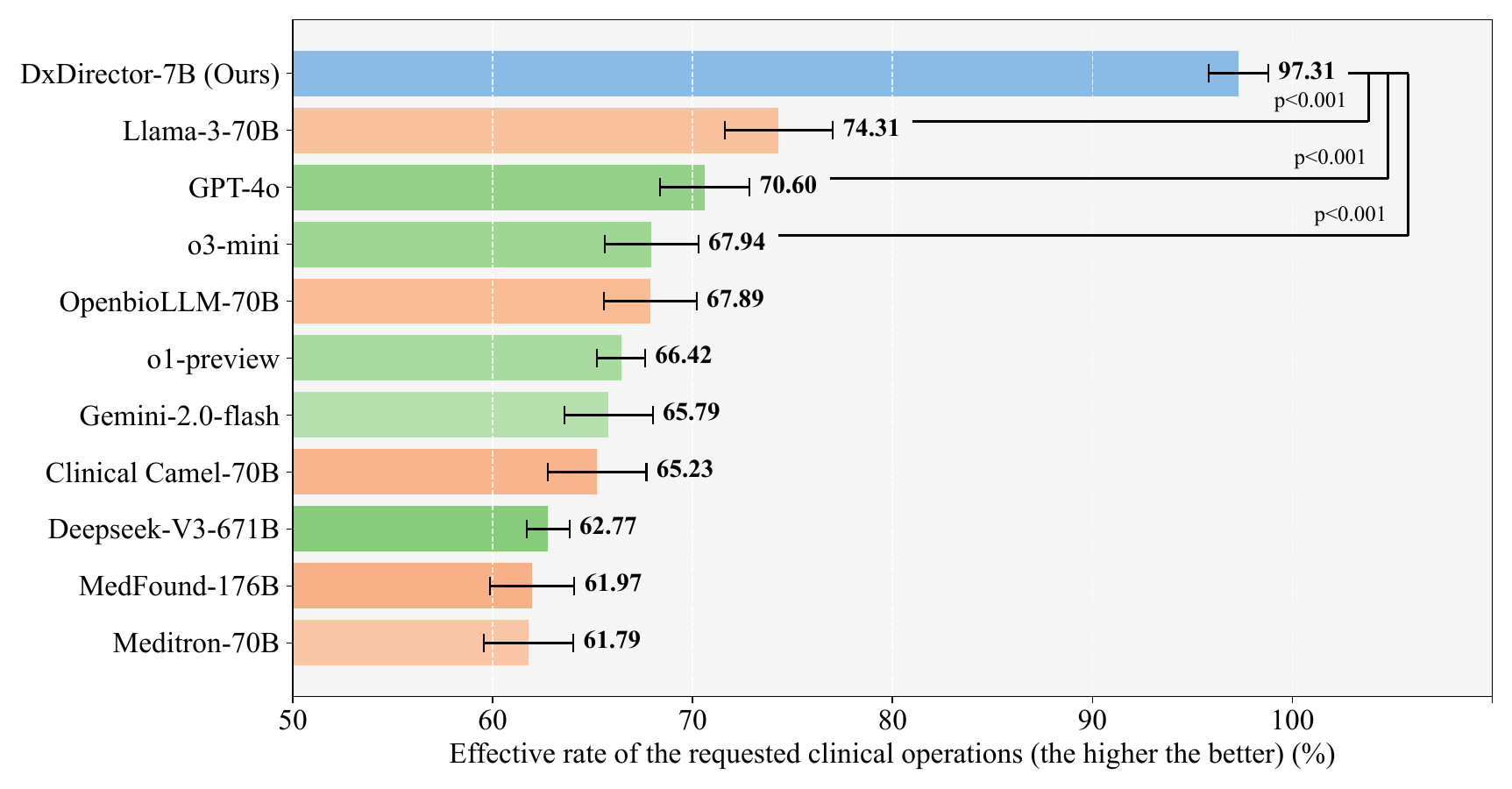}
                
    \vspace{-0.5em}
    \caption{Real-world Cases (ClinicalBench).}
    \label{clinical_rate}
  \end{subfigure}
  
  \caption{Proportion of operations that are truly useful for making a diagnosis out of all requested operations. (the higher the better). Error bars reflect $95\%$ confidence intervals determined by non-parametric bootstrap procedure with 1,000 samples on RareArena and ClinicalBench, and 200 samples on NEJM Cases. We perform statistical significance tests utilizing
two-side Mann-Whitney U test between DxDirector-7B and the top-3 baselines, with p-value
levels annotated on the bars.}
  \label{fig:rate}
\end{figure}

For the second metric, we determine whether an operation genuinely contributes to diagnosis by assessing whether it appears in the case report provided by medical specialists. This metric serves as an indicator of the LLMs' proficiency in discerning essential operations necessary for accurate diagnosis while avoiding any redundant operations, with higher values reflecting greater efficiency in engaging human physician assistance.  Experimental results across three datasets are presented in Fig.~\ref{rare_rate}, \ref{NEJM_rate}, and \ref{clinical_rate}. DxDirector-7B demonstrates consistently superior performance, achieving efficiency ranging from $97\%$ to $98\%$ across all datasets, significantly surpassing the baselines. The performance of the baselines varies. In the diagnosis of complex cases (NEJM), general-purpose LLMs consistently outperform medically adapted LLMs, whereas in the diagnosis of rare diseases (RareArena), medically adapted LLMs surpass general-purpose LLMs. This indicates that the improvement on efficiency in seeking human physicians’ assistance is jointly driven by reasoning capabilities and the retention of long-tail medical knowledge. General-purpose commercial LLMs excel in the former, while medically adapted LLMs excel in the latter.

\subsection{Department-level Fine-grained Evaluations}
In this part, we categorize the data from ClinicalBench and RareArena by clinical department and assess the diagnostic accuracy within each category, providing a more granular evaluation of LLMs.
\begin{figure}[H]
    \centering
        \includegraphics[width=1.0\linewidth]{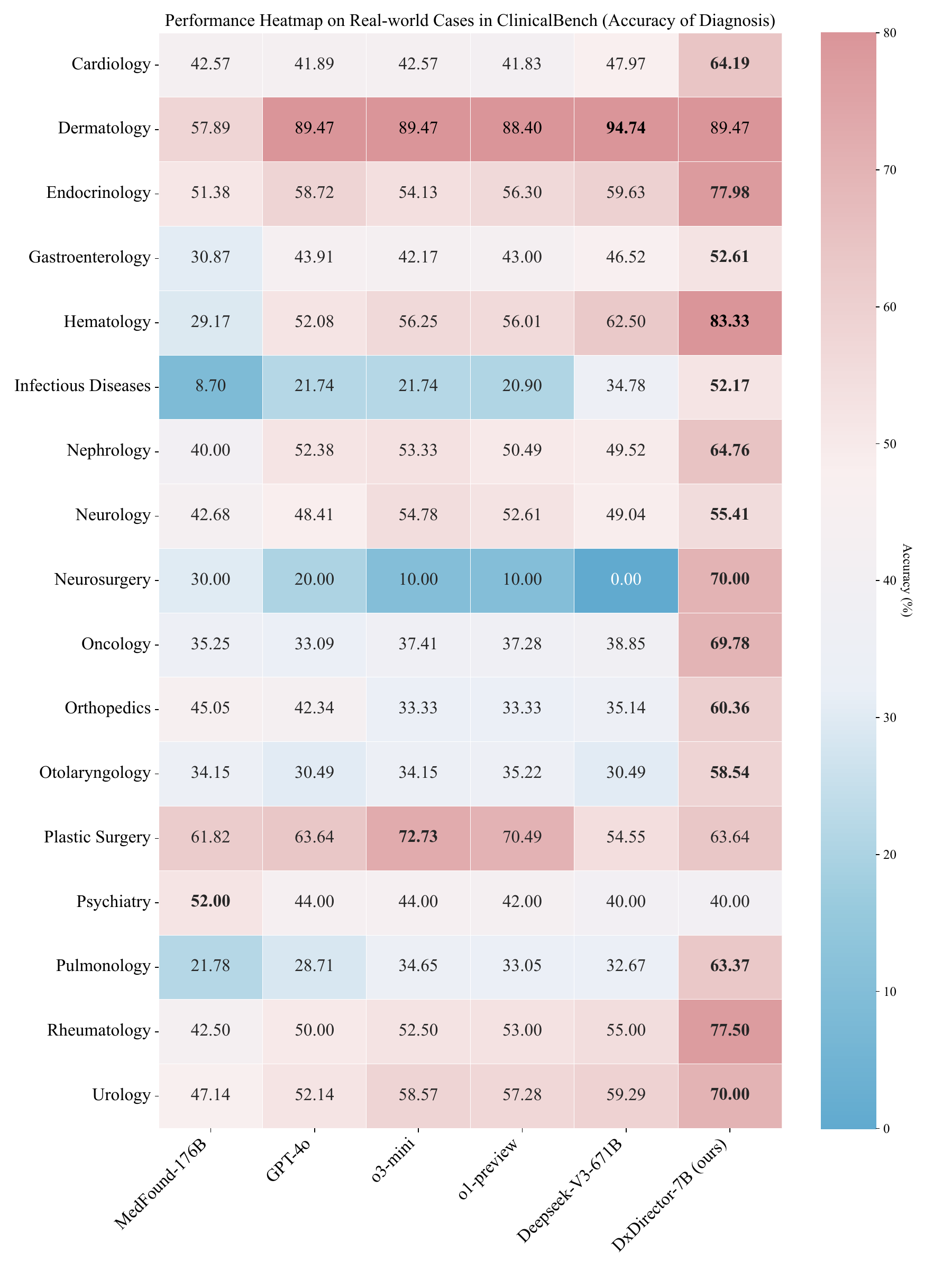}
        \caption{A comparative heatmap analysis of diagnostic accuracy: DxDirector-7B vs. state-of-the-art medically adapted and commercial general-purpose LLMs across 17 clinical departments consisting of 1,500 samples in ClinicalBench that is collected from real world. \textbf{Bold} indicates the best performance.}
        \label{clinical_heatmap}
\end{figure}
The heat map in Fig.~\ref{clinical_heatmap} illustrates the diagnostic accuracy across 17 clinical departments comprising 1,500 real-world cases within ClinicalBench. Our DxDirector-7B achieves the best performance on 14 out of 17 departments. ClinicalBench can reflect the true clinical distribution encountered in routine practice. DxDirector-7B significantly outperforms all baseline LLMs, with substantial margins observed particularly in Neurosurgery ($\Delta=40.0\%$), Oncology ($\Delta=30.93\%$) and Pulmonology
\begin{figure}[H]
    \centering
        \includegraphics[width=1.0\linewidth]{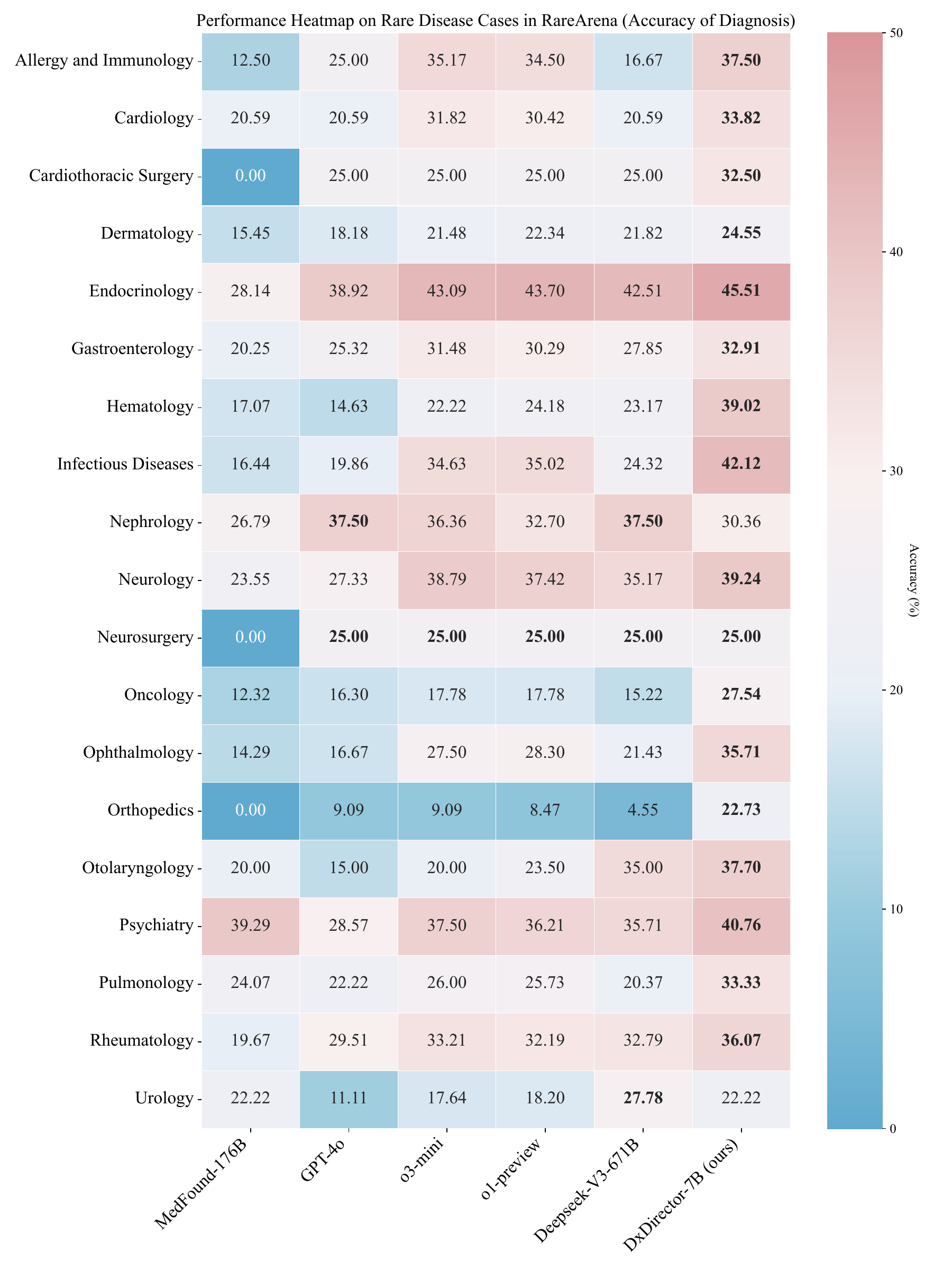}
        \caption{A comparative heatmap analysis of diagnostic accuracy: DxDirector-7B vs. state-of-the-art medically adapted and commercial general-purpose LLMs across 19 clinical departments in rare disease cases on 22,901 samples in RareArena. \textbf{Bold} indicates the best performance.}
        \label{rare_heatmap}
\end{figure}
($\Delta=28.72\%$). Diagnoses within these departments typically necessitate comprehensive integration of multiple diagnostic tests. It is a challenging scenario for existing state-of-the-art LLMs, which struggle to actively pursue and integrate necessary diagnostic information starting from only vague patient chief complaints. DxDirector-7B cannot achieve the best performance on Dermatology, Plastic Surgery and Psychiatry. It mainly because that the clinical diagnosis of these three departments is extremely dependent on frequent real contact, observation, and interactions between human physicians and patients, which cannot give full play to the advantages of DxDirector-7B.

The heat map in Fig.~\ref{rare_heatmap} illustrates the diagnostic accuracy across 19 clinical departments for rare disease diagnosis,  based on 22,901 samples from RareArena. Notably, our DxDirector-7B, outperforms all baselines in 16 out of 19 departments. In particular, DxDirector-7B demonstrates substantial improvements in diagnostic accuracy for rare diseases in Hematology ($\Delta=14.84\%$), Orthopedics ($\Delta=13.64\%$), Oncology ($\Delta=9.76\%$), Pulmonology ($\Delta=7.33\%$), and Infectious Diseases ($\Delta=7.1\%$). The diagnosis of rare diseases in these departments emphasizes that LLMs can accurately plan and integrate diagnostic tests at multiple stages, integrate travel and contact history with laboratory results, and perform image-test collaborative reasoning. These capabilities are the core of LLMs in driving full-process clinical diagnosis, demonstrating the superiority of our DxDirector-7B in this regard. DxDirector-7B cannot surpass Deepseek-v3-671B on Urology ($\Delta=-5.56\%$) and Nephrology ($\Delta=-7.14\%$). Given the overlap in medical knowledge related to the urinary system and kidney function between these two departments, this limitation suggests that DxDirector-7B may have gaps in its long-tail medical knowledge concerning rare diseases in these domains. This analysis shows the strengths and limitations of our DxDirector-7B compared to state-of-the-art LLMs in clinical diagnosis across various departments.

\vspace{-1.0em}
\subsection{Evaluations on Real-world Clinical Diagnosis} \label{beiyi}
In this section, we introduce medical specialists to participate in the evaluation of LLMs in real-world clinical diagnosis scenarios. The real clinical diagnostic scenario is set within an officially certified Grade 3A hospitals in China. The involved patients are inpatients presenting with more complex conditions than typical outpatients. Consequently, LLMs must engage in intricate reasoning to gather comprehensive clinical information effectively. To safeguard patients from potential harm, the evaluation environment is structured as follows: patient behaviors and medical specialist operations during clinical diagnosis are fully recorded using actual inpatient records. Subsequently, two GPT-4o-based agents replicate precisely the recorded behaviors of patients and specialists throughout the diagnostic process. In evaluation, LLMs interact with these agents to drive the full-process diagnosis, initiating solely from the patient's vague chief complaint. Within this controlled environment, LLMs do not directly interact with real patients, and their diagnostic outputs undergo rigorous review by medical specialists, thereby effectively mitigating ethical risks and potential harm. 

This evaluation is performed on 160 cases across 9 different clinical departments including Gastroenterology, Nephrology, Dermatology, Cardiovascular Medicine, Infectious Diseases, Endocrinology, Pulmonology, General Surgery, and Pain Management. We compare our DxDirector-7B with the most powerful commercial LLMs including GPT-4o, o1-preview, o3-mini and Deepseek-V3-671B, which possess tens of times more parameters than our DxDirector-7B. Medical specialists from each department participate in evaluating the diagnostic contents produced by these LLMs. This evaluation is conducted from two aspects: (1) scoring the diagnostic content generated by LLMs (on a scale from 0 to 10), and (2) assessing whether the diagnoses generated by LLMs could fully replace those made by medical specialists. To ensure objective assessments and mitigate potential biases in human specialists scoring, a double-blind adjudication approach is implemented. In this approach, both human specialists and LLMs independently diagnose the same patient cases without exposure to each other's diagnostic outputs. Additionally, a third-party evaluation agent, utilizing both GPT-4o and Deepseek-V3, assigns scores based on the alignment between LLM-generated diagnoses and those provided by medical specialists. The final score is calculated as the average of the scores given by GPT-4o and Deepseek-V3, thus ensuring robust and unbiased comparative assessment. The assessment of whether the diagnoses generated by LLMs could replace those made by specialists also follows the same pattern by observing the decisions of the third party agent (can or cannot).

\begin{figure}[H]
  \centering
  
  \begin{minipage}[t]{0.33\textwidth}
    \centering
    \subcaptionbox{Cardiovascular Medicine\label{fig:sub1}}[0.9\textwidth]{
      \includegraphics[width=\textwidth]{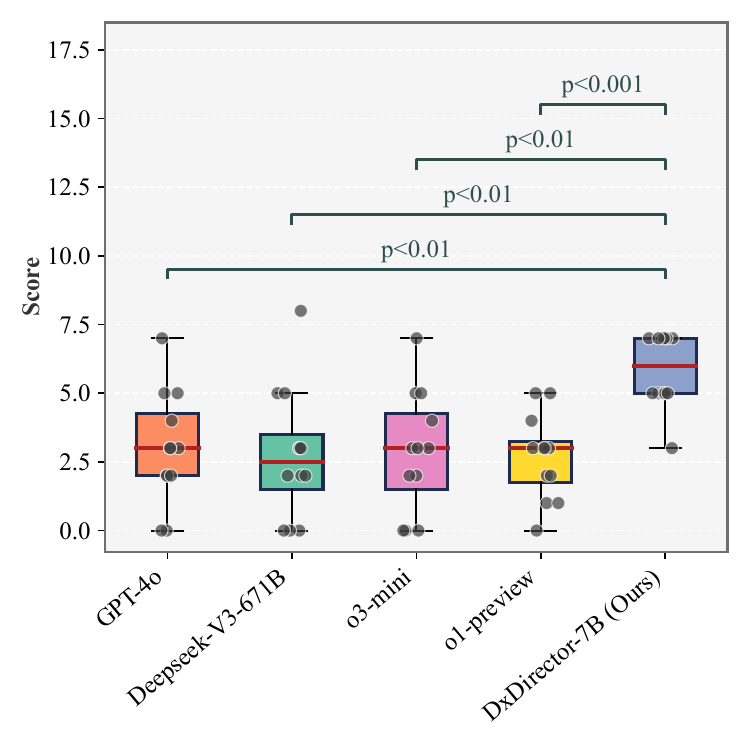}
    }
  \end{minipage}\hfill
  \begin{minipage}[t]{0.33\textwidth}
    \centering
    \subcaptionbox{Dermatology\label{fig:sub2}}[0.9\textwidth]{
      \includegraphics[width=\textwidth]{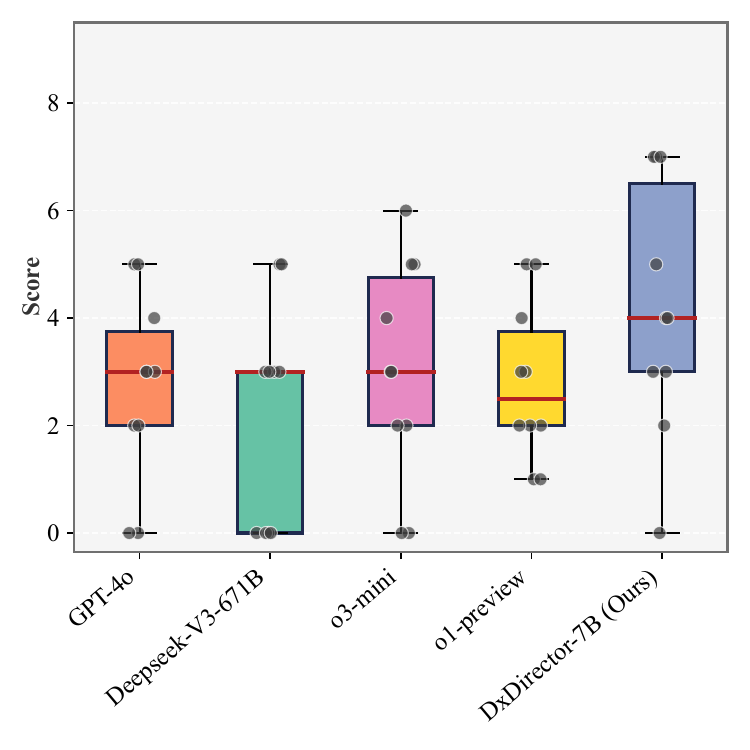}
    }
  \end{minipage}\hfill
  \begin{minipage}[t]{0.33\textwidth}
    \centering
    \subcaptionbox{Endocrinology\label{fig:sub3}}[0.9\textwidth]{
      \includegraphics[width=\textwidth]{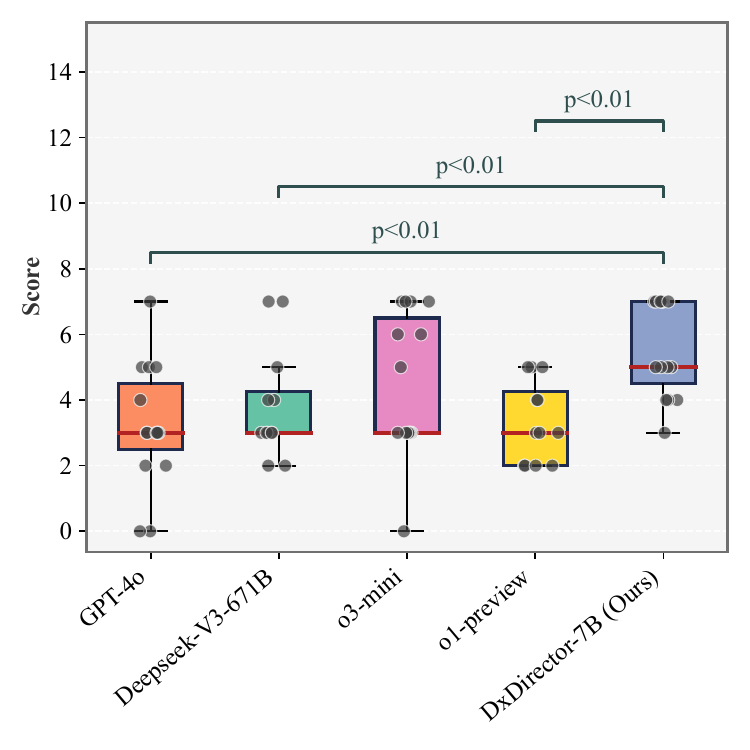}
    }
  \end{minipage}
  
  \begin{minipage}[t]{0.33\textwidth}
    \centering
    \subcaptionbox{Gastroenterology\label{fig:sub4}}[0.9\textwidth]{
      \includegraphics[width=\textwidth]{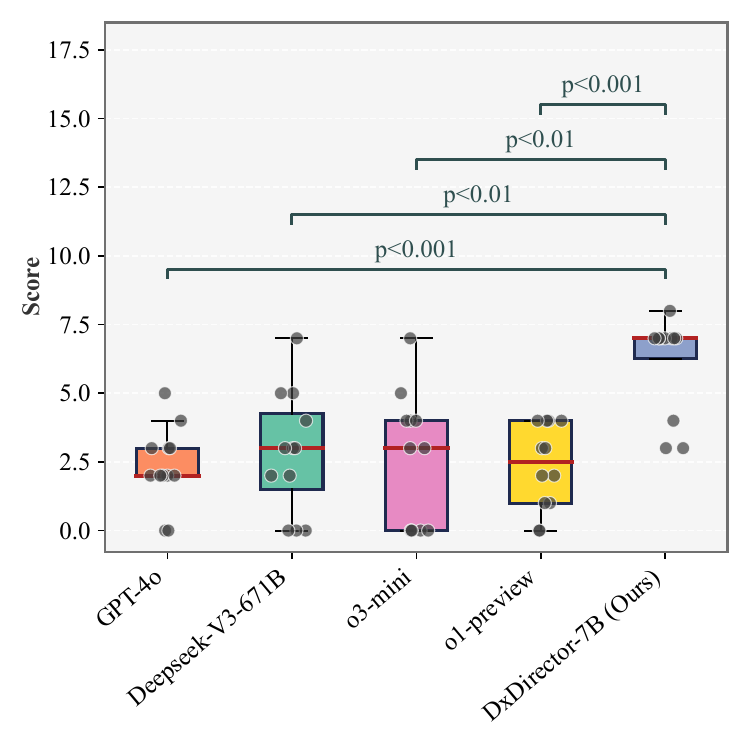}
    }
  \end{minipage}\hfill
  \begin{minipage}[t]{0.33\textwidth}
    \centering
    \subcaptionbox{General Surgery\label{fig:sub5}}[0.9\textwidth]{
      \includegraphics[width=\textwidth]{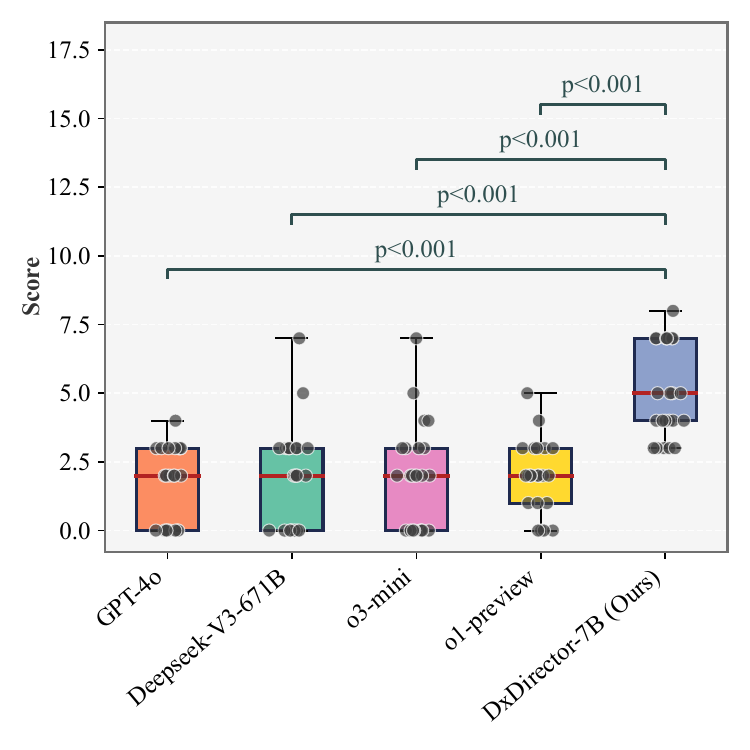}
    }
  \end{minipage}\hfill
  \begin{minipage}[t]{0.33\textwidth}
    \centering
    \subcaptionbox{Infectious Diseases\label{fig:sub6}}[0.9\textwidth]{
      \includegraphics[width=\textwidth]{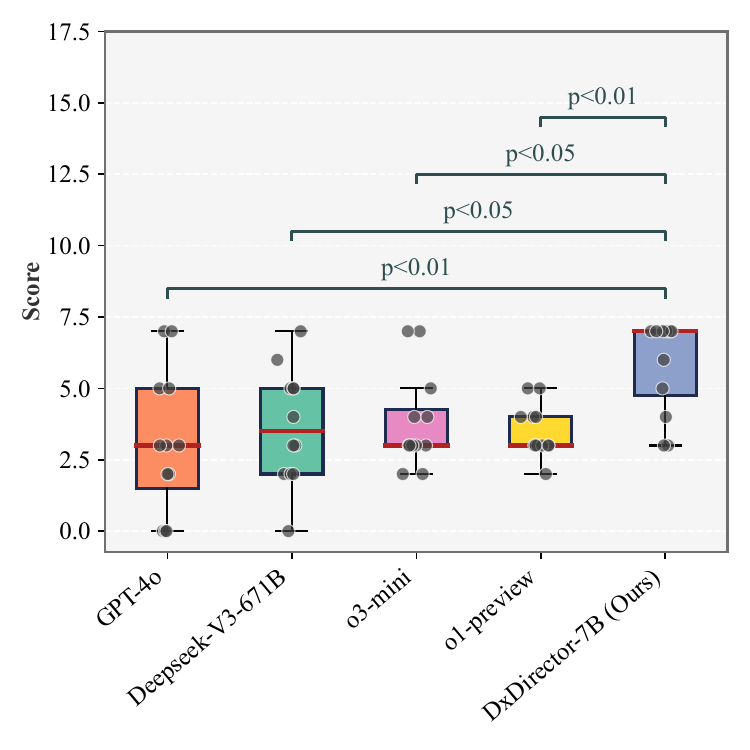}
    }
  \end{minipage}
  
  \begin{minipage}[t]{0.33\textwidth}
    \centering
    \subcaptionbox{Nephrology\label{fig:sub7}}[1.0\textwidth]{
      \includegraphics[width=\textwidth]{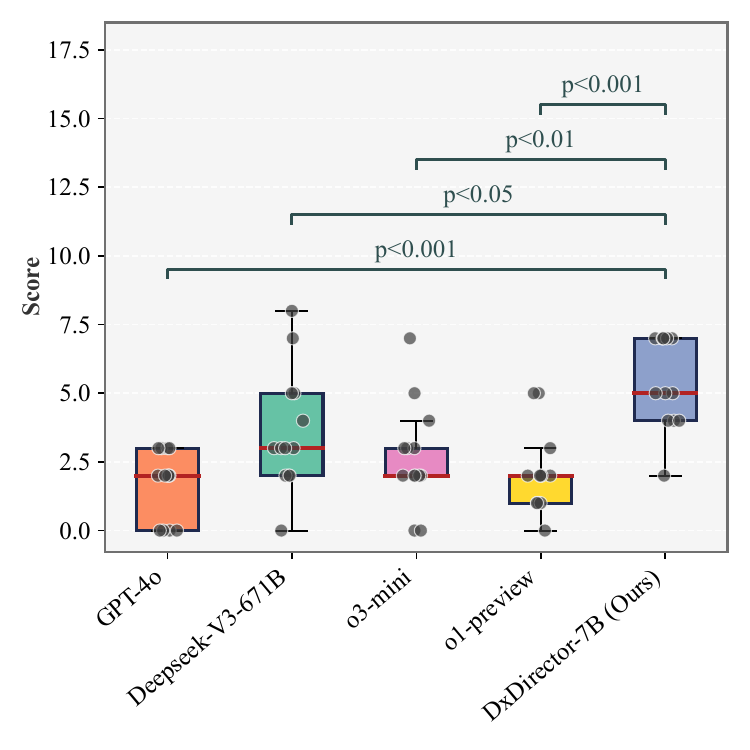}
    }
  \end{minipage}\hfill
  \begin{minipage}[t]{0.33\textwidth}
    \centering
    \subcaptionbox{Pain Management\label{fig:sub8}}[1.0\textwidth]{
      \includegraphics[width=\textwidth]{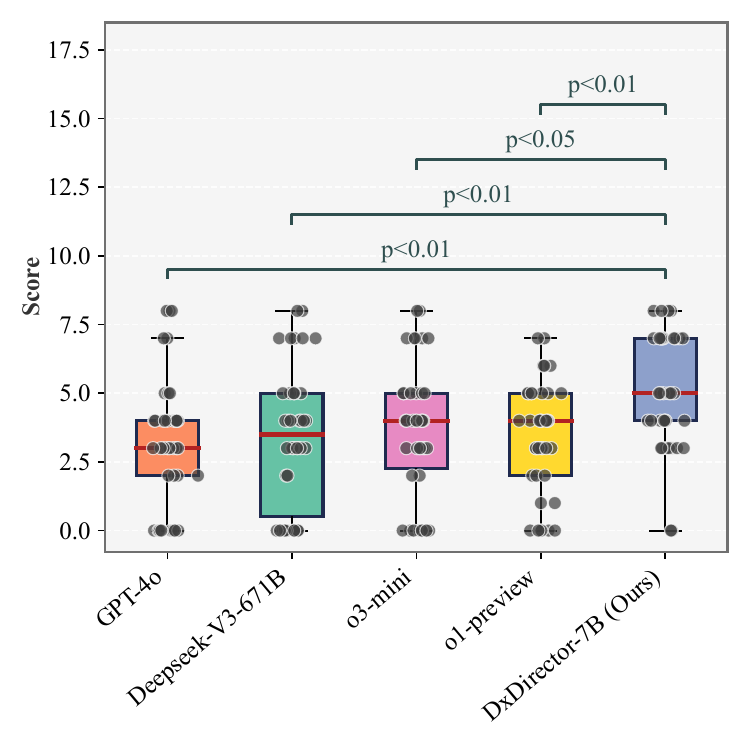}
    }
  \end{minipage}\hfill
  \begin{minipage}[t]{0.33\textwidth}
    \centering
    \subcaptionbox{Pulmonology\label{fig:sub9}}[1.0\textwidth]{
      \includegraphics[width=\textwidth]{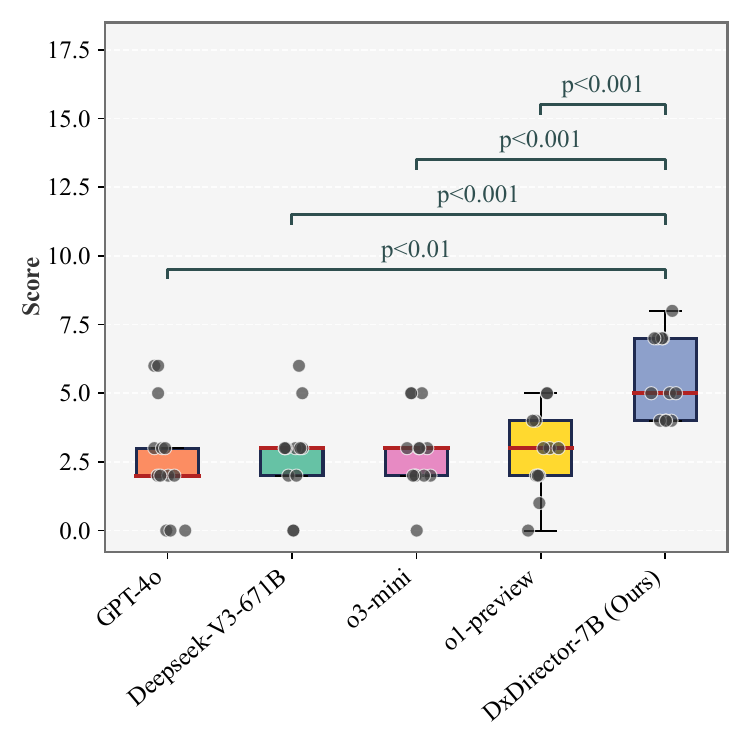}
    }
  \end{minipage}

  \caption{A comparison of the capabilities of different LLMs in 9 departments in real-world clinical diagnosis. The evaluation is conducted by double-blinded adjudication between LLMs and specialists in the corresponding departments, with scores ranging from 0 to 10. We perform statistical significance tests utilizing the two-side Mann-Whitney U test between DxDirector-7B and the baselines, with p-value levels annotated on the figures.}
  \label{fig:beiyi_xiangxian}
\end{figure}

\begin{figure}[H]
    \centering
        \includegraphics[width=0.9\linewidth]{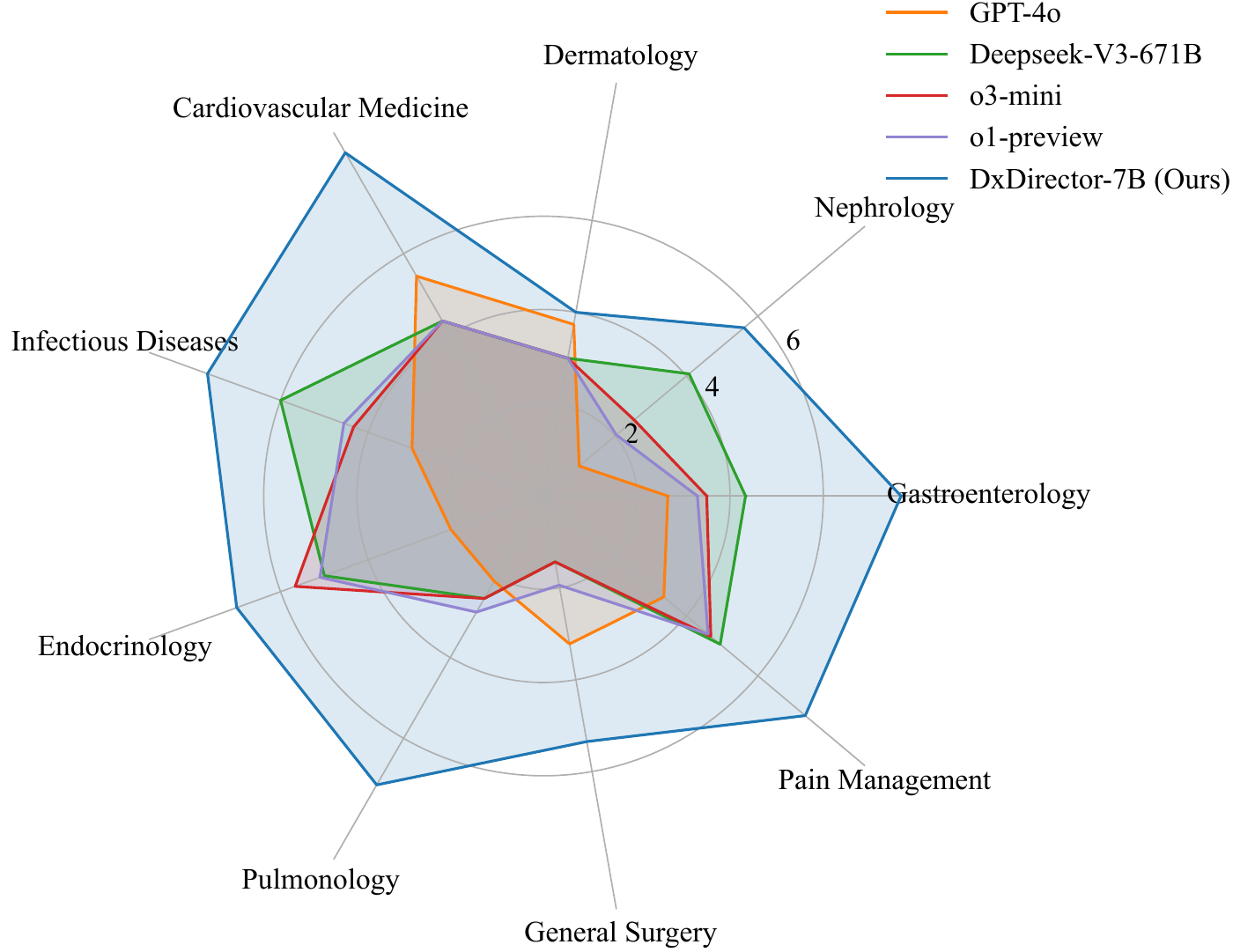}
        \caption{A comparison of the capabilities of different LLMs in 9 departments in real-world clinical diagnosis. The evaluation is conducted by double-blinded adjudication between LLMs and specialists in the corresponding departments, with scores ranging from 0 to 10.}
        \label{beiyi_leida}
\end{figure}

The results of the first aspect are shown in Fig.~\ref{fig:beiyi_xiangxian} and Fig.~\ref{beiyi_leida}. Overall, our DxDirector-7B achieves the highest alignment with medical specialists in all 9 clinical departments, which demonstrates that DxDirector-7B has greater usability and accuracy compared to the most advanced commercial LLMs in real-world clinical practice. The significant lead is evident in Cardiovascular, Pulmonology, and
Gastroenterology. In these departments, patients' chief complaints are far from sufficient to determine the final diagnosis, requiring additional diagnostic test results such as CT scans, angiography, blood tests, and more. These necessitate LLMs to actively acquire the complete clinical information by step-by-step reasoning to finish the entire diagnostic process, posing a substantial challenge for existing LLMs and our DxDirector-7B can effectively address.

\begin{figure}[H]
    \centering
    \subcaptionbox{GPT-4o vs. Specialists\label{fig:sub1}}[0.48\textwidth]
        {\includegraphics[width=0.48\textwidth]{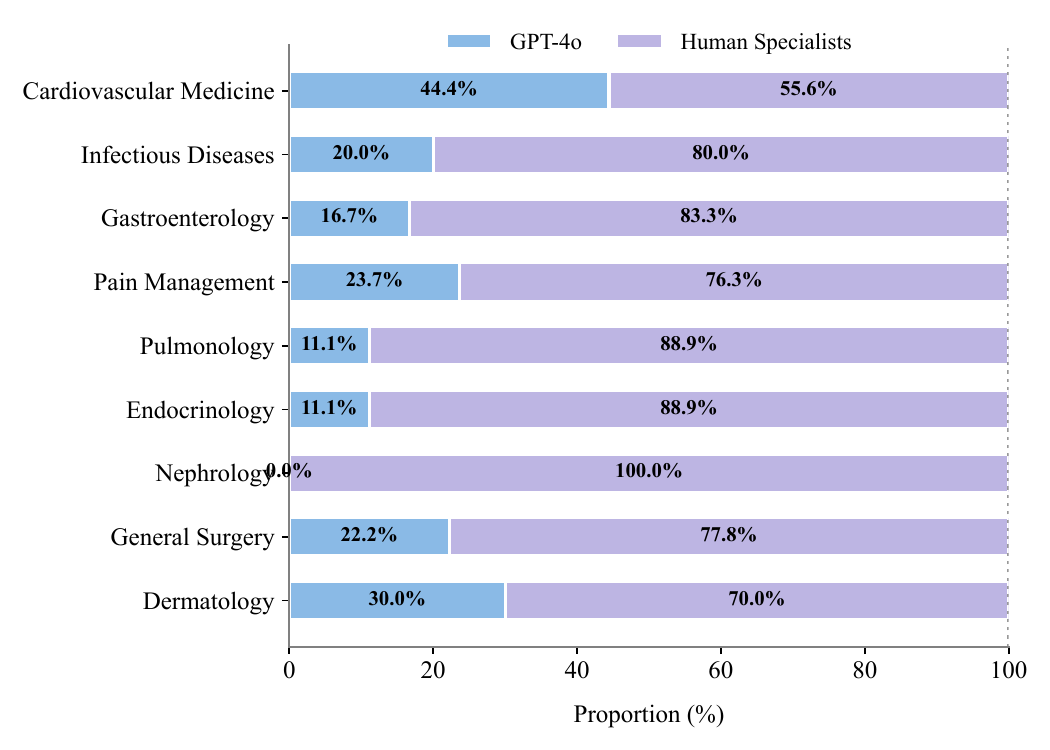}}
    \subcaptionbox{Deepseek-V3-671B  vs. Specialists\label{fig:sub2}}[0.48\textwidth]
        {\includegraphics[width=0.48\textwidth]{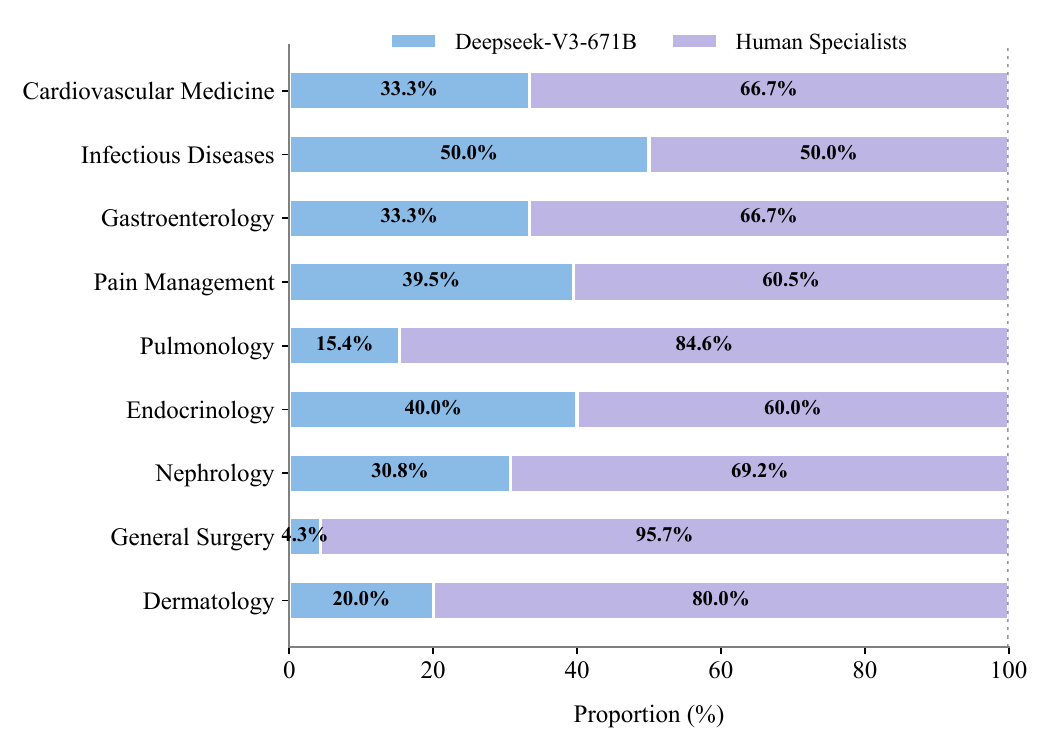}}
    \subcaptionbox{o1-preview vs. Specialists\label{fig:sub3}}[0.48\textwidth]
        {\includegraphics[width=0.48\textwidth]{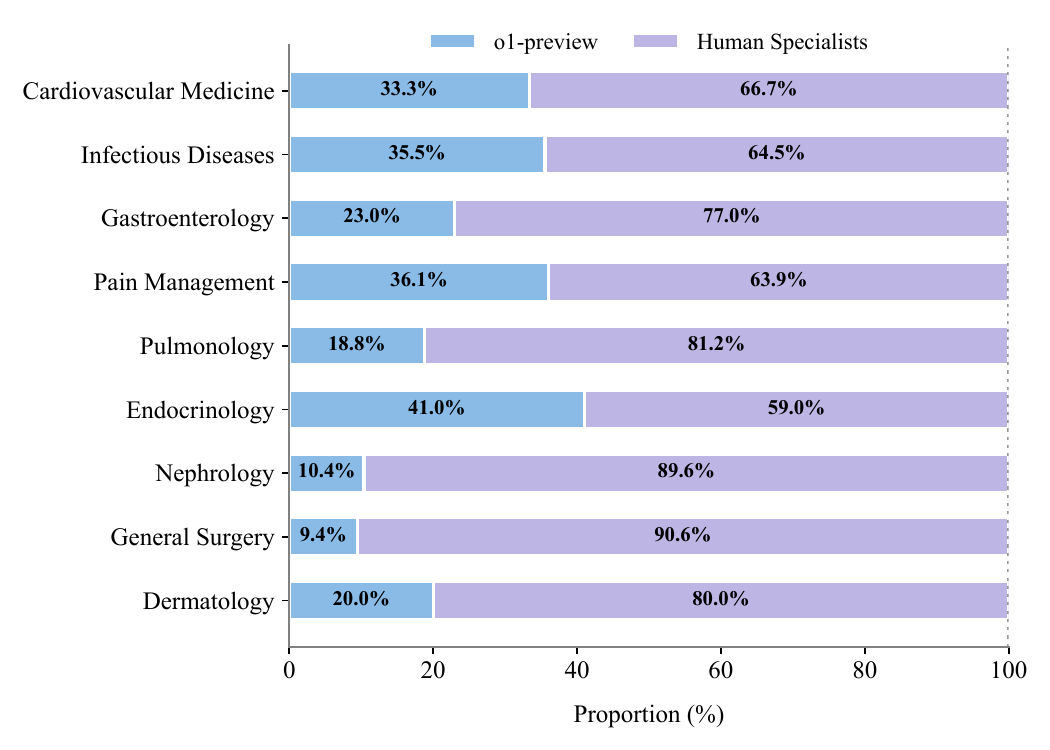}}
    \subcaptionbox{o3-mini vs. Specialists\label{fig:sub4}}[0.48\textwidth]
        {\includegraphics[width=0.48\textwidth]{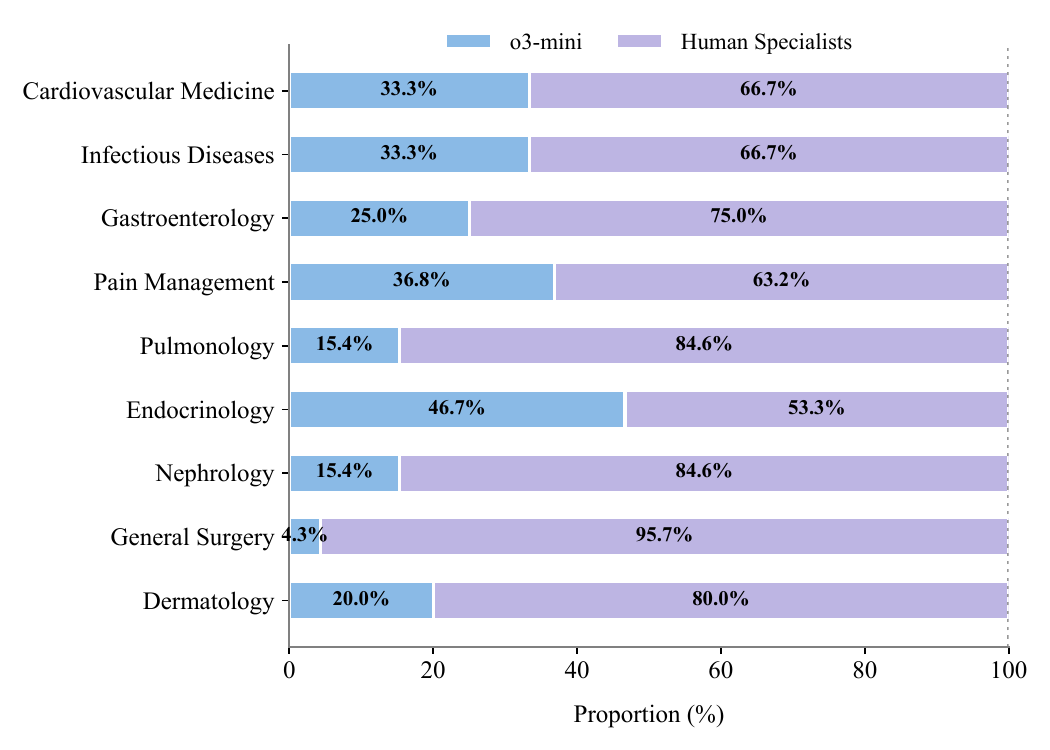}}
    \begin{center}
        \subcaptionbox{DxDirector-7B (Ours) vs. Specialists\label{fig:sub5}}[0.48\textwidth]
            {\includegraphics[width=0.48\textwidth]{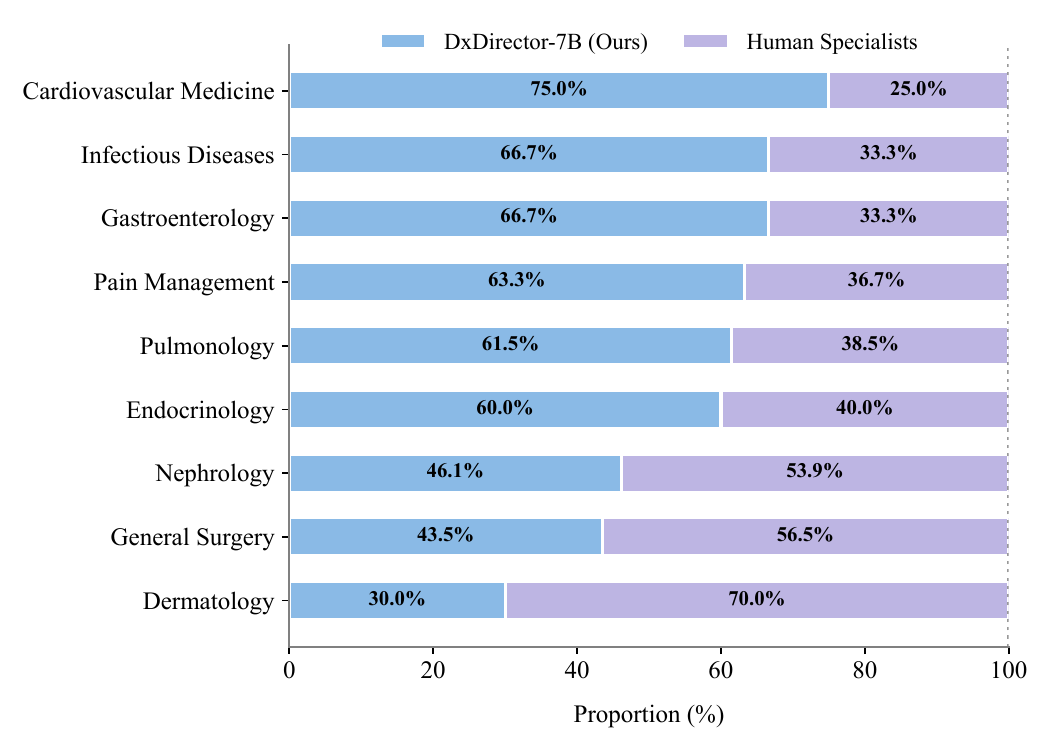}}
    \end{center}
    
    \caption{The proportion of the diagnoses generated by LLMs can completely replace those of medical specialists in each department. The assessment is conducted by double-blinded adjudication between LLMs and specialists in the corresponding department.}
    \label{beiyi_com}
\end{figure}

The results of the second aspect are shown in Fig.~\ref{beiyi_com}. In this evaluation, the third-party agent assesses whether the diagnoses generated by DxDirector-7B could replace those made by medical specialists. In the bar of Fig.~\ref{beiyi_com}, the proportion of various LLMs indicates the ratio of samples that LLMs can replace specialist physicians to the total number of samples. All baseline LLMs fail to outperform specialist physicians in all departments. On the contrary, as for our DxDirector-7B, the diagnostic
contents generated by DxDirector-7B in cardiovascular medicine achieves a 75.0\% replacement rate to specialist physicians. In infectious diseases, gastroenterology, pain management, pulmonology and endocrinology, DxDirector-7B achieves 60\%--66.7\% replacement rate to specialist physicians. These departments emphasize the comprehensive analysis and reasoning of clinical testing information. For departments such as dermatology and general surgery where physical operations such as contact, observation, and real-time response are dominant, DxDirector-7B cannot achieve replacement rates of more than $50\%$, because in real-world clinical diagnosis, these departments strongly rely on frequent real interactions with human physicians and patients.

\subsection{Evaluations on US Medical License Exam Consisting of Various Clinical Tasks}
In this section, we assess the performance of LLMs using the United States Medical Licensing Examination (USMLE) dataset, which comprises 1,273 publicly available cases covering various clinical tasks, such as diagnosis, differential diagnosis, prevention, etiological analysis, and so on. To better replicate realistic clinical scenarios and elevate the complexity, we convert the original multiple-choice format of the USMLE questions into open-ended questions. This transformation demands more sophisticated reasoning and clinical inference from the LLMs. In this transformed dataset, LLMs are required to address questions across various tasks under full-process diagnosis setting. Here, only the patient's chief complaint is initially provided, and the LLMs must actively infer and gather more detailed clinical information through more reasoning.

\begin{figure}[H]
    \centering
        \includegraphics[width=1.0\linewidth]{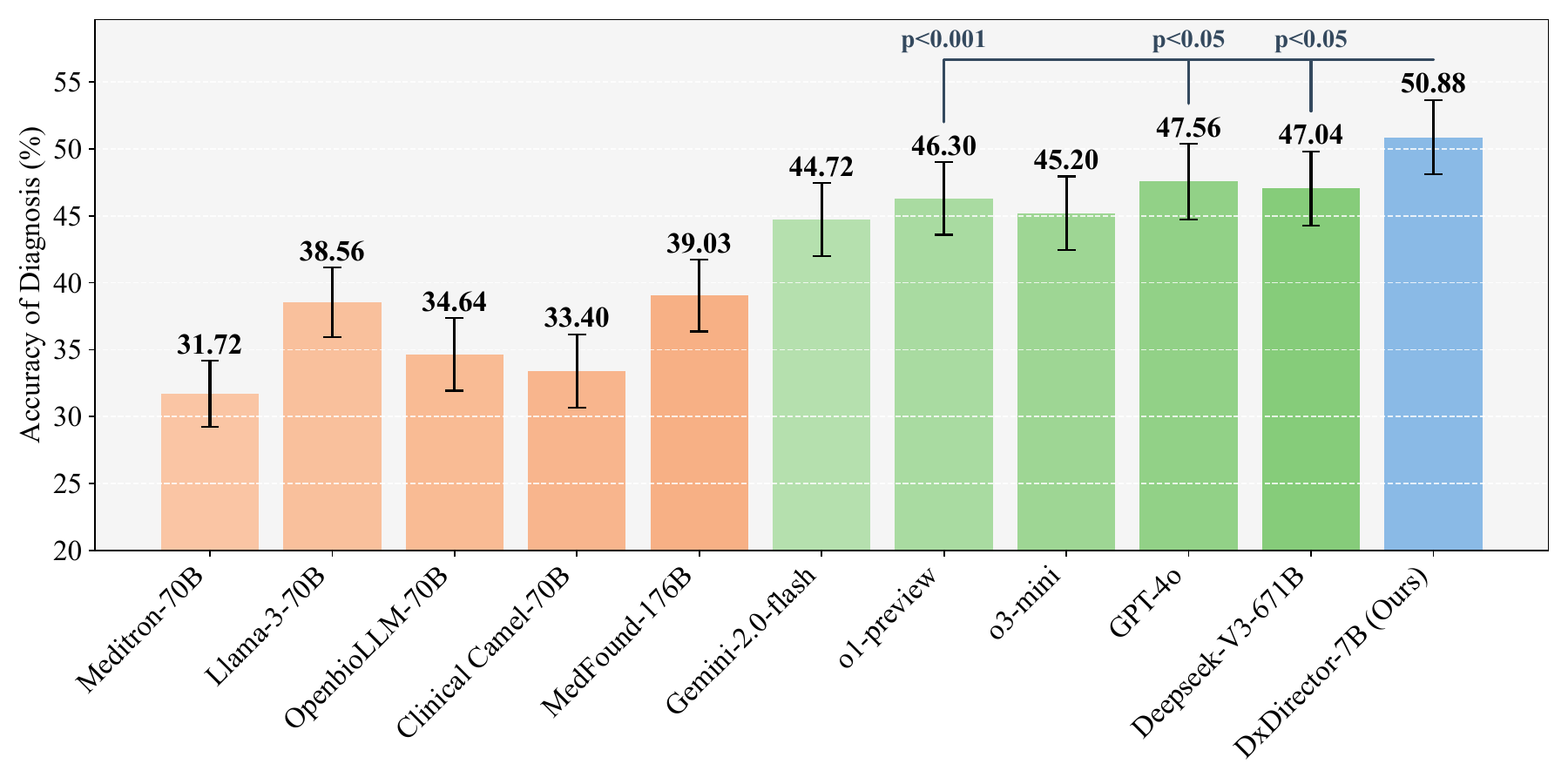}
        \caption{Accuracy of answering questions about various clinical tasks on US Medical License Exam in full-process diagnosis setting. Bars are annotated with the accuracy of each LLM. Error bars reflect $95\%$ confidence intervals determined by non-parametric bootstrap procedure with 1,000 samples. We perform statistical significance tests utilizing the two-side McNemar test between DxDirector-7B and the top-3 baseline, with p-value levels annotated on the bars.}
        \label{medqa_acc}
\end{figure}

\paragraph{Overall Accuracy}
The overall performance of various LLMs on the US Medical Licensing Examination (USMLE) is illustrated in Fig.~\ref{medqa_acc}. Our DxDirector-7B achieves the highest accuracy ($50.88\%$), underscoring its superior capabilities not only in making diagnosis but also across a broader array of clinical tasks, thereby highlighting its versatility for practical healthcare applications. Notably, DxDirector-7B outperforms medically adapted LLMs such as MedFound-176B, attaining a significant absolute improvement of $11.85\%$ despite having only approximately one-tenth of the parameter size
(7B compared to 70B and 176B). Furthermore, medically adapted LLMs with similar parameter sizes (OpenbioLLM-70B, Clinical Camel-70B, and Meditron-70B) demonstrate inferior performance (by $\Delta=-6.84\% \sim-3.92$\%) compared to the general LLM Llama-3-70B. This observation suggests that existing medical adaptation methods, while effective at enhancing diagnostic accuracy, may inadvertently compromise performance on other critical tasks in full-process clinical diagnosis setting. Collectively, these comparisons emphasize the efficacy and generalizability of our training method employed in developing DxDirector-7B.

\begin{figure}[H]
    \centering
        \includegraphics[width=0.9\linewidth]{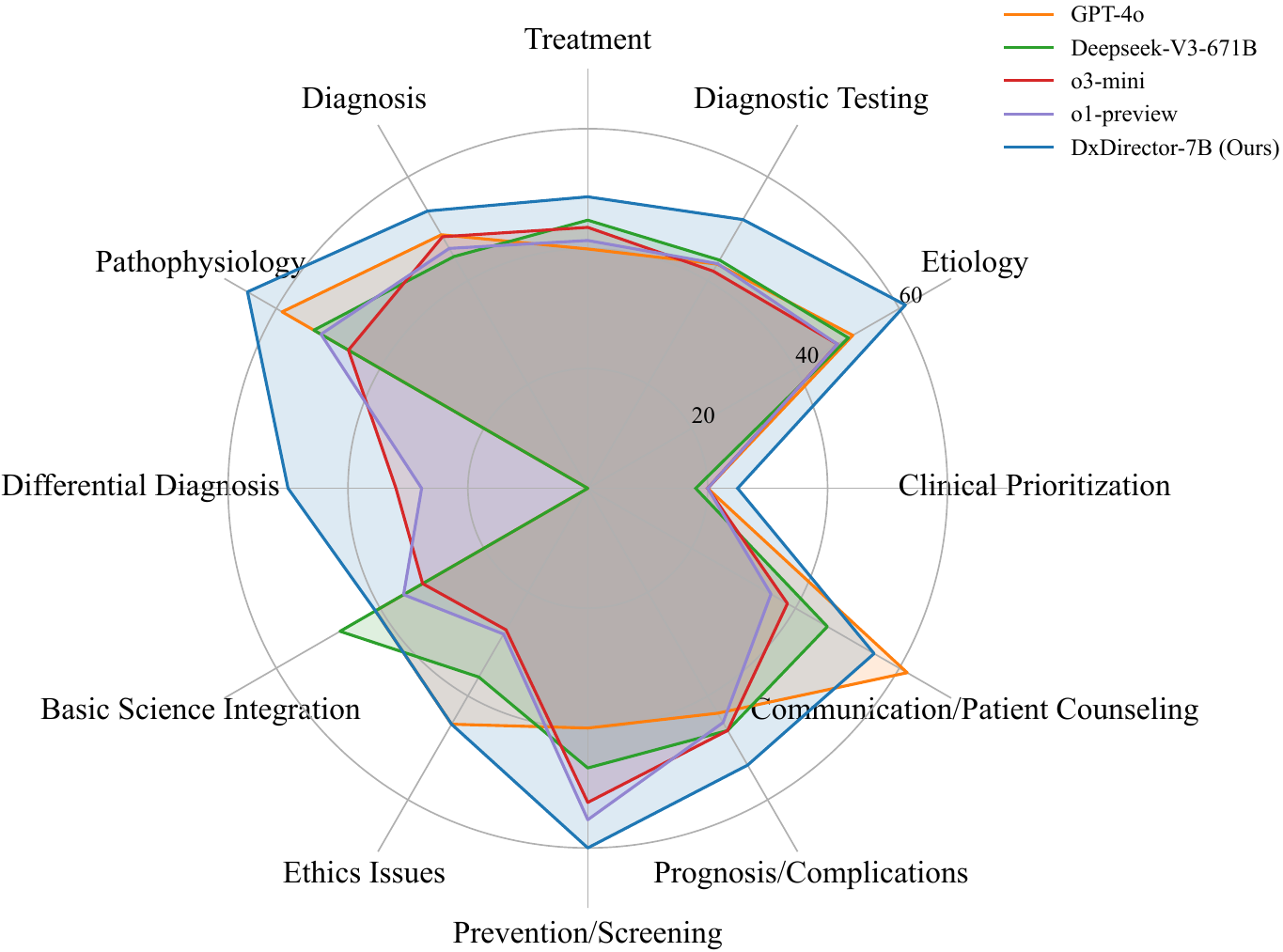}
        \caption{The visual comparison among various LLMs on 12 clinical tasks in USMLE.}
        \label{medqa_leida}
\end{figure}

\paragraph{Specific Accuracy on Twelve Clinical Tasks}
Fig.~\ref{medqa_leida} provides a detailed comparison of the performance of various LLMs across 12 clinical tasks within the USMLE dataset, offering granular insight into their comprehensive clinical capabilities in full-process diagnosis. Our DxDirector-7B outperforms all powerful commercial LLMs on 10 out of 12 tasks. Specifically, DxDirector-7B
achieves the large absolute improvement than all baselines in differential diagnosis ($\Delta=18.00\%$) and etiology ($\Delta=10.13\%$). Both of these two tasks require LLMs to obtain as detailed and accurate clinical test information as possible to rule out potential disease options and determine the etiology. So our DxDirector-7B, equipped with strong full-process diagnostic driving capability, significantly outperforms other LLMs in this regard. DxDirector-7B performs worse than Deepseek-V3-671B in basic science integration, this is primarily due to the substantial gap in the number of parameters between the two (nearly 100 times), which renders DxDirector-7B's ability to memorize basic medical knowledge comparatively weaker. Besides, GPT-4o with powerful chat capabilities is better than DxDirector-7B in patient communication. Compared with other tasks, basic science integration and patient communication are not strongly dependent on the ability to drive the entire diagnostic process, so these two are not specially optimized in DxDirector-7B. Overall, our DxDirector-7B has surpassed the existing strongest commercial LLMs on most critical clinical tasks in full-process diagnosis setting, despite having a significantly lower parameter count and training cost.


\begin{figure}[t]
  \begin{minipage}[t]{0.33\textwidth}
    \centering
    \subcaptionbox{MedFound-176B\label{mis_1}}[0.9\textwidth]{
      \includegraphics[width=\textwidth]{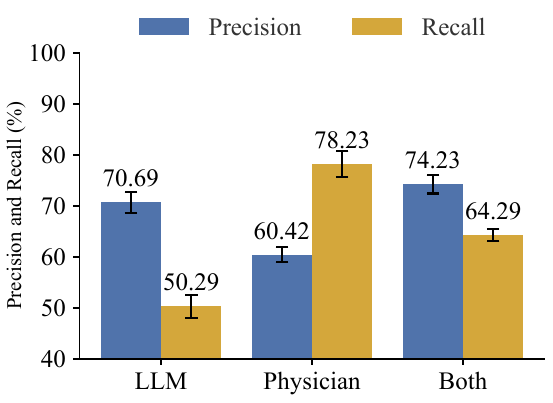}
    }
  \end{minipage}\hfill
  \begin{minipage}[t]{0.33\textwidth}
    \centering
    \subcaptionbox{GPT-4o\label{mis_2}}[0.9\textwidth]{
      \includegraphics[width=\textwidth]{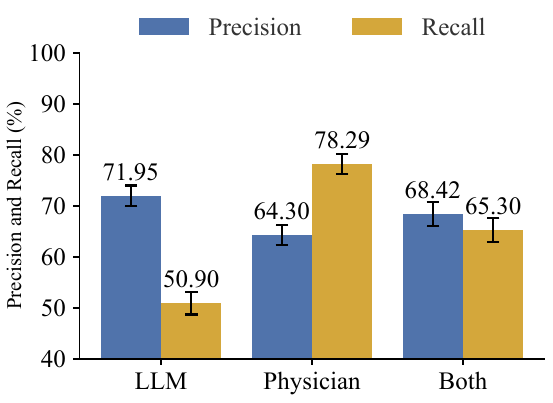}
    }
  \end{minipage}\hfill
  \begin{minipage}[t]{0.33\textwidth}
    \centering
    \subcaptionbox{o1-preview\label{mis_3}}[0.9\textwidth]{
      \includegraphics[width=\textwidth]{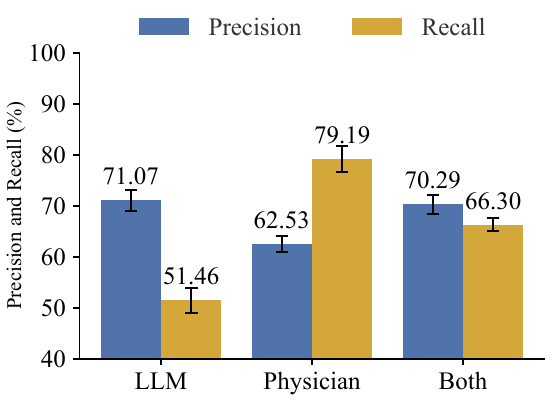}
    }
  \end{minipage}

    \begin{minipage}[t]{0.33\textwidth}
    \centering
    \subcaptionbox{o3-mini\label{mis_4}}[0.9\textwidth]{
      \includegraphics[width=\textwidth]{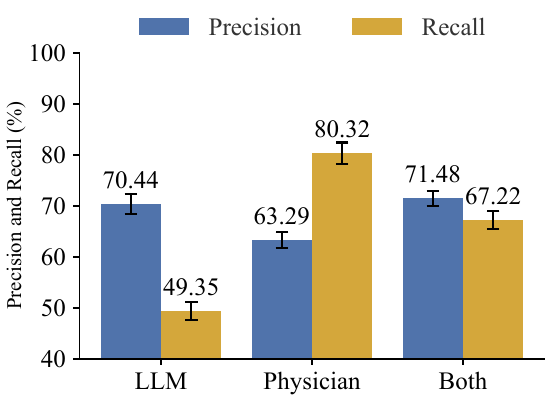}
    }
  \end{minipage}\hfill
  \begin{minipage}[t]{0.33\textwidth}
    \centering
    \subcaptionbox{Deepseek-V3-671B\label{mis_5}}[0.9\textwidth]{
      \includegraphics[width=\textwidth]{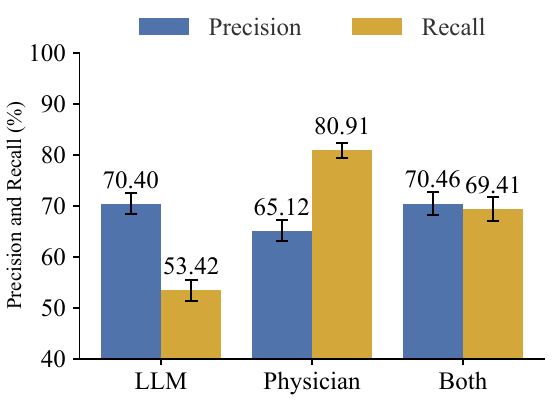}
    }
  \end{minipage}\hfill
  \begin{minipage}[t]{0.33\textwidth}
    \centering
    \subcaptionbox{DxDirector-7B (Ours)\label{mis_6}}[0.9\textwidth]{
      \includegraphics[width=\textwidth]{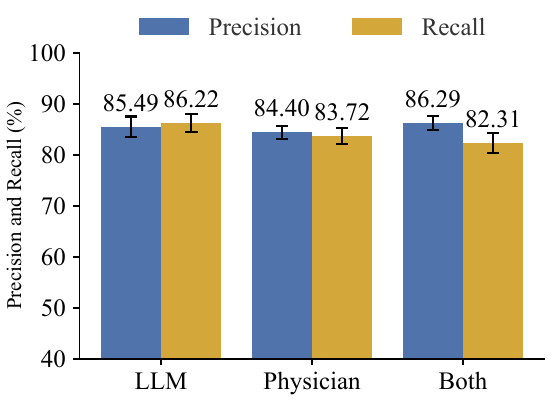}
    }
  \end{minipage}
  

  \caption{The effectiveness of the accountability mechanism for the diagnosis content generated by different LLMs in misdiagnosis scenarios. This is a classification task with 3 classes including the responsibilities of LLMs, human physicians or both of them. We use precision and recall for each task as the evaluation metrics. For example, bars for LLM class means the recall and precision for identifying the misdiagnoses caused by LLMs. Error bars reflect $95\%$ confidence intervals determined by non-parametric bootstrap procedure with 1,000 samples.}
  \label{fig:misdiagnosis}
\end{figure}
\subsection{Accountability in Misdiagnosis}
In this section, we evaluate the accountability of DxDirector-7B in cases of misdiagnosis. Unlike existing LLMs, which present the entire diagnostic reasoning process interwoven with multiple human physician operations without clear distinction, DxDirector-7B explicitly structures the diagnostic process (Fig.~\ref{DxDirector}). Each diagnostic step is clearly itemized, distinguishing the content generated by the LLM from that provided by human physicians, and each LLM-generated step is explicitly attached by authoritative medical literature. This structured approach enables precise identification of specific erroneous steps during a misdiagnosis and clarifies responsibility between the LLM and physicians.

To evaluate the accountability capability of DxDirector-7B in cases of misdiagnosis, we simulate diagnostic errors by introducing perturbations to randomly selected steps within the diagnostic process to make LLM generate the incorrect diagnosis. The specific perturbation method is employing Deepseek-V3 to rewrite the selected steps to generate new content that is factually inconsistent with the original content. These perturbations can impact either the LLM-generated content, the involvement of human physicians, or both. In this scenario, a GPT-4o-based agent is employed to assess whether it can correctly identify the source of misdiagnosis—whether attributed to human physicians, LLMs, or both—thereby evaluating the effectiveness of the accountability mechanism. This evaluation constitutes a three-class classification task, utilizing precision and recall as primary metrics. The evaluation dataset comprises 1,500 sampled cases that all LLMs can generate the correct diagnosis from RareArena and ClinicalBench. Results, presented in Fig.~\ref{fig:misdiagnosis}, indicate that DxDirector-7B's accountability mechanism achieves the highest precision and recall across all categories compared to baseline LLMs. Notably, baseline LLMs typically attribute errors disproportionately to human physicians over LLMs, reflected by higher recall but lower precision for physician accountability than LLM (recall: $78.23\% \sim 80.91\%$ vs. $49.35\% \sim 53.42\%$; precision: $60.42\% \sim 62.15\%$ vs. $70.40\% \sim 71.95\%$), which means that the diagnostic content generated by baselines makes physicians over-accountable compared to LLMs. In contrast, DxDirector-7B maintains comparable and high precision and recall for both LLM and physician accountability. This means that providing clear and fine-grained content attached by authoritative medical literature is of great significance for achieving an accurate medical accountability mechanism.

\vspace{-0.5em}
\section{Discussion}

\vspace{-0.5em}
\paragraph{Reverse the Physician-AI Relationship}
Existing AI remains largely as an assistant to physician. This AI-assisted working pattern limits AI’s ability to fully reduce physicians’ workload and enhance diagnostic efficiency. In this paper, we propose an innovative paradigm that reverses the relationship between AI and physicians, that is, training LLM to be a director in the entire process of clinical diagnosis, while physicians become assistants of the LLM, providing simple help only when necessary with the principle of minimizing physicians’ involvement. It advances effective AI deployment in full-process clinical workflows of the real world, reducing the workload of physicians to the greatest extent possible and indicating the efficient, accurate and scalable diagnostic solution.

\vspace{-1em}
\paragraph{Superior Diagnostic Accuracy in Full-process Clinical Diagnosis Setting}
Based on above design, we propose DxDirector-7B, an LLM with powerful deep thinking ability, can effectively drive the full-process clinical diagnosis with the minimal human efforts to reduce the workload and lessen the demand for specialized expertise of human physicians in practical clinical tasks as much as possible. We evaluate our DxDirector-7B in full-process setting across four publicly authoritative datasets including rare and complex cases, and a real-world clinical diagnostic scenario set in a top-tier hospital in China. As for accuracy of diagnosis, experimental results in Fig.~\ref{diag_acc},  Fig.~\ref{fig:beiyi_xiangxian} and Fig.~\ref{medqa_acc} indicate that our DxDirector-7B significantly surpasses the medically adapted LLMs with dozens of times more parameters, such as MedFound-176B and commercial general-purpose LLMs with nearly 100 times more parameters, such as GPT-4o, o1-preview, o3-mini, and Deepseek-V3-671B. This means that our DxDirector-7B is not only accurate but computationally economical, which is a solid step towards the low-cost and effective application of LLMs in practical clinical diagnostics.

\vspace{-1em}
\paragraph{Significantly Reduce the Physician Workload}
The workload human physicians need to pay in full-process diagnosis is crucial metrics for evaluating the empowerment of AI in clinical diagnosis. Experimental results in Fig.~\ref{fig:count} and Fig.~\ref{fig:rate} show that in the entire process of clinical diagnosis, our DxDirector-7B achieves the minimal physicians' workload, the maximum physicians' efficiency and the most accurate diagnosis compared with exisiting state-of-the-art LLMs. This means DxDirector-7B is an efficient, accurate and scalable diagnostic solution.

\vspace{-1em}
\paragraph{Comprehensive Understanding at Department-Level}
In order to gain a more comprehensive understanding of the advantages and disadvantages of our DxDirector-7B across different clinical departments. We report the comparison between DxDirector-7B and baselines at the departmental level in Fig.~\ref{clinical_heatmap} (17 departments, real-world cases) and Fig.~\ref{rare_heatmap} (19 departments, rare diseases). The results indicate that our DxDirector-7B achieves significant superior diagnostic accuracy on most departments than all powerful baselines, especially on departments that are complex and require many diagnostic tests, such as oncology. This means DxDirector-7B has more powerful ability to drive multiple appropriate diagnostic tests in full-process clinical diagnosis to iteratively enrich the clinical information for accurate diagnosis, which is because DxDirector-7B can continuously perform deep thinking to make the optimal decision. This emphasizes the importance of developing LLMs with greater deep thinking abilities for clinical diagnosis.

\vspace{-1em}
\paragraph{Promisingly Substitute Medical Specialists in Real-World diagnosis}
Experiments on real-world cases across 9 clinical departments in top-tier hospital in China further demonstrate the advantages of our DxDirector-7B in practical clinical diagnostic applications (Fig.~\ref{fig:beiyi_xiangxian}). Evaluations with the participation of human medical specialists show that the diagnoses generated by our DxDirector-7B achieve substitution for human medical specialists in 60\% to 75\% of cases in many department (Fig.~\ref{beiyi_com}). This results surpass all state-of-the-art LLMs and indicate the potential of DxDirector-7B to serve as a viable substitute for medical specialist in real-world diagnosis.

\vspace{-1em}
\paragraph{Superior Performance on Various Clinical Tasks}
Making a diagnosis is not the only task in clinical practice. We further evaluate LLMs on 12 clinical tasks (differential diagnosis, treatment, etiology and so on.) at US Medical License Exam level. Our DxDirector-7B outperforms all powerful commercial LLMs on 10 out of 12 tasks, especially achieves significant improvement on the tasks that require LLMs to obtain as detailed and accurate clinical test information as possible, such as differential diagnosis and etiology. DxDirector-7B is expected to become an all-around director in clinical diagnosis, accurately completing various clinical tasks by minimizing the need for assistance from human doctors.

\vspace{-1em}
\paragraph{Accurate Accountability Mechanism}
Establishing a clear and accurate accountability mechanism between human physicians and AI for misdiagnosis is very important at a time when both parties are working closely together. Our DxDirector-7B generates the clear diagnosis (Fig.~\ref{DxDirector}) in which each step is listed separately item by item, the content generated by LLM is clearly distinguished from the human physicians, and each step generated by LLM is attached by authoritative medical literature. This structured output provides a basis for clearly identifying the specific erroneous medical steps and distinguishing the responsibilities between physicians and LLMs in a misdiagnosis. Evaluation on a large misdiagnosis dataset shows that compared with all powerfull LLMs in baselines, the diagnoses generated by DxDirector-7B can provide a more accurate and fair mechanism for accountability in misdiagnosis.

\vspace{-1em}
\paragraph{Value and Impact}
These results show our DxDirector-7B successfully reshapes the collaborative relationship between AI and human physicians, which indicates the new era where AI, traditionally has always been regarded as a physician's assistant, now assumes a director role in autonomously steering the entire diagnostic process with the principle of minimizing the physician's involvement. This paradigm shift is designed to substantially alleviate the workload of human physicians, enhancing efficiency while improving diagnostic accuracy. DxDirector-7B significantly reduces the reliance on human physicians' workload and professional expertise, thus lowering barriers to quality medical diagnosis. By delivering a low-cost, efficient, and accurate clinical solution, DxDirector-7B offers profound implications, particularly for medically underserved and resource-limited regions and is promising to be applied into various clinical departments and tasks. Besides, the significantly lower parameter count and training and inference costs compared to existing state-of-the-art LLMs enable DxDirector-7B to be applied to various medical institutions at low cost.

\vspace{-1em}
\paragraph{Limitations and Future Work}
Although our DxDirector-7B has demonstrated impressive performance, there are still some points worth studying in the future. For example, more refined and diversified rules for human physician participation can be defined for each clinical department to further improve efficiency. Besides, the assistants for DxDirector-7B can be not only human physicians but other AI models for healthcare. For example, DxDirector-7B can call on various specialized pathology analysis models with visual understanding capabilities for radiology~\cite{lu2024visual,zhang2023knowledge,christensen2024vision,tanno2024collaboration}, echocardiography~\cite{huang2023visual}, cell slice analysis~\cite{chen2024slidechat}, pathology~\cite{chen2024towards}, and so on. This can further reduce the workload of human physicians and improve accuracy. From a higher perspective, DxDirector-7B can act as a director in establishing an efficient diagnostic framework that promotes effective collaboration among the three key entities: physicians, patients, and various specialized AI models. Future exploration in these directions will revolutionize the existing healthcare paradigm. A large language model with exceptional reasoning capabilities will enable the fully automated, efficient mobilization and integration of various medical resources, significantly enhancing both the efficiency and accuracy of healthcare delivery.

\section{Methods}
In this section, we introduce the detailed training method of our DxDirector-7B, which can be divided into three stages: (1) Continued pre-training on medical data; (2) Instruction-tuning for full-process clinical diagnosis; (3) Step-level strategy preference optimization. Then we introduce more details about our experiments.

\subsection{Continued Pre-training on Medical Data} \label{pretrain}
This stage enables the general LLM to acquire medical knowledge, which forms the foundation for its clinical diagnosis capabilities. Specifically, in this stage, we train open-source LLM (Llama-2-7B) on large-scale medical texts by cross-entropy loss function in the paradigm of next token prediction, being supervised by the learning signals from medical texts themselves. For example, given $\mathcal{C}$ is a dataset of texts, $C_i=[c_1,c_2,c_3,...c_n]$ is one of the text sequence in $\mathcal{C}$ ($C_i \in \mathcal{C}$), $\mathcal{M}(c_t|c_1,c_2,...c_{t-1}; \theta)$ is the probability of $c_t$ estimated by the LLM given prefix $[c_1,c_2,...c_{t-1}]$ and model parameters $\theta$. the training objective in this continued pre-training stage is minimizing the negative log-likelihood over set $\mathcal{C}$ as:
\begin{equation}
    \min_{\theta} \sum_{C_t \in \mathcal{C}} \sum_{t=1}^{n} -\log \mathcal{M}(c_i|c_1,c_2,...c_{t-1}; \theta).
\end{equation}
We collect publicly available medical data for this continued pre-training including $35$K articles from clinical guidelines, $16.1$M paper abstracts from PubMed and PubMed Central, $5$M full papers from PubMed and PubMed Central~\cite{chen2023meditron}. Llama-2-7B is trained on these datasets to memorize the basic medical knowledge. Besides, we use experience replay~\cite{chen2023meditron} to maintain the original general knowledge of LLama-2-7B by mixing $500$K general domain data from Wikipedia, ArXiv, books, and StackExchange into the training datasets.

\subsection{Instruction-tuning for Full-process Clinical Diagnosis} \label{sft_section}
Instruction-tuning is the process of training the LLM that has been pre-trained to generate expected responses for user's input instructions~\cite{zhang2023instruction}. Full-process clinical diagnosis in the real world, especially for rare and complex cases, often involves multiple complex medical knowledge and multiple diagnostic test procedures. This section introduces our proposed novel instruction-tuning method that enables our DxDirector-7B to drive full-process clinical diagnosis solely starting with ambiguous chief complaints, through a step-by-step reasoning and continuous deep thinking. The training at this stage aims to endow DxDirector-7B with four key capabilities:
\begin{enumerate}[leftmargin=*, noitemsep] 
    \item \textbf{Progressive Clinical Information Reasoning}: DxDirector-7B can gradually reason valuable clinical information starting with a patient's vague complaint in a step-by-step manner, ultimately completing clinical tasks such as diagnosis, treatment plan design, and so on. Each step may involve inference about medical knowledge, analyzing clinical phenomena, or designing diagnostic tests.
    \item \textbf{Step-Level Deep Thinking}: DxDirector-7B possesses deep thinking—that mimics human “slow thinking capabilities at the step level. When determining the specific strategy for each step, it first generates a thinking process that analyzes the currently available information and the expected goal to define the optimal strategy for the current step.
    \item \textbf{Human Assistance When Necessary}: In cases where the strategy requires clinical operations for diagnostic test---such as medical imaging, physical examinations, laboratory tests, and so on, which computer program-based LLMs cannot complete---DxDirector-7B can request assistance from human physicians and continue reasoning after receiving the assistance.
    \item \textbf{Autonomously Generate Final Diagnosis}: DxDirector-7B can autonomously decide whether it can make am accurate final diagnosis based on the currently available information. The generated final diagnosis is a concise and clear summary of the step-by-step reasoning process, with each involved medical knowledge attached by authoritative medical literature.
\end{enumerate}
We introduce the details about this stage in two parts: dataset construction and training.

\subsubsection{Data Construction}
Constructing the suitable training dataset is the prerequisite. Our data construction can be divided into three steps: (1) raw data collection, (2) data transformation, and (3) deep thinking injection. The pipeline of our data construction is shown in Fig.~\ref{data_construction_sft}.

In raw data collection, our raw data is collected from MedQA~\cite{jin2021disease}, a medical question-and-answer dataset enriched with extensive context, including detailed clinical information such as patient profiles, disease symptoms and histories, diagnostic test results, vital signs, and so on. Questions in this dataset covers multiple clinical tasks such as diagnosis, differential diagnosis, designing treatment plan, screening, analyzing etiology, and so on. This dataset contains $10,178$ samples. After collecting the raw dataset, we perform data transformation on it to construct the data aligning with full-process diagnosis. We use GPT-4o API for fully automated data transformation, and introduce human medical experts perform sample evaluations of the transformed data to ensure the quality. For each data sample, data transformation consists of three steps as shown in ``Data Transformation'' part in Fig.~\ref{data_construction_sft}: 
\begin{figure}[t]
    \centering
        \includegraphics[width=1.0\linewidth]{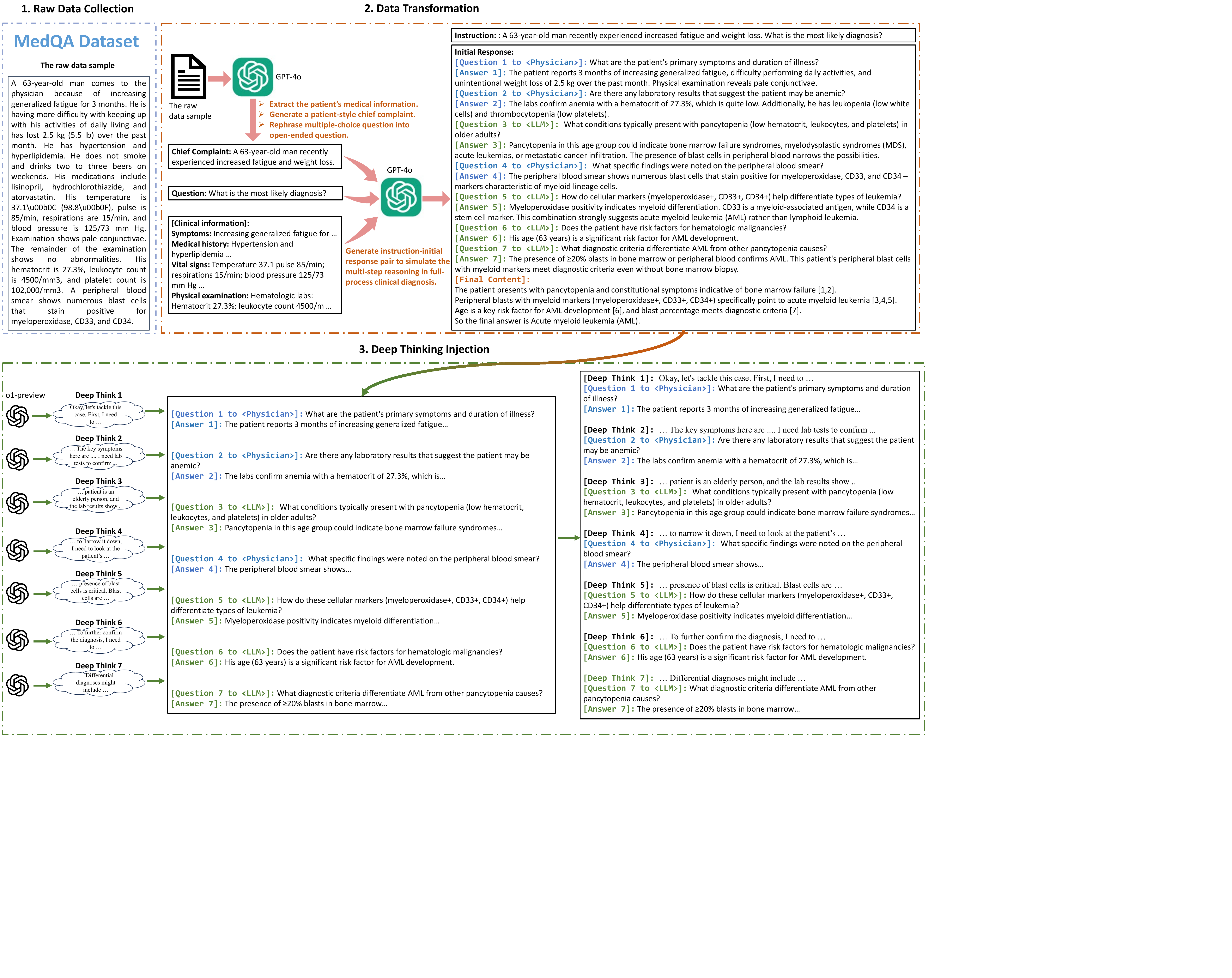}
        \caption{Pipeline of our data construction for instruction-tuning for full-process clinical diagnosis.}
        \label{data_construction_sft}
\end{figure}

Firstly, we use GPT-4o to extract the patient's detailed clinical information from the context, such as symptoms, medical history, all diagnostic test results, and vital signs, and so on. In full-process clinical diagnosis, this information cannot be obtained at the beginning, but is gradually obtained during the complex consultation process.

Secondly, we use GPT-4o API to rewrite the original data into a patient-style chief complaint, which is a simple, vague, and non-professional description provided by the patient about their condition without any specific clinical information. This is the only information that LLMs can obtain from patients at the beginning of real-world full-process clinical diagnosis. Besides, we rephrase multiple-choice question in original data into open-ended question, which is closer to real-world clinical diagnosis. The rewritten chief complaint and open-ended question is synthesized into the instruction for the latter tuning.

Thirdly, we give the clinical information to GPT-4o and and design examples and complex prompts (i.e., in-context learning~\cite{wies2023learnability}) to instruct GPT-4o to convert the provided information to simulate the step-by-step reasoning process in full-process clinical diagnosis starting with the patient's chief complaint. In this way, we construct the initial instruction-response pairs to fine-tune our DxDirector-7B. As shown in the ``Initial Response'' of data transformation stage in Figure~\ref{data_construction_sft}, each step consists of a question-answer pair in the paradigm of self-questioning and self-answering. The questions are divided into two types: one is inquiries or inference based on objective medical knowledge, such as the causes of a specific disease or determining the possible disease based on the patient's specific symptoms, and so on. These questions are marked as ``\texttt{$<$LLM$>$}'' and their answers can be finished by LLM itself. The other type of question involves inquiring about the clinical operations for diagnostic test or communication to patients. They are marked as ``\texttt{$<$Physician$>$}'' and their answer must be finished with the help of human physicians. When the reasoning is finished, the ``\texttt{[Final Content]}'' is generated, it is the summary of the reasoning process, with the number of reasoning steps marked at the corresponding positions. This enhances the credibility and error-correctability of the AI-generated diagnostic process. It is worth noting that GPT-4o  cannot effectively perform full-process diagnosis. Therefore, to construct this data, we add the patient's detailed clinical information to the context to GPT-4o. GPT-4o simulates the full-process diagnosis under the premise of already knowing all related information, which simplifies the task significantly. This approach enables us to construct a large amount of data that meets the requirements at a relatively low cost.

After transformation, we inject the deep thinking content for each step of ``Initial Response''. As shown in deep thinking injection stage in Figure~\ref{data_construction_sft}, we use o1-preview to generate detailed thinking content for each step of ``Initial Response''. This thinking should fully consider the clinical information at the current step and combine it with the ultimate clinical goal to reason about the optimal strategy that should be taken at the current step, which simulates the human "slow thinking" process. Deep thinking makes the logical connection between each step in the whole process of clinical diagnosis closer. These contents often only appear in the minds of human physicians and are not written in electronic medical records. The explicit generation of these contents enables DxDirector-7B to have the "slow thinking" ability like human physicians. We do not generate deep thinking end-to-end during the data transformation stage because we find that doing so will result in deep thinking revealing currently unknown clinical information in advance.

The data instance in final instruction-response pairs for instruction-tuning for full-process clinical diagnosis is the instruction consisting of patient's chief complaint and clinical question, and the response consisting of multi-step reasoning, deep thinking and final diagnosis.

During data construction, we randomly sample the transformed instruction-response pairs and provide them to human medical experts for evaluation to determine whether this data aligns with real clinical diagnostic scenarios. We collect feedback from the medical experts and continuously refine our prompts to optimize the quality of the data. The detailed prompts in data construction can be found in Supplementary Fig. 11 to 20. Finally, we obtain $10,178$ high-quality instruction-response pairs for training.

\subsubsection{Training with Decoupled Reasoning and Knowledge}
We train DxDirector-7B to perform full-process clinical diagnosis on the constructed dataset above. As shown in instruction-response pair of Figure~\ref{data_construction_sft}, given the instruction, DxDirector-7B is trained to generate the response consisting of numbered ``\texttt{[Deep Think]}'' and numbered ``\texttt{[Question]-[Answer]}'' pairs. ``\texttt{[Deep Think]}'' and ``\texttt{[Question]}'' emphasize reasoning capability while ``\texttt{[Answer]}'' emphasize medical knowledge recalling capability. We propose a decoupled training method based on loss-masking to enable DxDirector-7B to learn these two capabilities separately. It trains the two capabilities alternately in batches. When training reasoning ability, the loss function only computed over the content in ``\texttt{[Deep Think]}'' and ``\texttt{[Question]}'', while the content in ``\texttt{[Answer]}'' and ``\texttt{[Final Content]}'' is masked. This allows the LLM to focus on reasoning the questions to be addressed at each step rather than recalling their answers. The cross-entropy loss function $\mathcal{L}_1$ for this can be computed as:
\begin{align}
    \mathcal{L}_1 = \sum_{R_{i} \in \mathcal{Q}} -\log \mathcal{M}(R_{i}|I,R_{1:i-1}; \theta),
\end{align}
in which $\mathcal{M}$ is the distribution of next token prediction of LLM, $I$ and $R$ are the instruction and response respectively of instruction-response pair in Figure~\ref{data_construction_sft}. $\mathcal{Q}$ is set of tokens in ``\texttt{[Deep Think]}'' and ``\texttt{[Question]}'', and $R_{1:i-1}$ is the prefix for the token $R_i$.

When training knowledge recalling capability, the loss function only includes the content in ``\texttt{[Answer]}'' and ``\texttt{[Final Content]}'', while the content in ``\texttt{[Deep Think]}'' and ``\texttt{[Question]}'' is masked. Since some answers need the assistance from  human physicians to obtain patient's clinical information, the input data used to training this ability is accompanied by the extracted patient's clinical information. The cross-entropy loss function $\mathcal{L}_2$ for this can be computed as:
\begin{align}
    \mathcal{L}_2 = \sum_{R_{i} \notin \mathcal{Q}} -\log \mathcal{M}(R_{i}|I,P,R_{1:i-1}; \theta),
\end{align}
in which $P$ is the patient's clinical information that used to simulated the assistance from  human physicians in training. 
The details of the hyperparameters in training are provided in the Section~\ref{implementation}.

\subsection{Step-Level Strategy Preference Optimization} \label{sspo}
We call the ``\texttt{[Question]}'' to be solved in each step derived by deep thinking of DxDirector-7B as “strategy”. After instruction-tuning, our DxDirector-7B has initially demonstrated the ability to drive full-process clinical diagnoses by generating the strategy step-by-step. However, instruction-tuning with token-level cross-entropy loss function cannot effectively make LLM learn how to make optimal strategy at each step, i.e., generate the most appropriate ``\texttt{[Question]}''. The further optimization in this stage enables DxDirector-7B to implicitly compare multiple potential strategies in deep thinking at each step and select the optimal strategy. This ensures that each step in complex clinical reasoning is correct and efficient, so that the diagnosis can be completed accurately while relying on the minimal human physicians’ efforts. 

The training method we proposed in this stage is called Step-Level Strategy Preference Optimization, a reinforcement learning algorithm that assigns different rewards to different strategies at the same step and trains DxDirector-7B to generate the strategy with higher reward. To achieve this, first, we construct the training data that consists of multiple strategy labeled with different rewards for each step, and then design the specific method for preference optimization training.

\subsubsection{Data Construction}
\begin{figure}[t]
    \centering
        \includegraphics[width=1.0\linewidth]{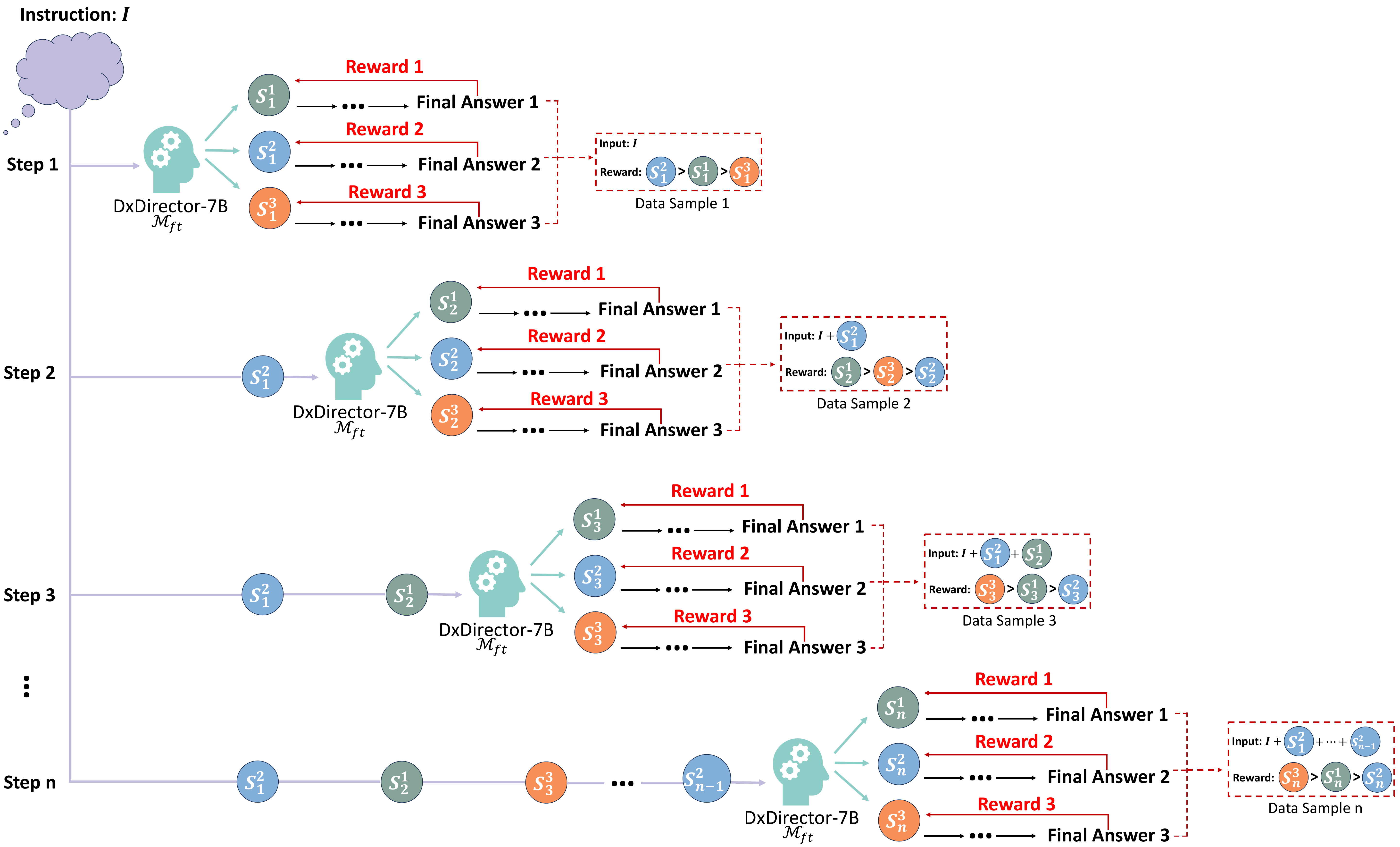}
        \caption{Pipeline of our data construction for Step-level Strategy Preference Optimization.}
        \label{data_construction_po}
\end{figure}
The overview of data construction for this stage can be found in Fig.~\ref{data_construction_po}. At step $t$, we give instruction $I$ and reasoning content from steps $1$ to $t-1$: $\{(d_1,q_1,a_1), (d_1,q_2,a_2),...,(d_t,q_{t-1},a_{t-1})\}$ as the input for $\mathcal{M}_{ft}$, in which $d_i$, $q_i$ and $a_i$ are the deep thinking, question (i.e., strategy) and answer respectively at the $i-th$ step, $\mathcal{M}_{ft}$ is the DxDirector-7B after instruction-tuning of the second stage. We change the random seed~\cite{kaczmarczyk2022backtesting} to make $\mathcal{M}_{ft}$ output $k$ different responses at step $t$:  $\{(d^1_t,q^1_t,a^1_t), (d^2_t,q^2_t,a^2_t),...,(d^k_t,q^k_t,a^k_t)\}$ when faced with this input. In our implementation, we set $k$ to $3$ and more sampling-related hyperparameters can be found in Section~\ref{implementation}. For each response $(d^i_t,q^i_t,a^i_t)$, we assign a reward $r^i_t$ to it based on the correctness of the final answer generated by the reasoning path continued with this response and the degree of reliance on the human physicians.  The correct answers are assigned higher rewards. For answers of the same correctness, the strategies that seek more assistance from human physicians will have a lower reward value than other strategies. The reward assigning strategy is as follows:
\begin{align}
    r^i_i = \begin{cases}
    \frac{10}{\gamma}, & \text{the final answer is correct} \\
    0,           & \text{the final answer is incorrect},
\end{cases}
\end{align}
in which $\gamma$ is the number of requesting for assistance from human physicians. We assign a corresponding reward to each response and obtain the set $\mathcal{S}_t=\{(d^1_t,q^1_t,a^1_t,r^1_t), (d^2_t,q^2_t,a^2_t,r^2_t),...,(d^k_t,q^k_t,a^k_t,r^k_t)\}$ with rewards at step $t$. We generate all unique ordered pairs over $\mathcal{S}_t$, which can be described as:
\begin{align}
    \mathcal{P}_t = \big\{ \big( (d^m_t, q^m_t, a^m_t, r^m_t), (d^n_t, q^n_t, a^n_t, r^n_t) \big) \mid r^m_t > r^n_t, 1 \leq m \leq k , 1\leq n \leq k \big\}.
\end{align}
In this way, we can get the data sample for strategy preference optimization training at step $t$, it consists of the input $X_t=I+\{(d_1,q_1,a_1), (d_2,q_2,a_2),...,(d_{t-1},q_{t-1},a_{t-1})\}$ and a set of paired responses $\mathcal{P}_t$. Our strategy preference optimization training method optimizes DxDirector-7B learn to make the better choice in each pair $(d^m_t, q^m_t, a^m_t, r^m_t), (d^n_t, q^n_t, a^n_t, r^n_t)$ of $\mathcal{P}_t$, maximizing the probability of generating the response with the higher reward while minimizing the probability of generating the response with the lower reward. Specific details about this will be introduced in Section~\ref{dpo_detials}. The response with the highest reward in $\mathcal{P}_t$ will be added to the prefix for data construction in step $t+1$. We use this strategy to iterate through each step of $2,000$ instruction-response pairs and finally get $23,608$ data samples for training, this dataset is denoted as $\mathcal{D}$ ($(X_t,\mathcal{P}_t) \in \mathcal{D}$).

\subsubsection{Step-Level Preference Optimization Training} \label{dpo_detials}
After data construction, we conduct step-level preference optimization training on $\mathcal{M}_{ft}$ to make it learn to implicitly compare multiple potential strategies in deep thinking at each step and select the optimal strategy with the principle of ensuring correctness while minimizing the workload of human physicians. The training objective is based on Direct Preference Optimization (DPO) loss function~\cite{rafailov2023direct}:
\begin{align}
    \mathcal{L}_{3}(\pi_\theta; \pi_{\text{ref}}) =
-\mathbb{E}_{(X_t, \mathcal{P}_t) \sim \mathcal{D}} \frac{1}{|\mathcal{P}_t|} 
\sum_{\substack{\big((d^m_t, r^m_t), \\ (d^n_t, r^n_t)\big) \in \mathcal{P}_t}} 
\left[
\log \sigma \left(
\beta \log \frac{\pi_\theta(d^m_t \mid X_t)}{\pi_{\text{ref}}(d^m_t \mid X_t)}
- \beta \log \frac{\pi_\theta(d^n_t \mid X_t)}{\pi_{\text{ref}}(d^n_t \mid X_t)}
\right)
\right],
\end{align}
in which $\pi_\theta(b|a)$ is probability of the policy model generating sequence $b$ given prefix $a$, $\pi_{\text{ref}}(b|a)$ is probability of the reference model generating sequence $b$ given prefix $a$, $\beta$ is a hyperparameter usually between $0.1$ to $0.5$. $\sigma$ is the sigmoid function. In this training objective, for each sample $(X_t, \mathcal{P}_t)$ in the training set $\mathcal{D}$, we traverse each response pair $\big((d^m_t, r^m_t), (d^n_t, r^n_t)\big)$ in $\mathcal{P}_t$ and align DxDirector-7B's deep thinking preference for the better strategy through the partial order relationship of rewards between $(d^m_t, r^m_t)$ and $(d^n_t, r^n_t)$, so as to enable DxDirector-7B to implicitly select the optimal strategy to generate among multiple potential strategies. For example, $X_t$ is the input to $\mathcal{M}_{ft}$ at the $t-th$ step, $r^m_t > r^n_t$, $d^m_t$ is the deep thinking for the strategy with higher reward and $d^n_t$ is the deep thinking for the strategy with lower reward. To optimize this loss function, DxDirector-7B should learn to maximize the probability 
of generating $d^m_t$ and minimize the probability of generating $d^n_t$. Since the strategy (``\texttt{[Question]}'') of each step is inferred from the content of deep thinking, optimization for deep thinking is more essential to enable DxDirector-7B to determine the optimal strategy for each step through more reasonable ``slow thinking'' like human. The details of the training are provided in the Section~\ref{implementation}.

\subsection{Training to Search Authoritative Medical Literature}
After instruction-tuning and preference optimization, as ``[\texttt{Final Content}]'' shown in Figure~\ref{data_construction_sft}, our DxDirector-7B can summarize after multi-step reasoning and mark the referenced reasoning step numbers at the corresponding positions. These numbers not only point to the reasoning steps but also indicate references to authoritative medical literature. This innovative design enhances the verifiability and credibility of AI-generated diagnostic content at a fine-grained level. This section introduces our detailed training method for medical literature search model.

\subsubsection{Training to Search}

\paragraph{Model Architecture. }The base model for search is Gemma-2B~\cite{team2024gemma}, a pre-trained language model with stacked $18$ transformer layers. We convert Gemma-2B into a vector representation model that can represent the query and each paragraph in medical literature to a dense vector as follows:

Given an input paragraph $P = \{x_1, x_2, \dots, x_T\}$ with $T$ tokens, the model outputs hidden states $\mathbf{H}^l \in \mathbb{R}^{T \times d}$ at each transformer layer $l \in \{1,2,\dots,18\}$. For text representation, we extract the last token's hidden state from the final transformer layer:
\begin{equation}
\mathbf{h}_{\text{text}} = \mathbf{H}^L[T, :] \in \mathbb{R}^d,
\end{equation}
in which $L=18$ denotes the last transformer layer, $d=2048$ is the hidden dimension of Gemma-2B, $T$ is the sequence length of the input paragraph.
This design leverages the autoregressive nature of Gemma-2B, where the final token's representation naturally aggregates contextual information from all preceding tokens through the transformer's self-attention mechanism. Given a query seeking medical knowledge, it represents the query and each paragraph in the corpus as vectors. The matching score between a query and a paragraph is determined by calculating the similarity between their vectors such as dot product. The paragraphs are then ranked in descending order based on their matching scores, and the top-k paragraphs are selected as the search results for the given query. To make the search model more accurate in the vector representation of medical text, we train Gemma-2B on large-scale medical data in contrastive learning method, which will be introduced below.

\paragraph{Data Collection and Process} We collect a large amount of text data from the medical domain to train our model. The training dataset consists of $11M$ medical articles come from medical textbooks, publications, and case reports. For each article, we use its title as the query and its abstract as the paragraph matched with this query. In this way we construct a large-scale query-paragraph paired data for training the search model.

\paragraph{Training}
 We use in-batch contrastive learning to train our model, enabling it to accurately represent text as vectors and rank texts on vector similarity. Specifically, each training batch consists of $b$ query-paragraph pairs. For the vector representation ($\mathbf{q}_i$) of a query in this batch, its positive sample is the paragraph paired with it ($\mathbf{p}_i$), while its negative samples are the $b-1$ paragraphs paired with other queries within the same batch ($\mathbf{p}_j$, $j \neq i$). For the query, in-batch contrastive learning aims to maximize the its vector similarity between the positive sample while minimizing its vector similarity between negative samples~\cite{xu2024search,xutheory}. The loss function to achieve this can be described as:
 \begin{align}
     \mathcal{L}_r = -\frac{1}{b} \sum_{i=1}^{b} \log \frac{e^{\mathbf{q}_i^\top \mathbf{p}_i}}{e^{\mathbf{q}_i^\top \mathbf{p}_i} + \sum_{\substack{j=1 \\ j \neq i}}^{b} e^{\mathbf{q}_i^\top \mathbf{p}_j}}
 \end{align}

\subsubsection{Fine-grand Medical Literature Search}

\paragraph{Indexing Corpus for Search} We build a large-scale database consisting of $23.9$M paragraphs from PubMed, $301.2$K paragraphs from StatPearls, $125.8$K paragraphs from medical textbooks as the corpus for medical literature search. We use our trained medical search model to represent each paragraph as a vector, and use IndexFlat~\footnote{\url{https://github.com/facebookresearch/faiss/blob/main/faiss/IndexFlat.h}} method based on Faiss~\footnote{\url{https://github.com/facebookresearch/faiss}} to index the vector for Approximate Nearest Neighbor (ANN)~\cite{li2019approximate} search.

\paragraph{Inference}
We use each ``\texttt{[Question]}'' in multi-step reasoning of DxDirector-7B as the query to our search model. For each query, we use our trained search model to represent it to a vector. Then, we use ANN to search the indexed corpus for the paragraph vector that is closest to the query vector and obtain the paragraph corresponding to this paragraph vector as top-1 ranked paragraph in the search result. We mark this paragraph with corresponding serial number and attach it to the ``\texttt{[Final Content]}''. In this way, we provide authoritative medical literature as a reference for each step in complex clinical diagnosis, which enables diagnostic readers to verify the diagnostic content more conveniently and judge the credibility of the diagnostic content in a fine-grained manner.

\subsection{Details about Experiments}

\subsubsection{Baselines}
The baselines of the experiments can be divided into two categories. One is the open source LLMs specifically optimized for medical scenarios and open source general-purpose LLMs:
\begin{enumerate}[leftmargin=*, noitemsep] 
    \item \textbf{Meditron-70B}, it is a medical adapted LLM based on Llama-2-70B, with the continued pretraining on a comprehensively curated medical corpus, including selected PubMed articles, abstracts, and internationally-recognized medical guidelines.
    \item \textbf{OpenbioLLM-70B}, it is an advanced LLM designed specifically for the biomedical domain based on Llama-3-70B. It achieves state-of-the-art performance on a wide range of biomedical tasks.
    \item \textbf{Clinical Camel-70B}, it is a medically adapted LLM fine-tuned on the Llama-2 70B architecture using QLoRA. It is tailored for the medical and clinical research, capable of processing and generating relevant content.
    \item \textbf{Meditron-176B}, it is a generalist medical LLM with 176 billion parameters, pre-trained on a large-scale medical text and real-world clinical records. It shows promising clinical diagnosis performance for cases with complete clinical information.
\end{enumerate}
The other is the current most powerful commercial general-purpose large language models:
\begin{enumerate}[leftmargin=*, noitemsep] 
    \item \textbf{GPT-4o}, it is a commercial general-purpose LLM developed by OpenAI. It shows better performance on clinical diagnosis than many medical adapted LLMs such as Meditron-70B, Clinical Camel-70B and Med-Palm-540B.
    \item \textbf{o1-preview}, it is a commercial general-purpose LLM developed by OpenAI. Compared with GPT-4o, it demonstrates more powerful reasoning capability and recent study has shown that it surpasses human accuracy in making the final clinical diagnosis~\cite{brodeur2024superhuman}.
    \item \textbf{o3-mini}, it is a commercial general-purpose LLM developed by OpenAI. Compared with o1-preview, its reasoning ability has been upgraded again, surpassing GPT-4o and o1-preview in many complex tasks.
    \item \textbf{Gemini-2.0-flash}, it is a commercial general-purpose LLM developed by Google. Its shows better performance than Google’s former PaLM 2.
    \item \textbf{Deepseek-V3-671B}, it is a general-purpose developed by Deepseek. It surpasses GPT-4o in general language understanding and generation capabilities.
\end{enumerate}

\subsubsection{Evaluation Metrics}
We propose four evaluation metrics in this paper:
\begin{enumerate}[leftmargin=*, noitemsep] 
    \item \textbf{Accuracy of Diagnosis.} This is a fully automatic evaluation metric driven by an LLM-based agent, which is used for four publicly available medical datasets including NEJM Clinicopathlogic Cases, RareArena, ClinicalBench and USMLE. Specifically, we use instructions to make the final diagnoses generated by LLMs conform to the format of ``So the final answer is ...'', so that we can extract the short and clear diagnoses generated by LLMs. For each sample, we compare the extracted diagnosis with the correct answer provided by the datasets to determine whether the generated diagnosis matches the correct answer. This is automatically done based on the gpt-4o-mini agent. We give gpt-4o-mini instructions and examples so that it can make the correct judgment. We statistically analyze the diagnostic accuracy to compare the capabilities of different LLMs.
    \item \textbf{Number of Clinical Operations.} This metric calculates the average number of operations that human physicians are required to perform when different LLMs complete the full diagnostic process over the whole datasets (the lower the better), which quantifies the workload of human doctors.
    \item \textbf{Effectiveness of Clinical Operations.} This metric calculates the average of the proportion of operations that are truly useful for making a diagnosis out of all requested operations over the whole datasets. We determine whether an operation is helpful for diagnosis by assessing whether it appears in the case report provided from medical specialists. This metric reflects the accuracy of LLMs in determining which operations are necessary for diagnosis. 
    \item \textbf{Scoring Referred from Medical Specialists.} This metric is applied in real-world diagnosis scenario. Evaluation by medical experts is an important way for us to understand the gap between AI and human physicians in real-world diagnosis. In order to establish an objective evaluation mechanism to prevent bias in human physicians’ scoring, we adopt Double-Blinded Adjudication, which makes LLMs and human physicians to give diagnoses to the same patient without seeing each other’s content. We introduce a third-party agent to score an LLM based on the degree of match between LLM's diagnosis and human physicians' diagnosis. The third-party agent is based on GPT-4o and Deepseek-V3, and the average of the scores given by them is taken as the actual score. The score range is between 0 and 10.
\end{enumerate}

\subsubsection{Ethical Review and Assurance of Clinical Study in Real-World Diagnosis}
Our real-world clinical diagnostic scenario is set within an officially certified Grade 3A hospitals in China. The involved patients are inpatients presenting with more complex conditions than typical outpatients. Consequently, LLMs must engage in intricate reasoning to gather comprehensive clinical information effectively. To safeguard patients from potential harm, the evaluation environment is structured as follows: patient behaviors and medical specialist operations during clinical diagnosis are fully recorded using actual inpatient records. Subsequently, two GPT-4o-based agents replicate precisely the recorded behaviors of patients and specialists throughout the diagnostic process. In evaluation, LLMs interact with these agents to drive the full-process diagnosis, initiating solely from the patient's vague chief complaint. Within this controlled environment, LLMs do not directly interact with real patients, and their diagnostic outputs undergo rigorous review by medical specialists, thereby effectively mitigating ethical risks and potential harm. The evaluation is performed on 160 cases across 9 different clinical departments including Gastroenterology, Nephrology, Dermatology, Cardiovascular Medicine, Infectious Diseases, Endocrinology, Pulmonology, General Surgery, and Pain Management.

To ensure objective assessments and mitigate potential biases in human specialists scoring, a double-blind adjudication approach is implemented. In this approach, both human specialists and LLMs independently diagnose the same patient cases without exposure to each other's diagnostic outputs. Additionally, a third-party evaluation agent, utilizing both GPT-4o and Deepseek-V3, assigns scores ranging from 0 to 10 based on the alignment between LLM-generated diagnoses and those provided by medical specialists. The final score is calculated as the average of the scores given by GPT-4o and Deepseek-V3, thus ensuring robust and unbiased comparative assessment. The assessment of whether the diagnoses generated by LLMs could fully replace those made by medical specialists also follows the same pattern by observing the decisions of the third party agent (can or cannot).

All data utilized in this research are exclusively for academic purposes, acquired ethically and legally, and have been reviewed and approved by the relevant institutional ethics committee (IRB00006761-M20250173), ensuring adherence to ethical and legal standards. Data collection rigorously complies with principles of patient privacy protection. No hospital-related information is disclosed, and to further protect patient confidentiality, all personally identifiable information (PII), treatment locations, and other sensitive details are systematically identified and removed by the medical team.

\subsubsection{Statistical Information}
The error bars reflecting $95\%$ confidence intervals are determined by non-parametric bootstrap procedure with 1,000 samples. As for accuracy, we perform statistical significance tests utilizing the two-side McNemar test between DxDirector-7B and the top-3 baseline, with p-value levels annotated on the bars. As for number of clinical operations, effective of clinical operations and scoring referred from specialists, we use two-side Mann-Whitney U test for statistical significance tests.

\subsubsection{Implementations} \label{implementation}
For continued pre-training and instruction-tuning stage, we use DeepSpeed framework in zero stage 3 to train DxDirector-7B with full parameter fine-tuning on 4 Nvidia A100 80G GPUS. In continued pre-training, we follow Meditron~\cite{chen2023meditron} to set $\beta_1=0.9$, $\beta_2=0.95$, eps$=10^{-5}$ for the AdamW optimizer. The learning rate is $3\times10^{-4}$. The weight decay is $0.1$. The batch size is $1$. In instruction-tuning stage, we set $\beta_1=0.9$, $\beta_2=0.95$, eps$=10^{-5}$ for the AdamW optimizer. The learning rate is $9.65\times10^-6$. The weight decay is $0$. The batch size is $1$ and training epochs is $3$. In step-level strategy preference optimization, first, we input the same prefix to make DxDirector-7B generate multiple different replies by changing random seed at each generation, with sampling parameters as $0.6$ temperature, $0.95$ top-p and $20$ top-k. After data construction, we use the open-source reinforcement learning framework trl~\footnote{\url{https://github.com/huggingface/trl}} to train our model with learning rate as $5\times10^{-7}$, gradient accumulation steps as $8$ and batch size as $1$. During the evaluation, we cancel the random sampling setting so that the content generated by LLMs can be fully reproduced.

\section{Data Availability}
After the external review, we promise to provide the datasets for continued pre-training, instruction-tuning and step-level stragety optimization on GitHub~\footnote{\url{https://github.com/}}. This ensures unrestricted access for anyone to use the data for any purpose, fostering future research and development. We can provide the public available datasets here: RareArena:~\url{https://github.com/zhao-zy15/RareArena}, NEJM Clinicopathologic Cases:~\url{https://www.nejm.org/browse/nejm-article-category/clinical-cases?date=past5Years}, ClinicalBench:~\url{https://github.com/WeixiangYAN/ClinicalLab}, USMLE and MedQA:~\url{https://drive.google.com/file/d/1ImYUSLk9JbgHXOemfvyiDiirluZHPeQw/view}.

\section{Code Availability}
 To prompt the future research, we promise to release all source codes to train our DxDirector-7B on Github, and the model weights on Huggingface~\footnote{\url{https://huggingface.co/}} after the external review. This ensures unrestricted access for anyone to use the code and model weights for any purpose, fostering future research and development. Combined with our detailed introduction to the method and the open source datasets, we believe that researchers will find it very easy to reproduce our work. 

\bibliographystyle{plain}
\bibliography{custom}

\section{Extended Data}
\begin{figure}[t]
    \centering
        \includegraphics[width=1.0\linewidth]{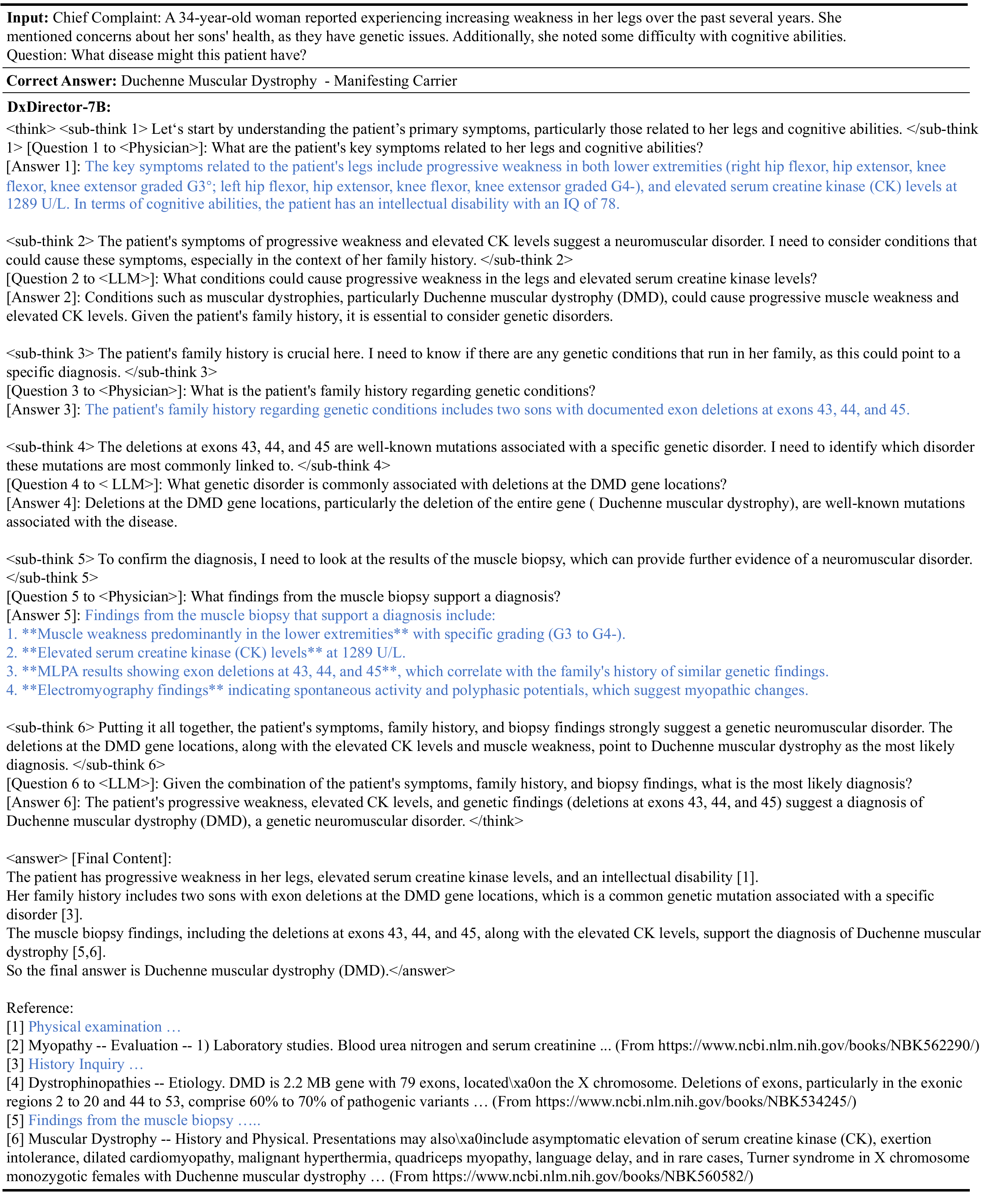}
        \caption{Our DxDirector-7B on case 1. \textcolor{customblue2}{Blue fonts} represent physician input, and others are generated by DxDirector-7B. In reference, the \textcolor{customblue2}{blue fonts} record the physician actions, and others are retrieved from authoritative medical literature to provide a verification mechanism for each step in the reasoning. The reference numbers correspond to the steps of thinking-question-answer.}
        \label{case1}
\end{figure}

In Fig.~\ref{case1} to Fig.~\ref{case3-bs}, we show the comparison of the actual output of DxDirector-7B and baseline LLMs for specific cases.

\begin{figure}[t]
    \centering
\includegraphics[width=1.0\linewidth]{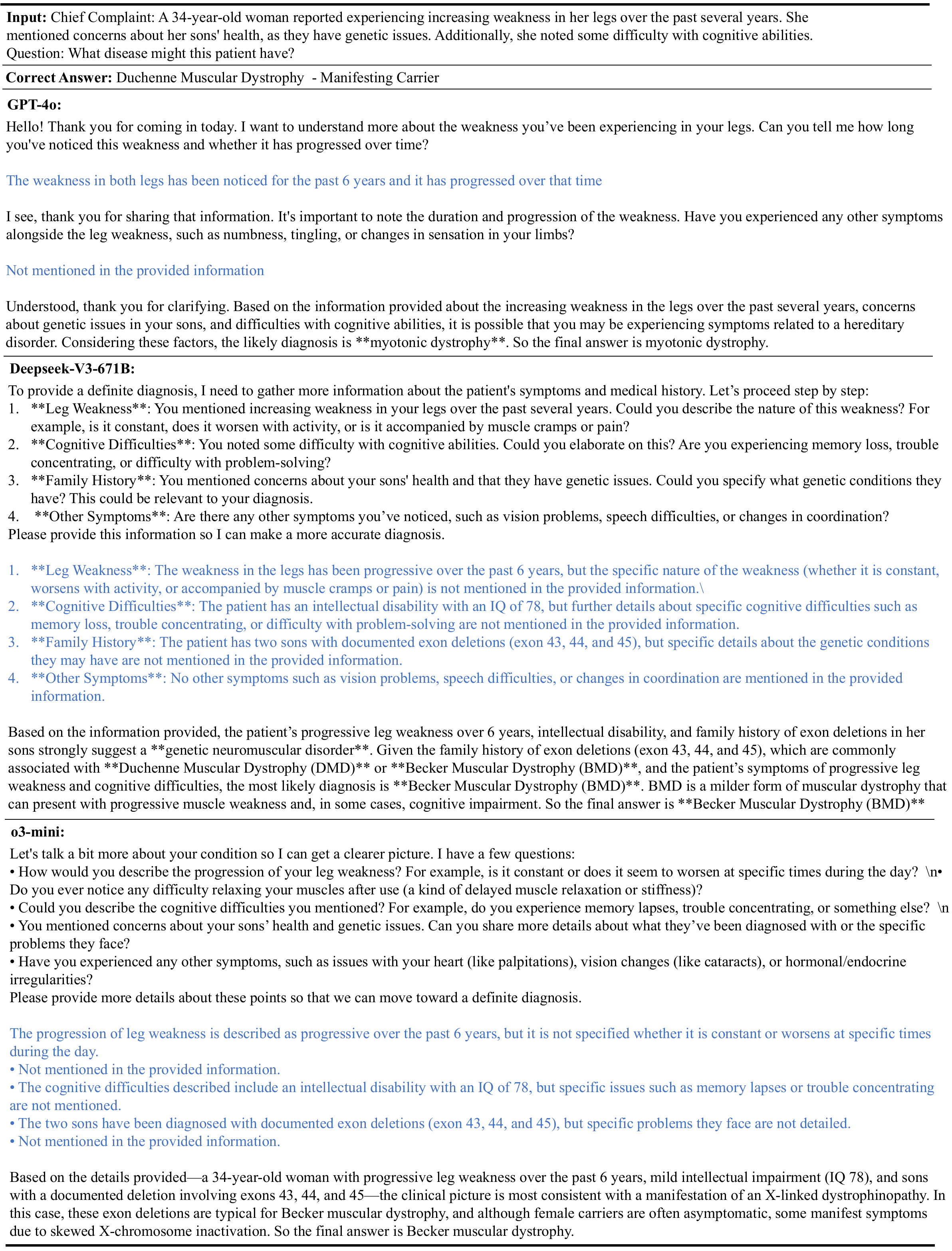}
        \caption{Baselines on case 1. \textcolor{customblue2}{Blue fonts} represent physician input. ``Not mentioned'' means the required clinical operations are not provided in clinical information of the case reports. The specific prompt enables baselines to perform full-process diagnosis is in Supplementary Fig.21}
        \label{case1-bs}
\end{figure}

\begin{figure}[t]
    \centering
        \includegraphics[width=1.0\linewidth]{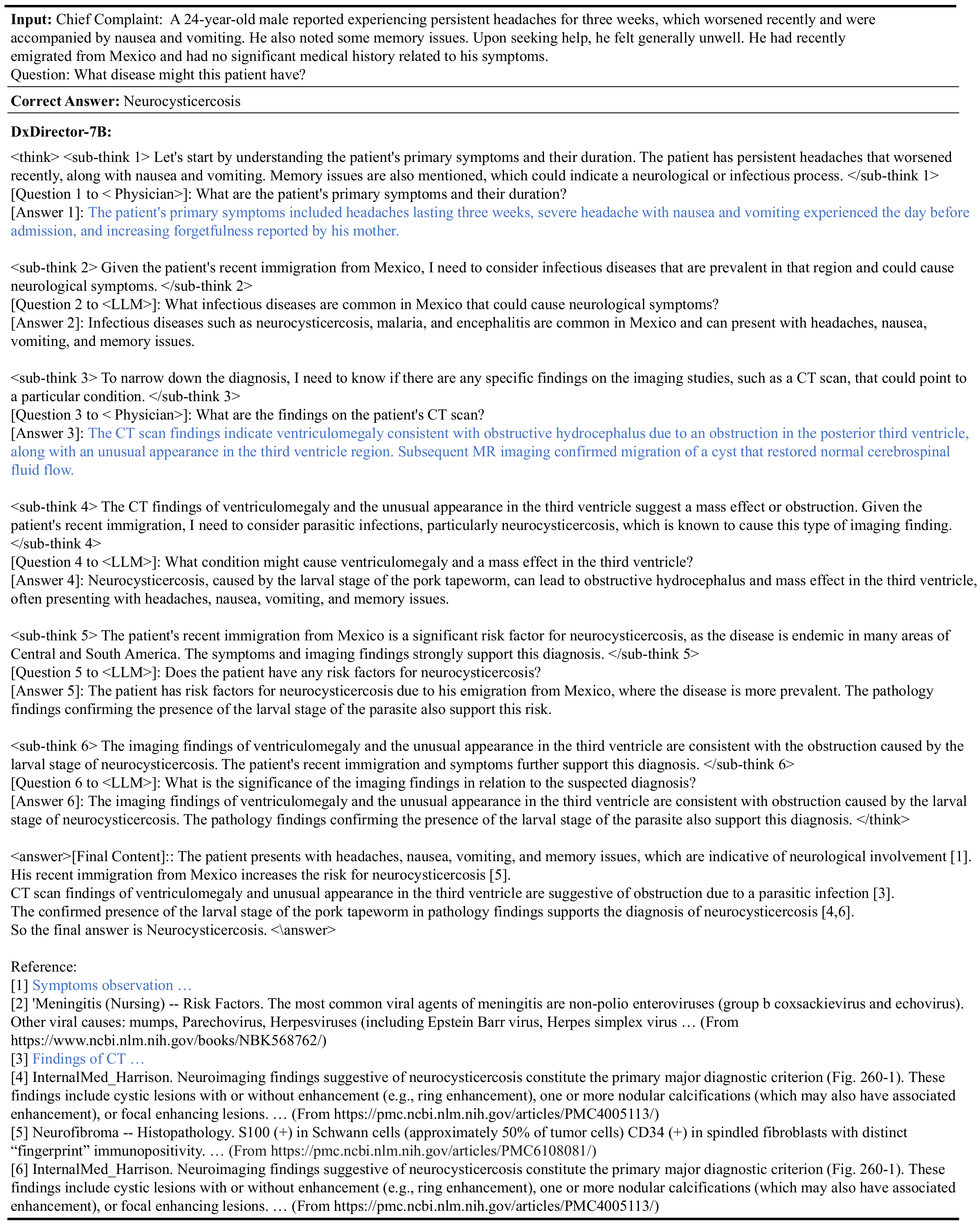}
        \caption{Our DxDirector-7B on case 2. \textcolor{customblue2}{Blue fonts} represent physician input, and others are generated by DxDirector-7B. In reference, the \textcolor{customblue2}{blue fonts} record the physician actions, and others are retrieved from authoritative medical literature to provide a verification mechanism for each step in the reasoning. The reference numbers correspond to the steps of thinking-question-answer.}
        \label{case2}
\end{figure}

\begin{figure}[t]
    \centering
\includegraphics[width=1.0\linewidth]{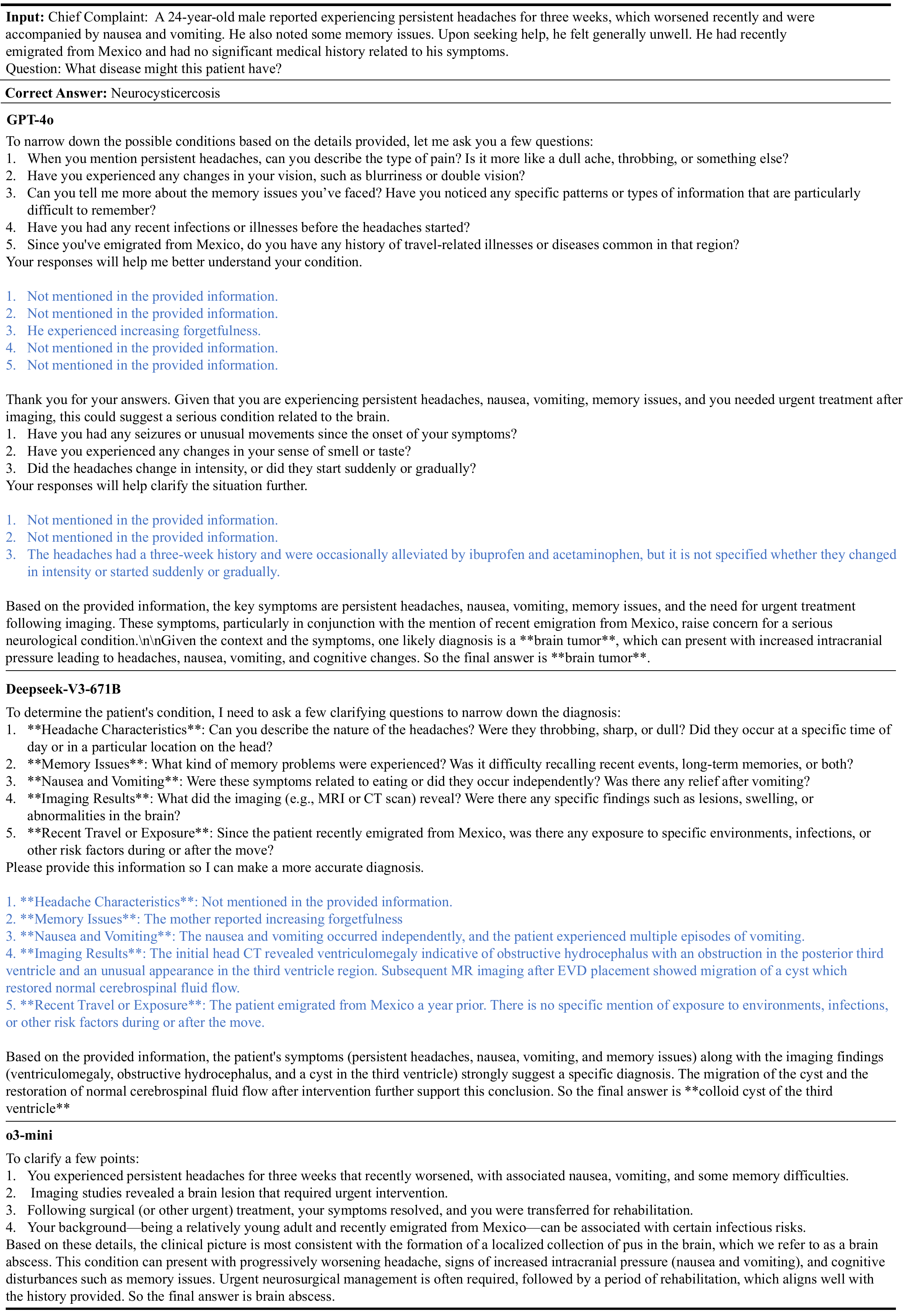}
        \caption{Baselines on case 2. \textcolor{customblue2}{Blue fonts} represent physician input. ``Not mentioned'' means the required clinical operations are not provided in clinical information of the case reports. The specific prompt enables baselines to perform full-process diagnosis is in Supplementary Fig.21}
        \label{case2-bs}
\end{figure}

\begin{figure}[t]
    \centering
        \includegraphics[width=1.0\linewidth]{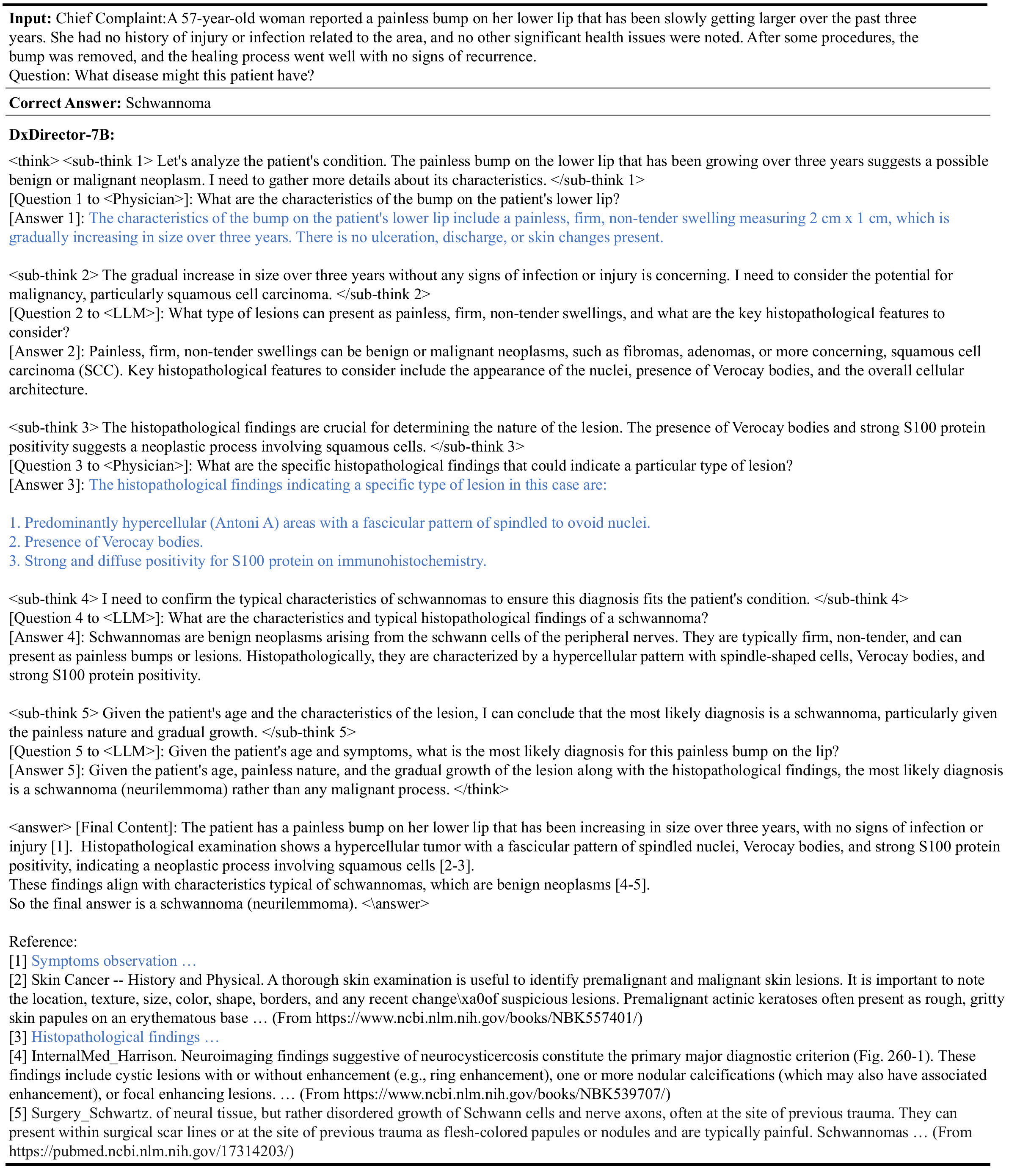}
        \caption{Our DxDirector-7B on case 3. \textcolor{customblue2}{Blue fonts} represent physician input, and others are generated by DxDirector-7B. In reference, the \textcolor{customblue2}{blue fonts} record the physician actions, and others are retrieved from authoritative medical literature to provide a verification mechanism for each step in the reasoning. The reference numbers correspond to the steps of thinking-question-answer.}
        \label{case3-bs}
\end{figure}

\begin{figure}[t]
    \centering
\includegraphics[width=1.0\linewidth]{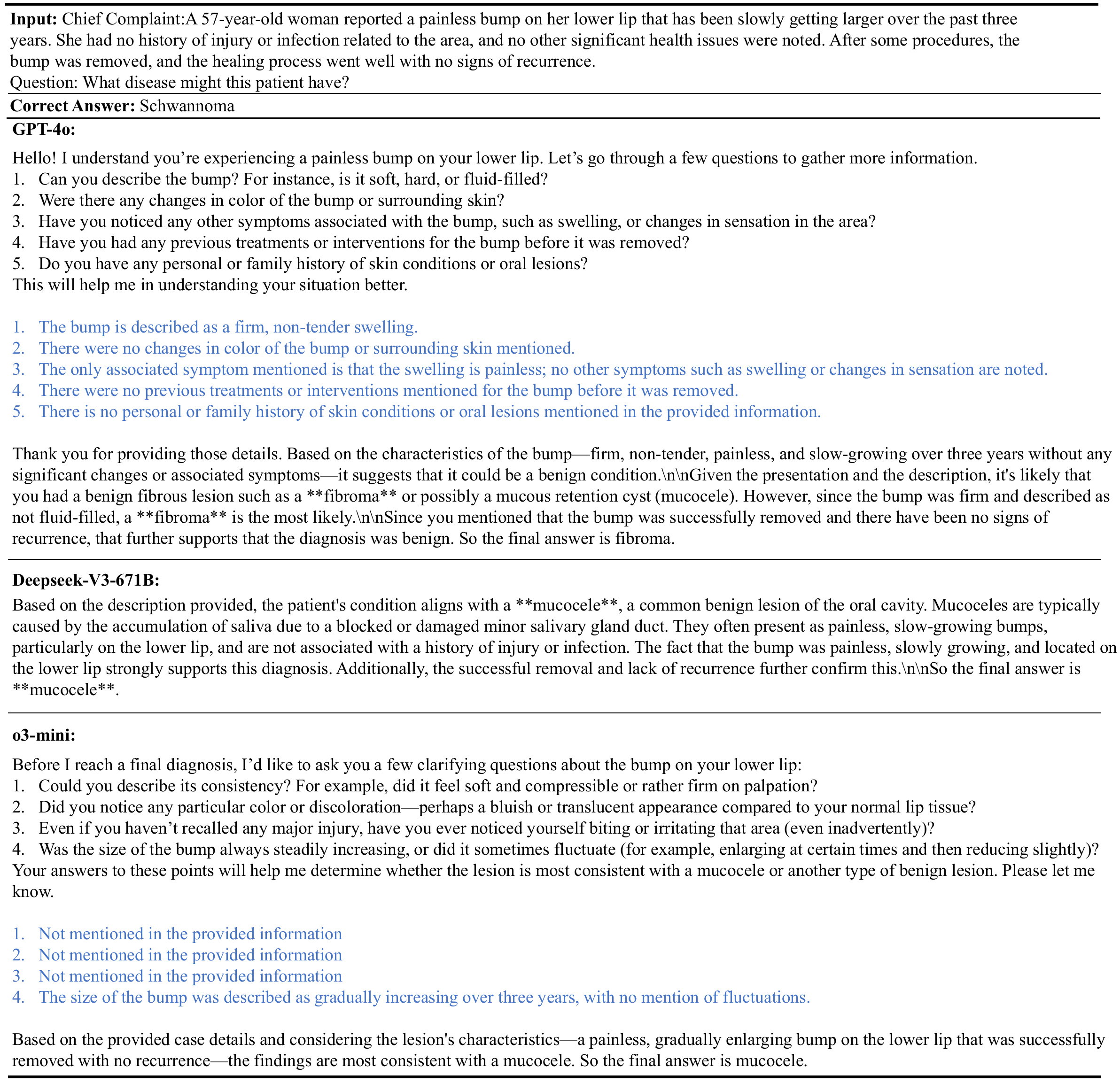}
        \caption{Baselines on case 3. \textcolor{customblue2}{Blue fonts} represent physician input. ``Not mentioned'' means the required clinical operations are not provided in clinical information of the case reports. The specific prompt enables baselines to perform full-process diagnosis is in Supplementary Fig.21}
        \label{case3-bs}
\end{figure}

\end{document}